\documentclass{article}

\usepackage{microtype}
\usepackage{graphicx}
\usepackage{subcaption}  
\usepackage{booktabs}

\usepackage{amsmath}
\usepackage{amssymb}
\usepackage{mathtools}
\usepackage{amsthm}

\usepackage{marvosym}

\usepackage{url}
\usepackage{caption}
\usepackage{multirow}
\usepackage{makecell}
\usepackage{xcolor}
\usepackage{colortbl}
\usepackage{wrapfig}
\usepackage{enumitem}

\usepackage{hyperref}

\usepackage[accepted]{icml2026}

\usepackage[capitalize,noabbrev]{cleveref}

\theoremstyle{plain}

\theoremstyle{definition}

\theoremstyle{remark}

\newcommand{\method}{ShotKV}
\newcommand{\bcmk}{KVFundaBench}

\icmltitlerunning{Semantic Integrity Matters: Benchmarking and Preserving High-Density Reasoning in KV Cache Compression}

\begin{document}

\twocolumn[
\icmltitle{Semantic Integrity Matters: Benchmarking and Preserving High-Density Reasoning in KV Cache Compression}

\begin{icmlauthorlist}
\icmlauthor{Xiang Liu}{hkustgz}
\icmlauthor{Zhenheng Tang}{hkust}
\icmlauthor{Hong Chen}{hkustgz}
\icmlauthor{Peijie Dong}{hkustgz}
\icmlauthor{Zeyu Li}{hkustgz}
\icmlauthor{Xiuze Zhou}{hkustgz}
\icmlauthor{Bo Li}{hkust,FYT}
\icmlauthor{Xuming Hu\texorpdfstring{$^{\text{\Letter}}$}{}}{hkustgz}

\icmlauthor{Xiaowen Chu\texorpdfstring{$^{\text{\Letter}}$}{}}{hkustgz}
\end{icmlauthorlist}

\icmlaffiliation{hkustgz}{The Hong Kong University of Science and Technology (Guangzhou), Email: xliu886@connect.hkust-gz.edu.cn}
\icmlaffiliation{hkust}{The Hong Kong University of Science and Technology}
\icmlaffiliation{FYT}{Guangzhou HKUST Fok Ying Tung Research Institute}

\icmlcorrespondingauthor{Xuming Hu}{xuminghu@hkust-gz.edu.cn}
\icmlcorrespondingauthor{Xiaowen Chu}{xwchu@hkust-gz.edu.cn}

\icmlkeywords{Large Language Models, KV Cache Compression, Benchmark, Inference Efficiency}

\vskip 0.3in
]

\printAffiliationsAndNotice{}

\begin{abstract}

While Key-Value (KV) cache compression is essential for efficient LLM inference, current evaluations disproportionately focus on \textbf{retrieval-oriented} long-context tasks, potentially masking the degradation of High-Density Reasoning where Chain-of-Thought (CoT) coherence is critical. We introduce KVFundaBench to systematically evaluate this gap, revealing a sharp dichotomy: while retrieval tasks remain robust, reasoning tasks exhibit severe Task-Dependent Degradation under aggressive compression due to disrupted CoT links. Extending our analysis to the DeepSeek-R1 model, we uncover that its specialized attention patterns offer unique insights into the fragility of reasoning chains. Guided by these findings—specifically the necessity of preserving few-shot examples as indivisible \textbf{Semantic Units}—we propose ShotKV. This approach explicitly separates prefill and decoding phases to prioritize semantic integrity. Empirical results demonstrate that ShotKV achieves 9\%-18\% accuracy improvements on long-context generation tasks and effectively generalizes to document QA, all while delivering an 11\% latency reduction compared to full cache inference.

\end{abstract}

\section{Introduction}
\label{sec:introduction}

The evolution of Large Language Models (LLMs) to process large documents for tasks such as answering and summarizing questions~\citep{raffel2020exploring, chowdhery2022palm,yu2026rethinking,touvron2023llama, touvron2023llama2,tang2026memory-survey}, spurred by breakthroughs in system architectures~\citep{flash-attn, flash-attn2, jacobs2023deepspeed, streamingllm, zhang2023dissecting,huang2026mdnparallelizingstepwisemomentum} and model design~\citep{chen2023extending, xiong2023effective, chen2023longlora, peng2024yarn}, has significantly increased GPU memory demands during inference~\citep{jamba, grok, geminiteam2024gemini, claude3, deepseekv2,deepseekv3}, making the development of efficient key value (KV) cache compression strategies a critical focus for LLM deployment and optimization. This concern is shared across the broader efficiency stack, including memory-efficient fine-tuning~\citep{pan2024lisa}, token-efficient adaptive inference~\citep{liu2026diffadapt}, and reasoning-model inference serving~\citep{li2026reasoningserving}; KV-level intervention is the axis we focus on here.

\begin{figure}[t]
    \centering
    \begin{subfigure}[b]{0.48\textwidth}
        \includegraphics[scale=0.2, trim=60 0 30 0]{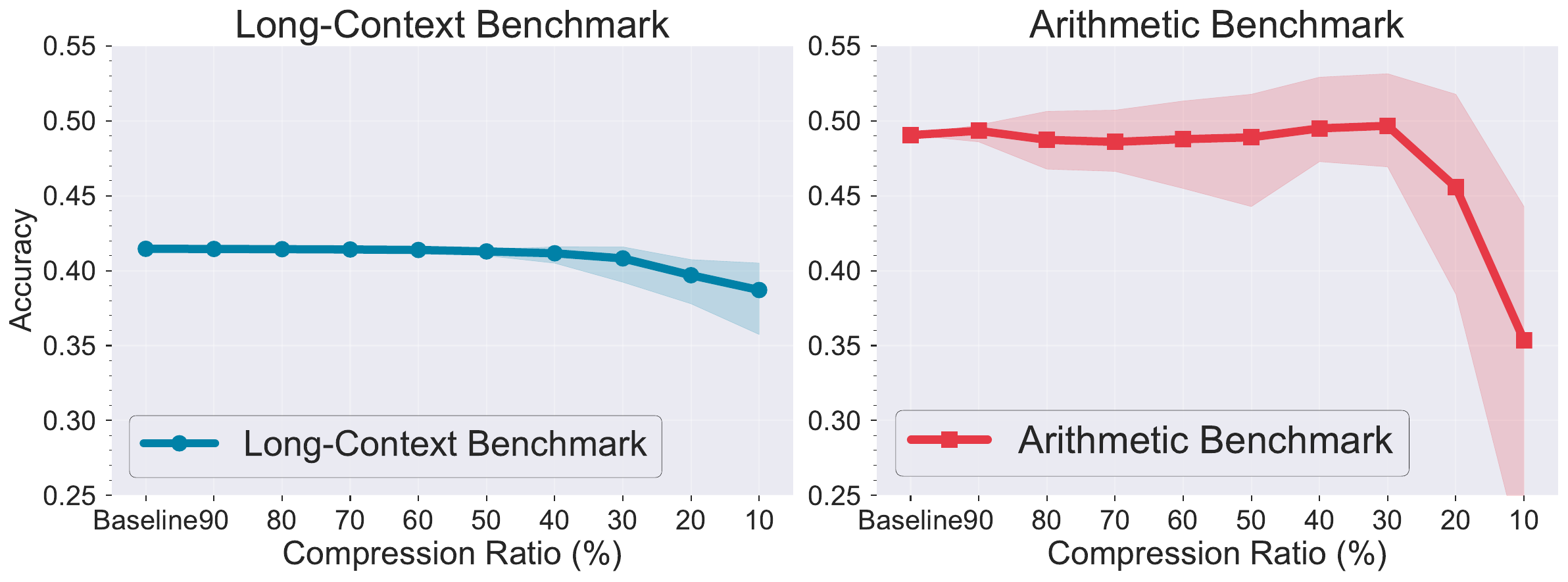}
        \caption{KV cache compression methods on long-context and arithmetic benchmarks.}
        \label{fig:first_1a}
    \end{subfigure}
    \vspace{-15pt}

    \begin{subfigure}[b]{0.48\textwidth}
        \includegraphics[scale=0.5, trim=30 15 40 0]{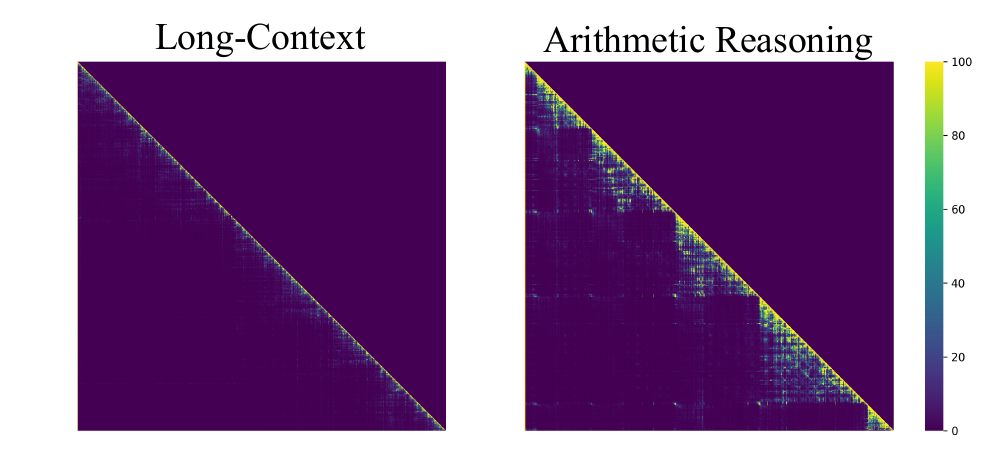}
        \caption{Attention heatmap.}
        \label{fig:first_1b}
    \end{subfigure}
    \vspace{-15pt}
    \caption{KV cache compression methods on long-context and arithmetic benchmarks.
    (a) Arithmetic benchmark shows more performance degradation than long-context benchmark.
    (b) Long-Context benchmark shows more sparsity in attention heatmap.}
    \vspace{-20pt}
    \label{fig:first_1}
\end{figure}

To address this, numerous studies have proposed selective token retention strategies~\citep{streamingllm, h2o, snapkv, ge2023model, pyramidkv, fu2024lazyllm, yang2024pyramidinfer, adnan2024keyformer, liu2024scissorhands, tang2024quest}, with pioneering works such as H2O~\citep{h2o} and SnapKV~\citep{snapkv} showing that retaining approximately 50\% of KV cache entries can balance model performance with significant memory savings.
However, current research predominantly evaluates these methods on retrieval-oriented long-context benchmarks, where the dominant difficulty is locating salient spans in a long context---e.g., Needle-In-A-Haystack (NIAH)~\citep{needle} and many tasks in LongBench~\citep{longbench,longbenchv2}. We argue that this evaluation focus can mask the true fragility of compressed models on a different regime, which we term \textbf{High-Density Reasoning}: tasks where nearly every token in a multi-step prompt (e.g., chain-of-thought few-shot exemplars) is reasoning-critical, in contrast to retrieval tasks where only a small ``needle'' subset truly matters. As illustrated in \cref{fig:first_1}(a), under matched compression methods, an arithmetic-reasoning workload exhibits sharp degradation while a retrieval-oriented long-context workload remains comparatively stable; the corresponding attention heatmap in \cref{fig:first_1}(b) shows the contrasting sparsity patterns that motivate this study. Under matched settings, many-shot reasoning at moderate context lengths can be \emph{more} compression-sensitive than retrieval over much longer contexts, because breaking a single semantic link in the Chain-of-Thought (CoT) can lead to catastrophic reasoning failure. Consequently, the impact of compression on a spectrum of fundamental LLM capabilities---such as \textit{arithmetic reasoning}, \textit{world knowledge}, \textit{commonsense reasoning}, and \textit{safety}---remains largely unexplored, particularly concerning their distinct attention patterns.

To this end, we introduce \textbf{\bcmk{}}, a benchmark designed to systematically assess the effects of KV cache compression across these diverse fundamental capabilities and their underlying attention dynamics. The benchmark includes 5 categories of tasks: \textit{world knowledge}, \textit{commonsense reasoning}, \textit{arithmetic reasoning}, \textit{code generation}, and \textit{safety}. Utilizing \bcmk{}, we conduct an extensive evaluation of \textbf{six representative KV cache compression methods} across \textbf{four state-of-the-art LLMs}, including both standard instruction-tuned models and specialized reasoning models. Our \bcmk{}  reveals several key findings: (1) \textit{Task-Dependent Degradation}: Performance degradation is highly task-dependent, with arithmetic reasoning tasks showing particular sensitivity to aggressive compression; (2) \textit{Model-Type Robustness}: Multi-step reasoning LLMs demonstrate higher compression robustness compared to instruction-tuned models; (3) \textit{Prompt Length Vulnerability}: Shorter prompts are more vulnerable to compression effects; (4) \textit{Chunk-Level Superiority}: Chunk-level compression strategies show superior performance on complex long-context reasoning tasks; (5) \textit{Prompt-Gain Sensitivity}: Tasks with larger prompt-based performance gains exhibit higher compression sensitivity; and (6) \textit{Long-Context Generation Sensitivity}: Long-context generation tasks are particularly sensitive to compression. These findings motivate \textbf{\method{}}, a lightweight, \emph{insight-driven proof-of-concept} that operationalizes the central conclusion of \bcmk{}: preserving few-shot examples as indivisible \textbf{Semantic Units}. \method{} is intentionally simple---a task-motivated combination of shot-aware prefill preservation and dynamic decoding compression---and its purpose is to validate, not to claim algorithmic novelty over, the benchmark's hypothesis that semantic-unit preservation is key to robustness under aggressive compression.



Our main contributions are summarized as follows:
\begin{itemize}[leftmargin=*, topsep=0pt, itemsep=2pt, parsep=0pt]
    \item\noindent We introduce \textbf{KVFundaBench} to systematically evaluate the effects of KV cache compression on \textbf{High-Density Reasoning} and fundamental capabilities.
    
    \item\noindent Our systematic investigation reveals multiple critical factors influencing compression sensitivity, including model training dynamics, prompt length characteristics, task-specific requirements, long-context reasoning, and long-context generation capabilities. 

    \item\noindent  We propose \method{}, a semantic-aware strategy that treats few-shot examples as indivisible \textbf{Semantic Units}. By explicitly separating prefill and decoding compression, \method{} achieves \textbf{9\%-18\%} accuracy gains on long-context generation and generalizes to document QA, while delivering an \textbf{11\% reduction in end-to-end latency}
\end{itemize}
\vspace{-5pt}

\vspace{-5pt}
\paragraph{Conflict of Interest Disclosure.} All authors are affiliated with academic institutions; financial support is fully disclosed in the Acknowledgments. The authors declare no additional competing financial or non-financial interests that could be perceived to influence this work.

\vspace{-5pt}
\section{Related Work}
\label{sec:related_work}
\vspace{-0.2cm}
\paragraph{KV cache compression.}
A growing line of work reduces KV memory by selecting a subset of tokens to retain. Token-level methods such as StreamingLLM~\citep{streamingllm}, H2O~\citep{h2o}, SnapKV~\citep{snapkv}, FastGen~\citep{ge2023model}, Scissorhands~\citep{liu2024scissorhands}, Quest~\citep{tang2024quest}, Keyformer~\citep{adnan2024keyformer}, and value-aware variants~\citep{guo2024attention} score individual tokens via attention statistics and evict the rest. Layer- or budget-aware variants such as PyramidKV~\citep{pyramidkv}, PyramidInfer~\citep{yang2024pyramidinfer}, and Ada-KV~\citep{feng2024ada} adapt the budget across layers or heads, while cross-layer designs (CLA~\citep{brandon2024reducing}, MiniCache~\citep{liu2024minicache}, YOCO~\citep{sun2024yoco}, LCKV~\citep{wu2024layercondensedkvcacheefficient}) share or merge KV caches across layers. Most relevant to us, ChunkKV~\citep{chunkkv} preserves contiguous chunks rather than isolated tokens, and SCOPE~\citep{scope} separates prefill and decoding compression. Recent work also benchmarks compression breadth~\citep{yuan2024kv,kvpress,kim2025kvzip,liu2025flowkv,zhu2025oraclekv,chen2026sonic}, asks which capabilities should be preserved under compression~\citep{tang2025lottery,dong2025compressedllm}, and reuses caches for RAG~\citep{yao2024cacheblend}. Orthogonal to token retention, KV \emph{quantization}~\citep{li2025antkv} reduces memory by lower-precision representations and can be combined with retention-based methods.
\vspace{-0.2cm}
\paragraph{Long-context evaluation.}
Existing long-context benchmarks predominantly emphasize \emph{retrieval-oriented} settings: NIAH~\citep{needle}, RULER~\citep{hsieh2024ruler}, $\infty$-Bench~\citep{zhang2024infty}, and many tasks in LongBench~\citep{longbench,longbenchv2} measure how well a model locates salient spans within long contexts. Generation-oriented evaluation~\citep{longgenbench} and many-shot ICL~\citep{agarwal2024many} are more recent and remain comparatively underexplored under compression. \bcmk{} complements these by systematically probing fundamental capabilities~\citep{mmlu,bbh,gsm8k,csqa,chen2021evaluating,lin2021truthfulqa,hartvigsen2022toxigen} under matched compression budgets, and \method{} positions itself as a shot-aware combination of (i) chunk-style semantic-unit preservation in prefill and (ii) dynamic decoding compression---designed not as a new algorithmic primitive, but as a targeted instantiation of the insight that emerges from \bcmk{}. An extended discussion is provided in \cref{appendix:related_work}.

\vspace{-5pt}
\section{Preliminary}
In this section, we provide comprehensive preliminaries of KV cache compression and LLM evaluation.
\vspace{-0.3cm}
\paragraph{Key-Value Cache in LLMs}
With the increasing long-context capabilities of LLMs, the Key-Value (KV) cache has become crucial for improving inference efficiency. During LLM inference, the KV cache stores intermediate computation results to avoid redundant calculations. For a given input sequence $x = (x_1, x_2, ..., x_n)$, each transformer layer $l$ maintains its key cache $K^l = (k^l_1, k^l_2, ..., k^l_n)$ and value cache $V^l = (v^l_1, v^l_2, ..., v^l_n)$, where $k^l_i, v^l_i \in \mathbb{R}^d$ represent the key and value vectors for token $x_i$ at layer $l$.

\vspace{-0.3cm}
\paragraph{KV Cache Compression}
KV cache compression aims to reduce memory usage by selectively storing or merging cached vectors. A compression operation can be denoted as $C(K,V) = (K',V')$, where $K'$ and $V'$ are compressed caches with size $m < n$, where $C$ is the compression method, $m$ is the number of retained tokens, and $n$ is the original number of tokens. The core idea is token selection - identifying and retaining important tokens based on attention patterns or other metrics while discarding less important ones. The compression ratio $r = m/n$ indicates how aggressively the cache is compressed, where a smaller ratio means more aggressive compression.

\vspace{-0.3cm}
\paragraph{Evaluation Protocol}
To thoroughly evaluate the impact of KV cache compression on LLMs' capabilities, we assess five benchmark categories: world knowledge, commonsense reasoning, arithmetic reasoning, code generation, and safety.

For each task category and compression method $C$, we calculate the relative performance change as follows:
\vspace{-5pt}
\begin{equation} 
\label{eq:performance_change}
    \Delta P = \frac{P_C - P_{\text{base}}}{P_{\text{base}}}
\vspace{-5pt}
\end{equation}
where $P_C$ and $P_{\text{base}}$ represent the performance scores with and without compression, respectively.


\vspace{-5pt}
\section{Benchmark Design}
\label{sec:benchmark_design}
\subsection{Benchmark Setups}

In this section, we will introduce the \bcmk{} setups, including the datasets, models, and evaluation environment.

\begin{figure}[t]
    \centering
    \includegraphics[scale=0.9, trim=20 0 0 0]{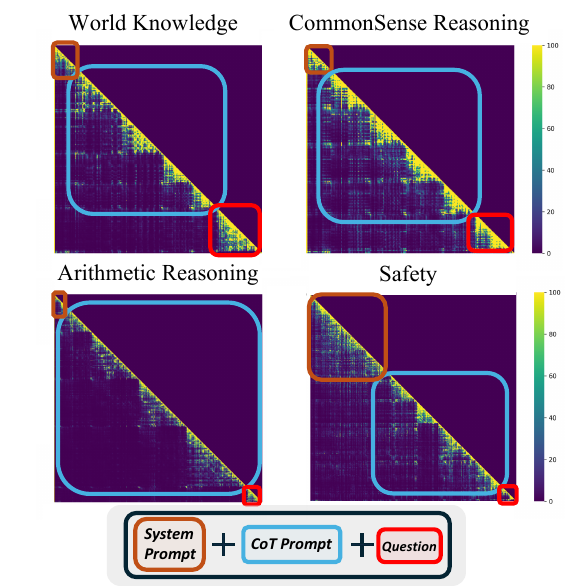}

    \vspace{-5pt}
    \caption{Attention heatmap on different tasks.}
    \vspace{-15pt} 

    \label{fig:main}
\end{figure}

\vspace{-0.3cm}
\paragraph{Datasets}
To evaluate the performance of KV cache compression on LLMs' overarching capabilities, we assess five benchmark categories: \textbf{\textit{World Knowledge (WK)}} using MMLU~\citep{mmlu}, measured by accuracy; \textbf{\textit{CommonSense Reasoning (CSR)}} using CommonsenseQA~\citep{csqa} , evaluated through multiple-choice accuracy; \textbf{\textit{Arithmetic Reasoning (AR)}} using GSM8K~\citep{gsm8k}, assessed by solve rate; \textbf{\textit{Code Generation (CG)}} using HumanEval~\citep{chen2021evaluating}, measured by pass@1 rate on test cases; and \textbf{\textit{Safety (SA)}} using JailBreakV~\citep{luo2024jailbreakv}, evaluated by attack success rate. Furthermore, we test the performance of KV cache compression on LongGenBench~\citep{longgenbench}, a \textbf{\textit{long-context generation (LG)}} benchmark. Detailed statistics for all datasets are provided in \cref{app:eval_bench_dataset}.

\vspace{-0.2cm}
\paragraph{Models}
We conduct experiments on a series of LLMs, including LLaMA-3.1-8B, LLaMA-3.1-8B-Instruct~\citep{dubey2024llama}, Mistral-7B-Instruct~\citep{jiang2023mistral}, and multi-step reasoning LLM DeepSeek-R1-Distill-Llama-8B~\citep{deepseekr1}.

\vspace{-15pt}
\paragraph{KV Cache Compression Methods}
To thoroughly investigate the potential impact on KV cache compression methods, we select the following methods: StreamingLLM~\cite{streamingllm}, SnapKV~\cite{snapkv}, H2O~\cite{h2o}, PyramidKV~\cite{pyramidkv}, PyramidInfer~\cite{yang2024pyramidinfer}, and ChunkKV~\cite{chunkkv}. 

\vspace{-15pt}
\paragraph{Hyperparameters}
The hyper-parameters for different observations are shown in \cref{tab:hyperparameters}. The temperature for the experiments are set to 0 for ensuring the deterministic results.

\vspace{-15pt}
\paragraph{Evaluation Environment}
We use the lm-evaluation-harness~\citep{eval-harness} library to load the models and evaluate the performance. The evaluation environment is a NVIDIA A40 GPU server.

\subsection{Attention Pattern Analysis on \bcmk{}}
\label{sec:attention_analysis}

\begin{figure}[t]
    \centering
    \begin{subfigure}[b]{1\linewidth}
        \includegraphics[width=1\linewidth]{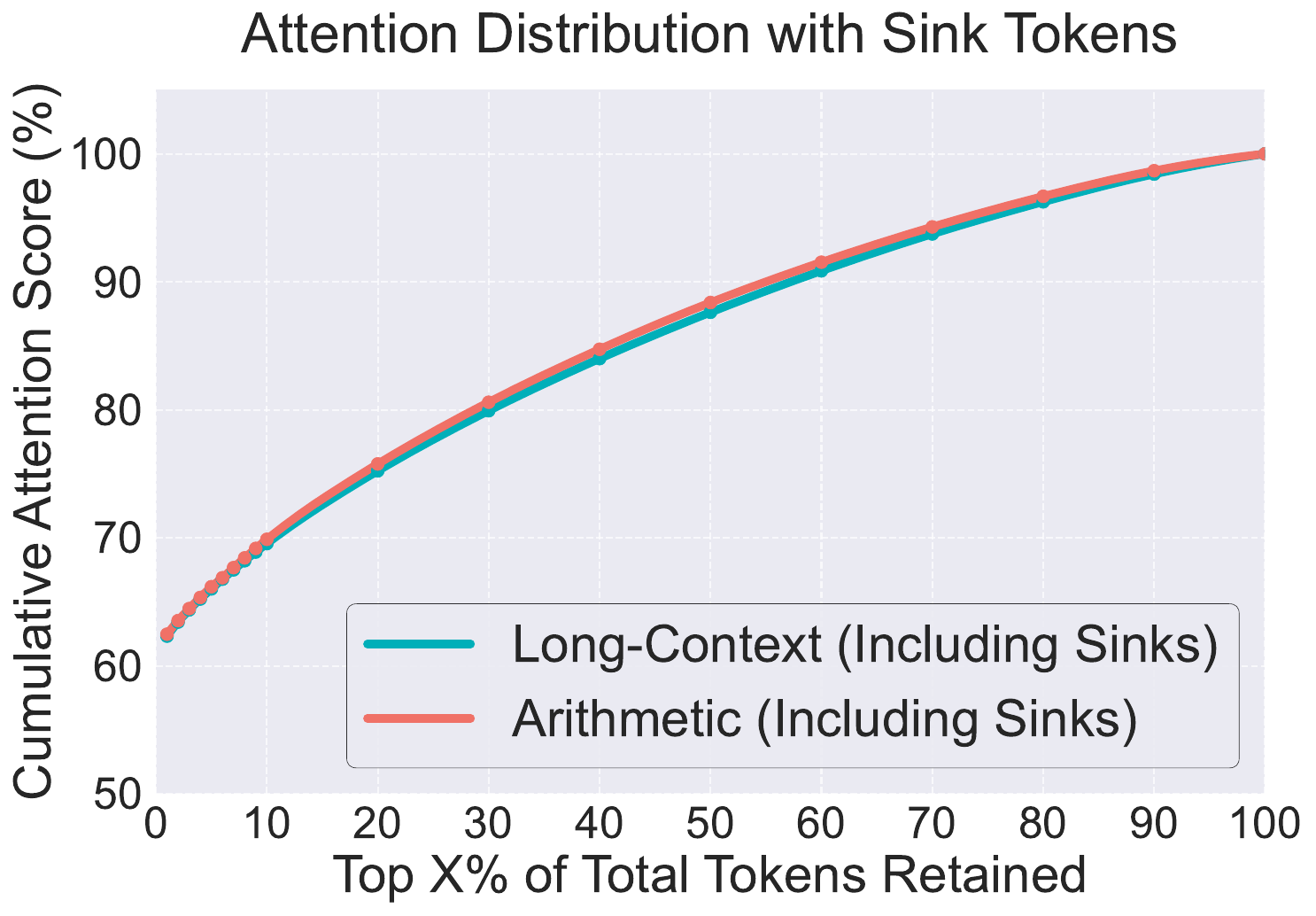}%
        \caption{Attention Distribution with Sink Tokens}
        \label{fig:subfig_with_sinks}%
    \end{subfigure}
    \vspace{-5pt}

    \begin{subfigure}[b]{1\linewidth}
            \vspace{-10pt}
        \includegraphics[width=1\linewidth]{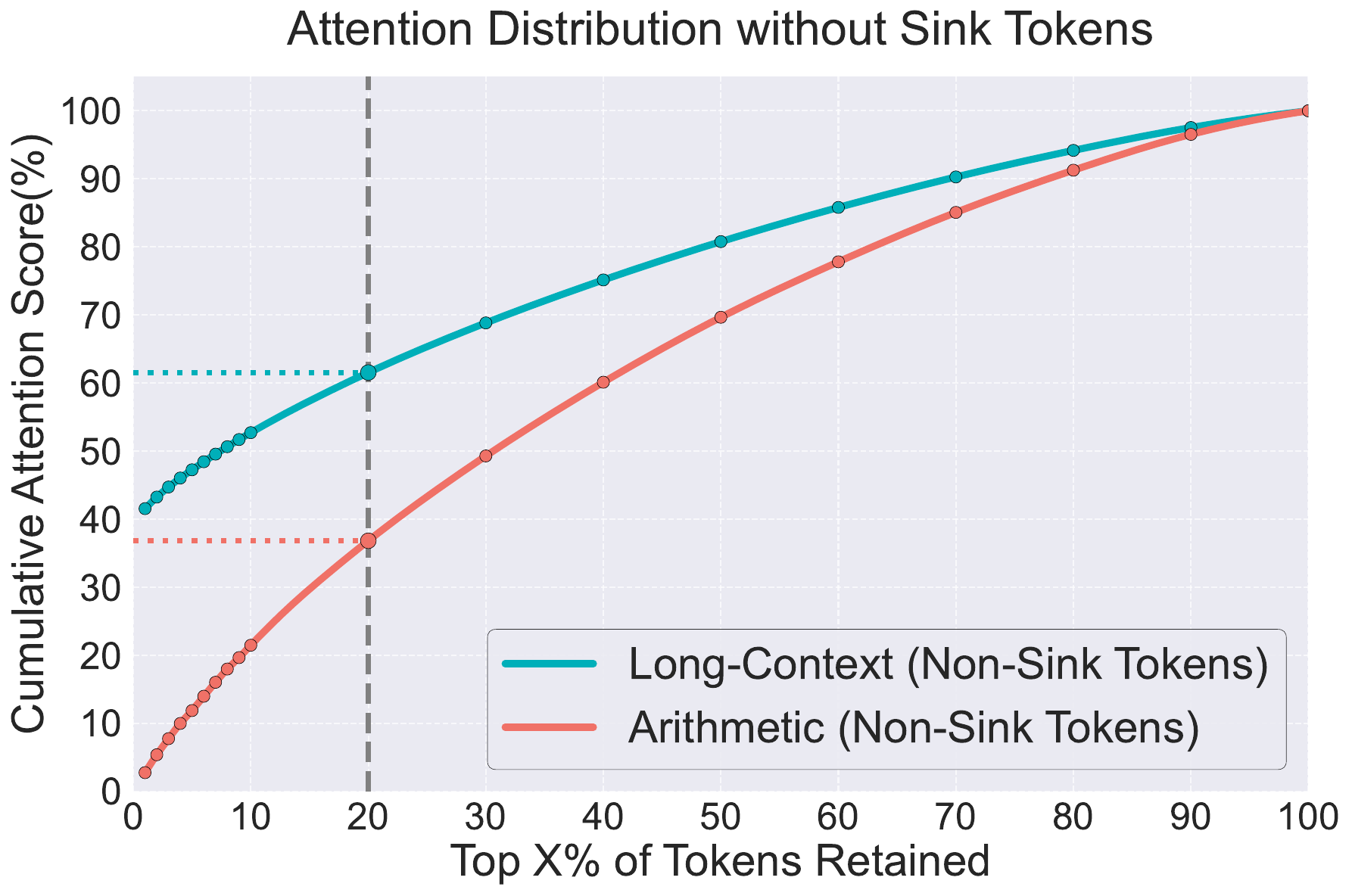}%
        \caption{Attention Distribution without Sink Tokens}
        \label{fig:subfig_without_sinks}%
    \end{subfigure}
    \vspace{-15pt}
    \caption{Cumulative attention score distribution for Long-Context and Arithmetic. (a) Overall distribution including initial sink tokens, showing high initial concentration. (b) Distribution without sink tokens (first 4 tokens removed), revealing that Arithmetic's non-sink attention is more diffuse compared to Long-Context's.}
    \vspace{-20pt} 
    \label{fig:attention_distributions_comparison} 
\end{figure}

To better understand the task-specific sensitivity to KV cache compression observed in our experiments, we analyze the attention patterns of LLMs on \bcmk{}. This analysis provides key insights into why different tasks respond differently to compression, from two perspectives: (1) task-specific attention patterns; and (2) attention distributions.

As shown in \cref{fig:main}, our benchmark reveals different attention behaviors in tasks: world knowledge and common sense reasoning tasks exhibit \textbf{universal attention distributions}, while arithmetic reasoning and safety tasks display more \textbf{specialized patterns}. In particular, arithmetic reasoning tasks show increased attention sparsity (i.e. focused attention on individual prompt examples), and safety tasks concentrate attention on system prompts, in contrast to the more uniform attention distribution seen in world knowledge and commonsense reasoning.

To further investigate the attention dynamics that might explain the task-specific sensitivities to KV cache compression, we analyzed the cumulative attention score distributions, as illustrated in \cref{fig:attention_distributions_comparison}. Figure~\ref{fig:attention_distributions_comparison}(a) depicts the overall attention distribution, which includes the initial sink tokens~\cite{streamingllm}. In this view, both long-context and arithmetic tasks demonstrate a very similar pattern: a steep initial rise where the top 1\% of tokens capture over 60\% of the total attention mass. This highlights the predominant role of sink tokens in attracting attention, regardless of the specific task.

However, a more distinct pattern emerges when these initial sink tokens (specifically, the first four tokens) are excluded from the analysis, as shown in \cref{fig:attention_distributions_comparison}(b). Within the remaining non-sink tokens, the attention distribution for arithmetic tasks becomes notably more diffuse, with a slower accumulation of attention mass. This divergence suggests that while sink tokens provide a common, strong attentional anchor, the subsequent distribution of attention across task-relevant (non-sink) tokens varies. The more diffuse attention in arithmetic's non-sink tokens implies a reliance on a broader set of contextual cues for its structured reasoning, potentially making it more vulnerable when compression begins to impact these non-sink tokens.

These detailed analyses of attention distributions (\cref{fig:main} and \cref{fig:attention_distributions_comparison}) reveal that LLMs engage different contextual information and attention strategies when performing long-context tasks versus tasks requiring fundamental abilities such as arithmetic reasoning. This highlights the necessity of evaluating KV cache compression beyond long-range dependencies to specifically assess its impact on diverse fundamental capabilities, owing to their distinct attentional mechanisms.

\begin{figure}[t]
    \centering

    \begin{subfigure}[b]{1\linewidth}
        \includegraphics[scale=0.29, trim=30 0 0 0]{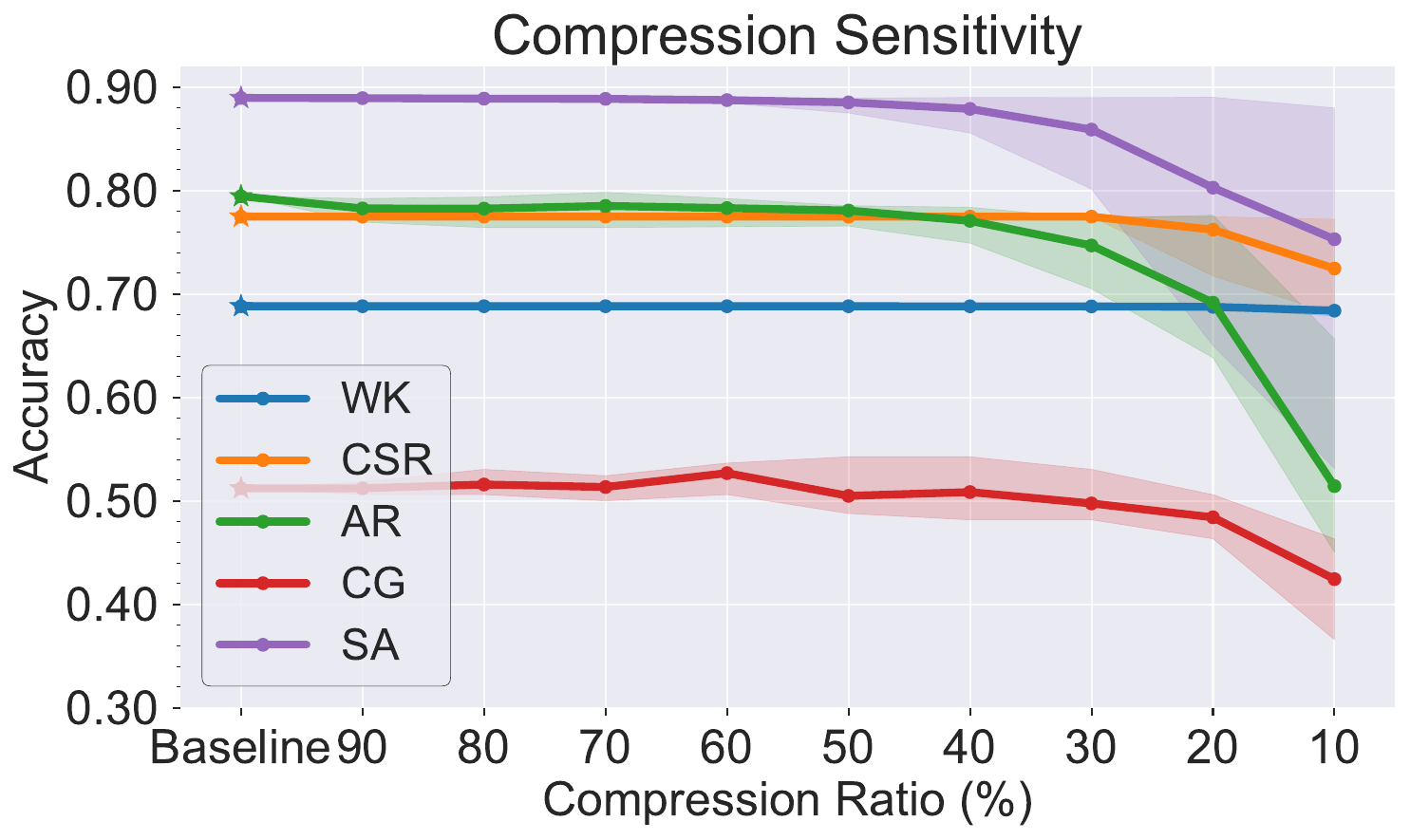}
        \caption{Sensitivity Analysis of Different Benchmark Categories to KV Cache Compression}
    \end{subfigure}
    \vspace{-10pt}

    \begin{subfigure}[b]{1\linewidth}
        \includegraphics[scale=0.29, trim=30 0 0 0]{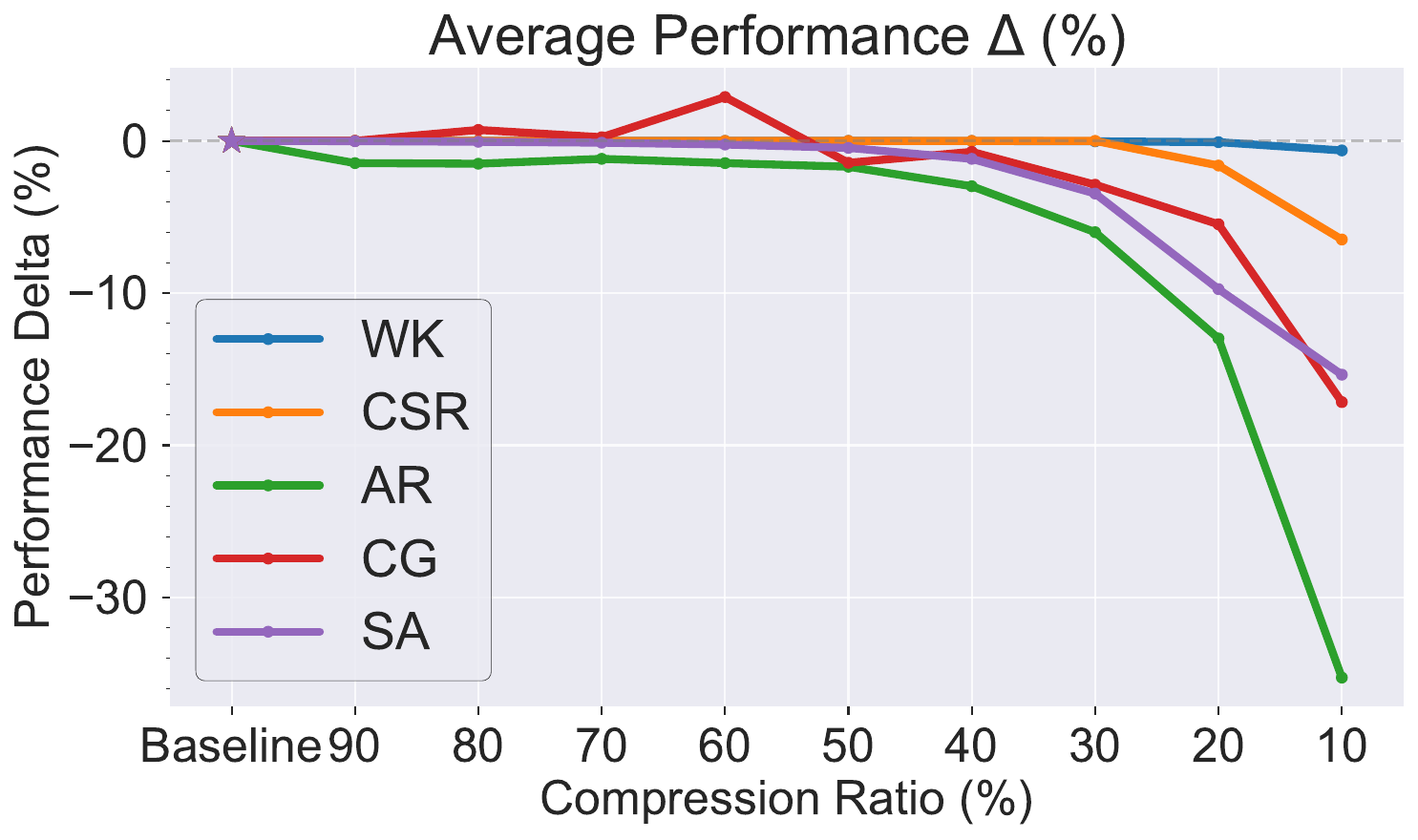}
        \caption{Performance Delta Lines with Baseline}
        
    \end{subfigure}
    \vspace{-15pt}

    \caption{Sensitivity Analysis of Different Benchmark Categories to KV Cache Compression. The benchmark categories include \textbf{WK} (World Knowledge), \textbf{CSR} (CommonSense Reasoning), \textbf{AR} (Arithmetic Reasoning), \textbf{CG} (Code Generation), and \textbf{SA} (Safety). The performance delta lines are calculated by \cref{eq:performance_change}.}

    \vspace{-15pt}
    \label{fig:benchmark_sensitivity}
\end{figure}

\subsection{Results and Analysis}
\label{sec:results_and_analysis}

In this section, we present the results and an analysis of the experiments. For detailed results, see \cref{sec:detail-results}.

\begin{figure*}[t]
    \centering
    \vspace{-5pt}
    \includegraphics[scale=0.4, trim=30 0 0 0]{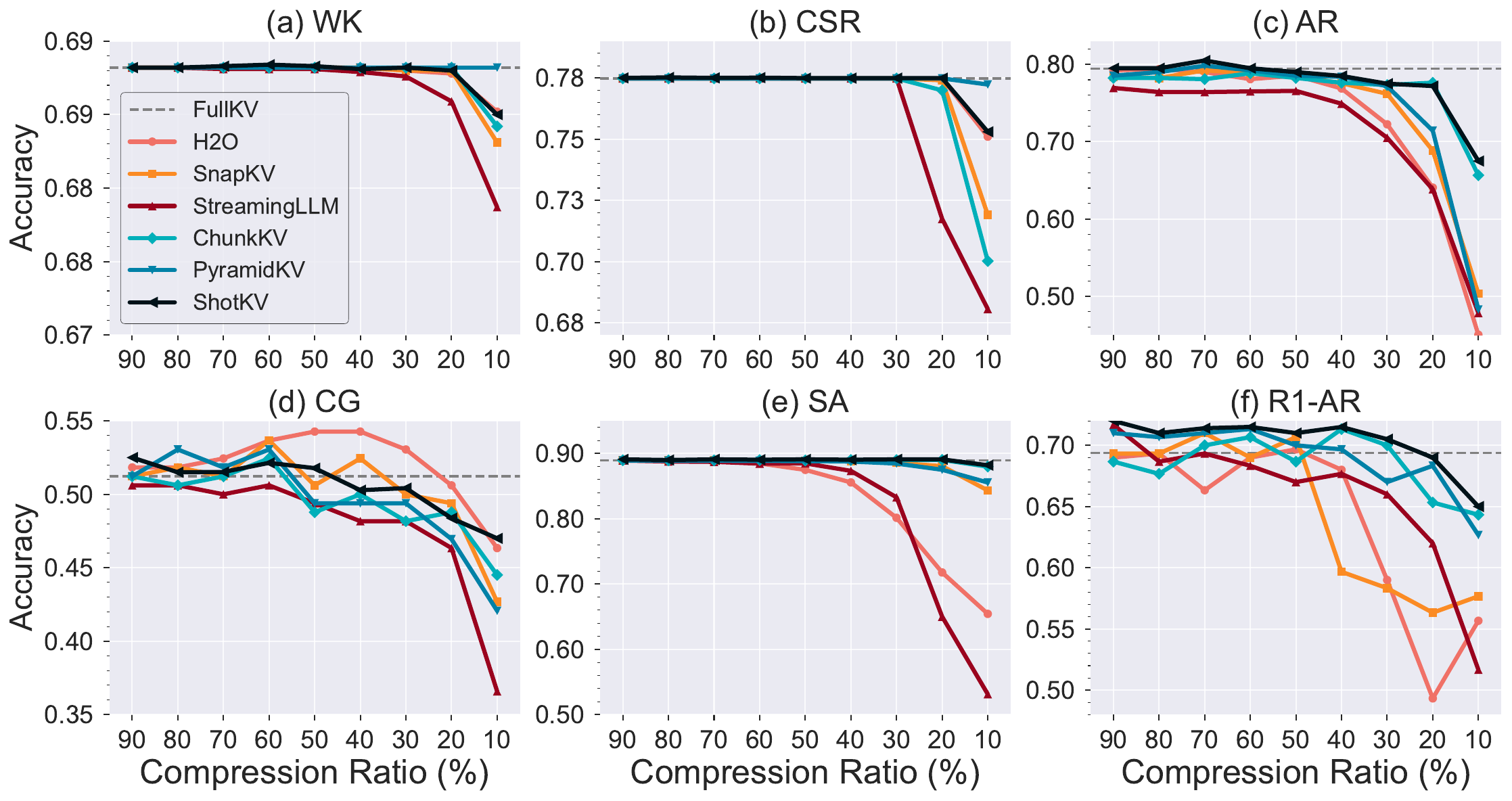}
    \vspace{-10pt}
    \caption{Performance Comparison of KV Cache Compression Methods on \bcmk{}. Results for R1-AR (f) were obtained using the DeepSeek-R1-Distill-Llama-8B model. \method{} is our proposed method; details can be found in Section~\ref{sec:shotkv}.}
    \label{fig:kv_cache_compression}
    \vspace{-15pt}
\end{figure*}

\vspace{-0.2cm}
\paragraph{\it{\textbf{Observation 1.}}} \textbf{Task-Dependent Degradation}: KV cache compression methods show task-dependent performance degradation, WK and CSR are more robust to KV cache compression.
As demonstrated in \cref{fig:benchmark_sensitivity}, all tasks maintain stable performance at compression ratios above 40\%, but exhibit distinct degradation patterns below this threshold. \textit{Arithmetic reasoning}, \textit{code generation}, and \textit{safety} tasks demonstrate the highest compression sensitivity, characterized by precipitous performance declines. \cref{fig:kv_cache_compression} illustrates the detailed performance impact of various KV cache compression methods across different tasks. This degradation is most pronounced in \textit{arithmetic reasoning} (\cref{fig:kv_cache_compression}(c)), where performance deteriorates significantly below the compression ratio of 20\%, with accuracy dropping from approximately 0.75 to below 0.5. Among the evaluated methods, ChunkKV~\cite{chunkkv} and PyramidKV~\cite{pyramidkv} consistently demonstrate superior stability in most tasks, while StreamingLLM~\cite{streamingllm} exhibits increased sensitivity to aggressive compression. Additionally, \textit{R1-Arithmetic reasoning} (\cref{fig:kv_cache_compression}(f)) indicates that reasoning LLMs demonstrate enhanced robustness to KV cache compression. Highlighting \textit{World Knowledge} and \textit{Common Sense Reasoning} as the most robust tasks, indicating that these tasks are less sensitive to KV cache compression.

\begin{figure}[t]
    \centering

    \begin{minipage}{0.49\textwidth}
        \centering
        \vspace{-5pt}
        \includegraphics[scale=0.29, trim=30 0 0 0]{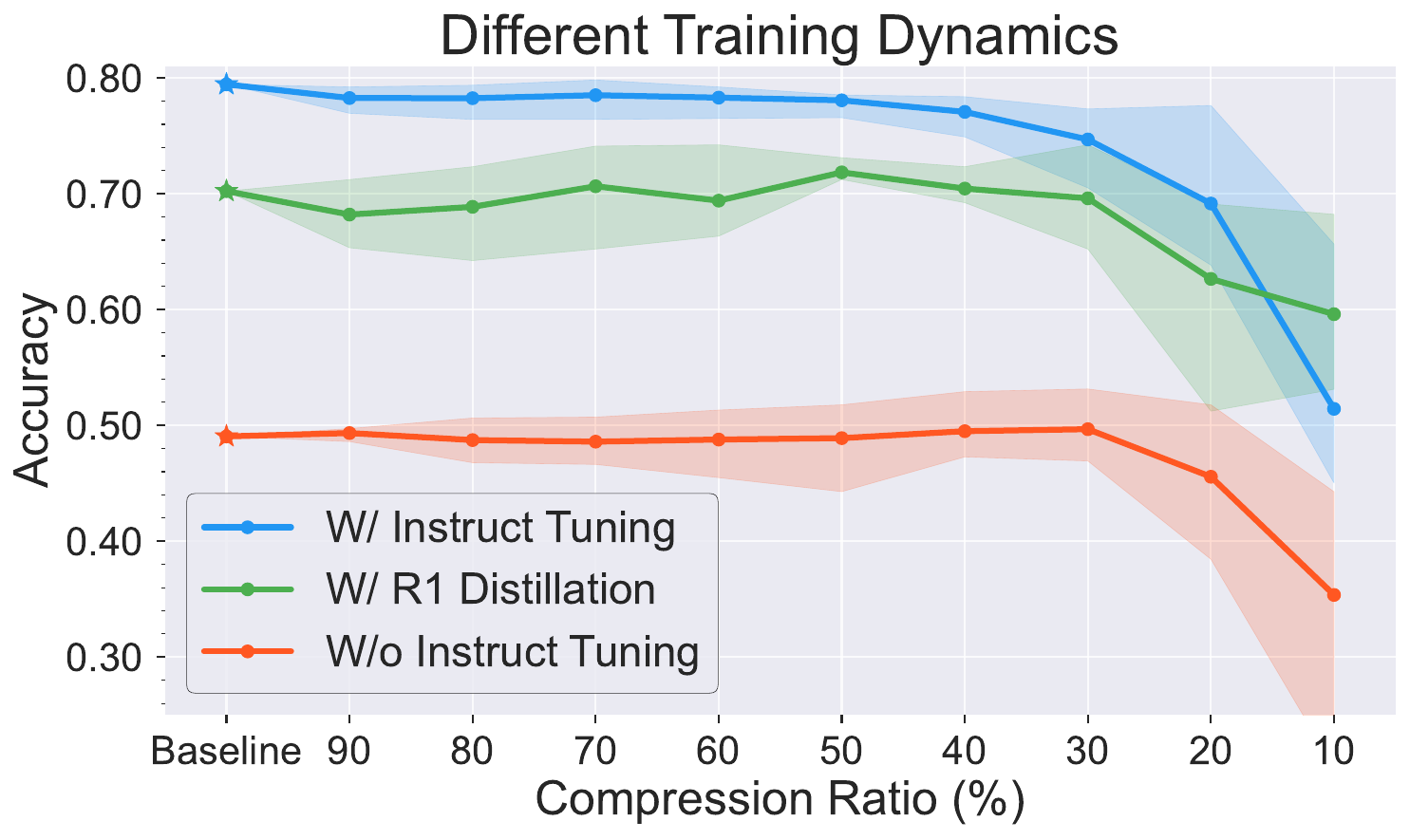}
        \vspace{-10pt}
        \caption{Performance Comparison of KV Cache Compression Methods on different training dynamics on \textit{Arithmetic Reasoning}}
        \label{fig:kv_cache_compression_instruct}
    \end{minipage}%
    \hfill
    \begin{minipage}{0.49\textwidth}
        \centering
        
        \includegraphics[scale=0.29, trim=30 0 0 0]{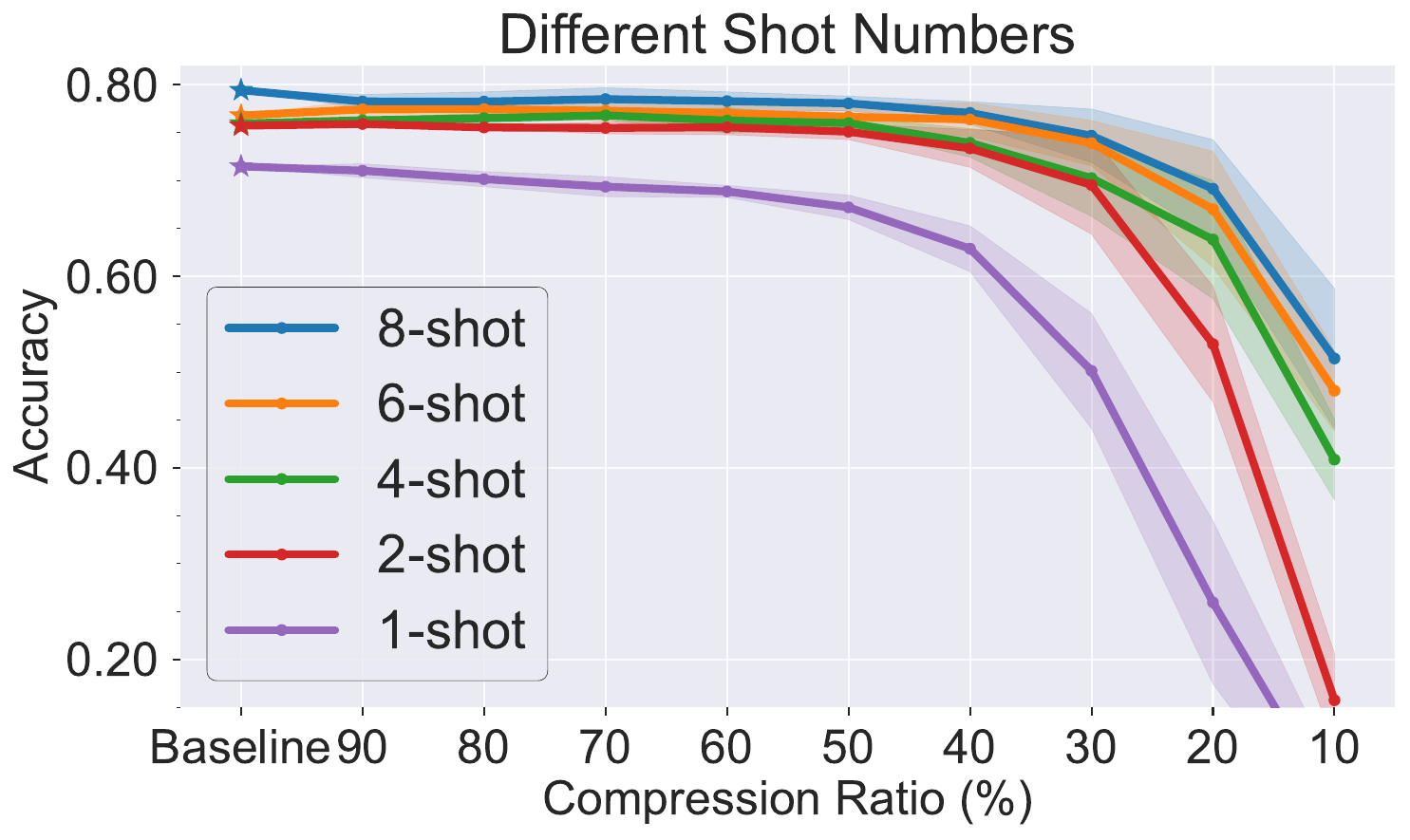}
        \vspace{-10pt}

        \caption{Average Performance Across Different Shot Numbers}
        \vspace{-10pt}

        \label{fig:shot_comparison_average}
    \end{minipage}
    \vspace{-10pt}
\end{figure}

\vspace{-0.8em}
\paragraph{\it{\textbf{Observation 2.}}} \textbf{Model-Type Robustness}: Multi-step reasoning LLMs are more robust to KV cache compression. \cref{fig:kv_cache_compression_instruct} presents a comparative analysis of LLaMA-3.1-8B across its base (w/o instruct tuned), instruct-tuned, and DeepSeek-R1 distilled variants, illustrating their averaged performance in five compression methods with confidence intervals. The R1 distilled model demonstrates superior stability, maintaining performance around 0.60 even at a compression ratio 10\%. The instruct-tuned model achieves a higher initial accuracy (0.8), but it exhibits heightened compression sensitivity, with performance deterioration beginning at 30\% compression ratio and declining sharply to approximately 0.5 at 10\% ratio. These findings suggest that while multi-step reasoning LLMs demonstrate enhanced robustness to KV cache compression, and instruct-tuning improves overall model performance, the latter may inadvertently increase model vulnerability to aggressive compression, particularly at compression ratios below 30\%.

\vspace{-0.8em}
\paragraph{\it{\textbf{Observation 3.}}} \textbf{Prompt Length Vulnerability}: Shorter prompts are more vulnerable to KV cache compression. 
As illustrated in \cref{fig:shot_comparison_average}, the effect of KV cache compression is markedly different with varying prompt lengths (shot numbers). Scenarios with fewer shots (for example, one-shot and two-shot) demonstrate heightened sensitivity to compression; their performance degrades more precipitously below a compression ratio of 30\% compared to scenarios with a greater number of shots (e.g., 4-8 shots). For example, in 1-shot settings, performance decreases from 0.5 to 0.05 as the compression ratio decreases from 30\% to 10\%. In contrast, 8-shot settings experience a less severe reduction, from 0.75 to 0.5, under the same compression conditions. This suggests that prompts with more shots, by virtue of containing more contextual examples, offer a richer set of reference points for the model. Consequently, the model's reliance on any single example being perfectly preserved in the compressed KV cache is reduced, leading to greater robustness against aggressive compression.

\begin{figure}[t]

    \centering
    \begin{subfigure}[b]{1\linewidth}
    \includegraphics[scale=0.29, trim=30 0 0 0]{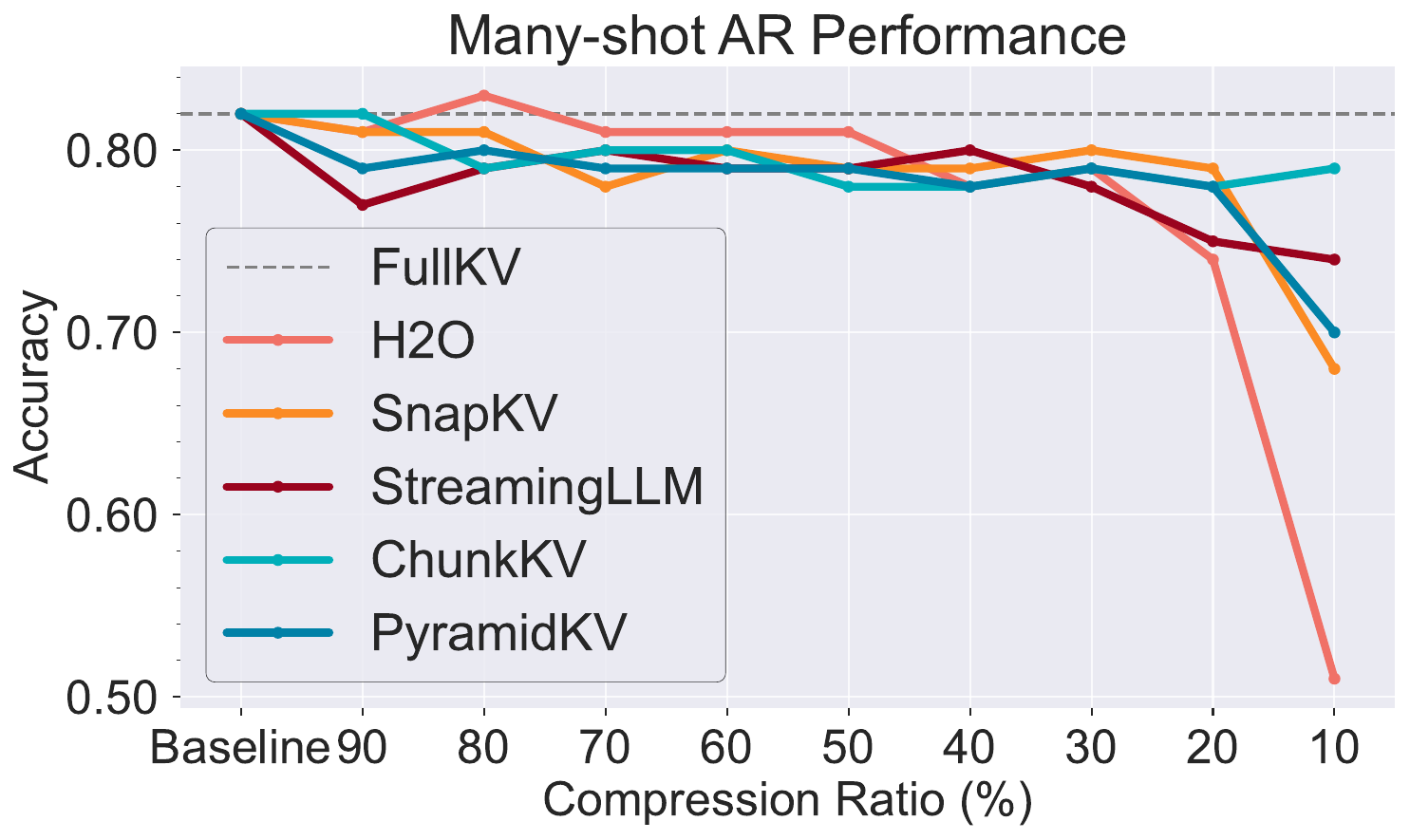}
    \caption{Many-shot \textit{Arithmetic Reasoning} on LLaMA3.1-8B-Instruct}
    \end{subfigure}
    \begin{subfigure}[b]{1\linewidth}
        \includegraphics[scale=0.29, trim=30 0 0 0]{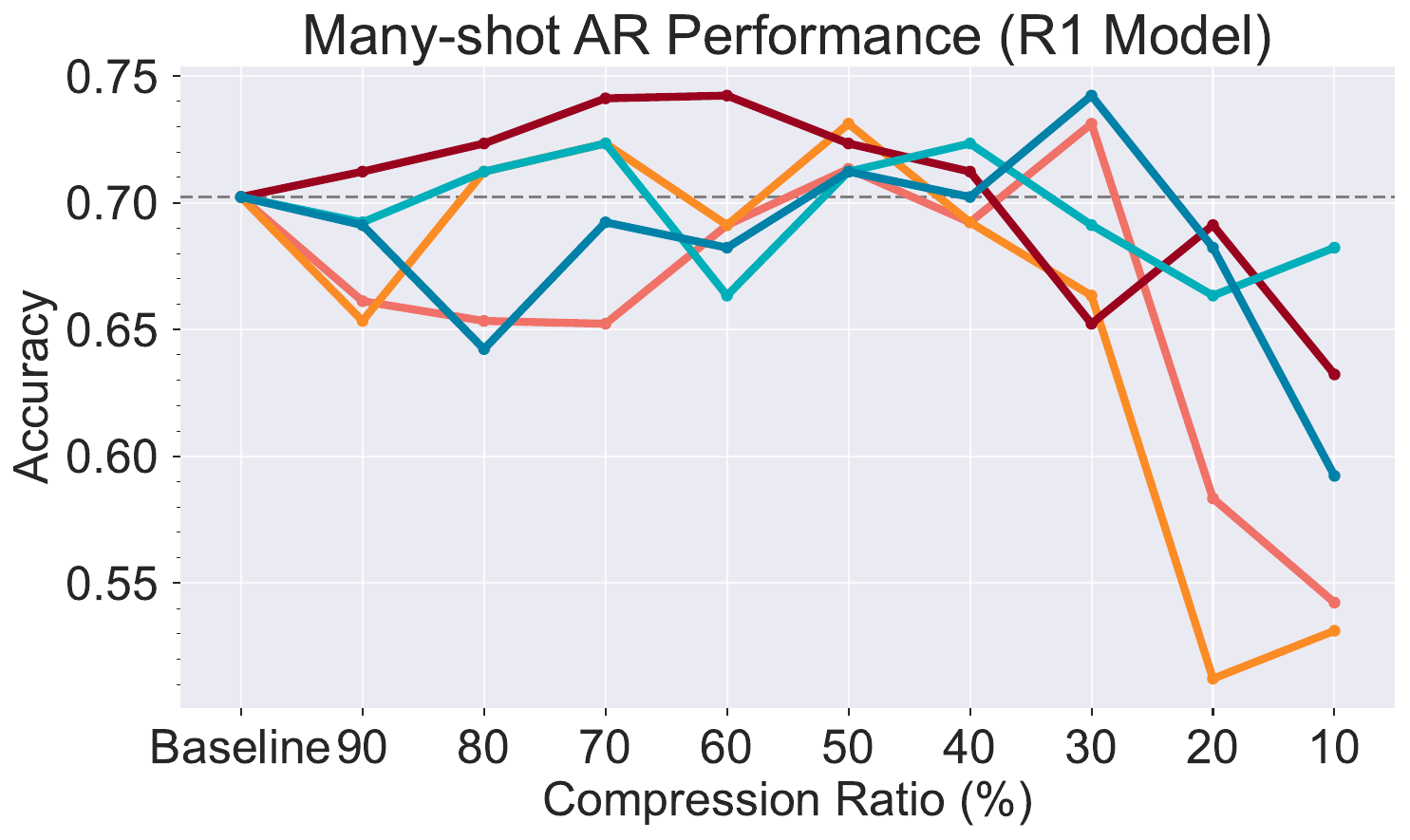}
        \caption{Many-shot \textit{Arithmetic Reasoning} on DeepSeek-R1-Distill-Llama-8B}
    \end{subfigure}
    \vspace{-15pt}
    \caption{Many-shot scenario on KV cache compression}
    \label{fig:many_shot_kv_cache_compression}
    \vspace{-20pt}
\end{figure}

\vspace{-8pt}
\paragraph{\it{\textbf{Observation 4.}}} \textbf{Chunk-Level Superiority}: Chunk-level compression is more effective for long-context structured reasoning tasks.
Inspired by \citet{agarwal2024many}, we consider many-shot in-context learning as a long-context reasoning task, which is more complex than existing long-context benchmarks, such as LongBench and NIAH. \cref{fig:many_shot_kv_cache_compression} shows the performance of KV cache compression methods on a 50-shot GSM8K task, where the prompt length exceeds 4K tokens. From the figure, we observe that ChunkKV~\cite{chunkkv} demonstrates the most stability when the compression ratio is below 10\% on both LLaMA-3.1-8B-Instruct and DeepSeek-R1-Distill-Llama-8B, indicating that in more complex long-context arithmetic reasoning tasks, chunk-level retention is more effective at preserving semantic information. This highlights the effectiveness of chunk-level compression for long-context structured reasoning tasks.

\begin{table}[t]
    \caption{Zero-shot vs Few-shot Performance Comparison}
    \centering
    \vspace{-5pt}
    \resizebox{1\linewidth}{!}{

    \begin{tabular}{l|cc|c}
    \toprule
    Benchmark & Zero-shot $\uparrow$ & CoT $\uparrow$ & Delta$\Delta$ \\
    \midrule
    \textit{Arithmetic Reasoning} & $29.04$ & $79.45$   & $+50.41$ \\
    \textit{World Knowledge} & $62.62$ & $68.82$  & $+6.20$ \\
    \bottomrule
    \end{tabular}
    }
    
    \vspace{-10pt}
    \label{tab:shot_comparison}
\end{table}

\vspace{-0.8em}
\paragraph{\it{\textbf{Observation 5.}}} \textbf{Prompt-Gain Sensitivity}: Tasks with higher performance gains from ICL and CoT are disproportionately more sensitive to KV cache compression. As shown in Table~\ref{tab:shot_comparison}, different tasks exhibit varying levels of performance improvement when moving from zero-shot to CoT prompting. \textit{Arithmetic reasoning} shows a dramatic improvement of 50.41\%, whereas \textit{World Knowledge} demonstrates a much more modest gain of 6.20\%.

From \cref{fig:benchmark_sensitivity}, we observe a strong correlation between prompt-based performance gains and compression sensitivity. Tasks that benefit significantly from CoT prompting, such as \textit{Arithmetic Reasoning}, are markedly more vulnerable to KV cache compression. This suggests a high dependency on the precise preservation of prompt information: when a model relies heavily on the reasoning patterns provided in the prompt to boost performance, any loss of this information via compression leads to a disproportionate drop in accuracy. Conversely, tasks like \textit{World Knowledge}, which show modest gains from CoT, exhibit greater resilience, likely because the model draws primarily from its internal parametric knowledge rather than the specific examples in the context. 

\begin{table}[t]
    \caption{KV cache compression methods' performance on \textit{LG-GSM8K}. Comp. \% indicates the compression ratio.}
    \centering
    \vspace{-5pt}
    \resizebox{1\linewidth}{!}{
    \begin{tabular}{l|c|cccc}
    \toprule
    \textbf{Method \& Comp. \%} & \textbf{100\%} & \textbf{40\%} & \textbf{35\%} & \textbf{30\%} & \textbf{25\%} \\
    \midrule
    \rowcolor{gray!30} FullKV & 46.00 & - & - & - & - \\
    \midrule
    StreamingLLM & - & 39.50 & 28.67 & 14.83 & 6.33 \\
    H2O & - & 32.66 & 25.17 & 19.83  & 14.83  \\
    PyramidInfer & - & 38.33 & 27.67 & 20.50 & 16.67 \\
    \rowcolor{red!20} \method{}(Ours) & - & \textbf{47.33} & \textbf{41.33} & \textbf{38.33} & \textbf{26.83}\\
    \bottomrule
    \end{tabular}
    }

    \vspace{-15pt}
    \label{tab:longgenbench_gsm8k_performance}
\end{table}

\vspace{-0.8em}
\paragraph{\it{\textbf{Observation 6.}}} \textbf{Long-Context Generation Sensitivity}: KV cache compression exhibits significant performance degradation in long-context generation tasks. As demonstrated in \cref{tab:longgenbench_gsm8k_performance}, our evaluation of three unified compression methods—StreamingLLM, H2O, and PyramidInfer—on \textit{LG-GSM8K} reveals substantial performance limitations. \textbf{We exclude context-compression methods like ChunkKV and SnapKV from this setting, as they primarily target static prefill compression and lack the dynamic token eviction mechanisms required to manage the rapidly growing cache during extended generation phases (e.g., 4k+ tokens).} In this arithmetic reasoning task with approximately 4k token generation duration, the evaluated unified compression methods show notable deterioration, with performance declining by more than 20\% at compression ratios below 30\%. The \method{} is our proposed method that aims to improve the performance of KV cache compression on Long-Context Generation tasks, details in \cref{sec:shotkv}.

\vspace{-5pt}
\paragraph{Summary and Design Implications}
Our comprehensive analysis reveals distinct challenges for KV cache compression: 
(1) The structural integrity of few-shot examples is crucial for reasoning tasks (\textit{Obs. 4}), suggesting that compression should operate at the \textit{semantic unit (shot) level} rather than the token level. 
(2) The conflicting requirements of preserving static prompt instructions (\textit{Obs. 1 \& 5}) versus dynamic generation contexts (\textit{Obs. 6}) indicate that a unified compression strategy is suboptimal. 
These insights motivate the design of \method{}, which explicitly separates prefill and decoding compression strategies and incorporates a shot-aware selection mechanism to preserve semantic coherence.

\vspace{-5pt}
\section{\method{}}
\label{sec:shotkv}

Based on the insights from \bcmk{}, we identify two critical design principles for effective KV cache compression on fundamental abilities:
\begin{enumerate}[leftmargin=*, topsep=0pt, itemsep=2pt, parsep=0pt]
    \item \textbf{Semantic Completeness (Addressing Obs. 4 \& 5):} Since reasoning tasks rely on coherent logical chains found in few-shot examples, compression algorithms must preserve these shots as atomic units. Token-level dropping (e.g., H2O) fragments these units, leading to performance collapse in arithmetic reasoning.
    \item \textbf{Phase Separation (Addressing Obs. 6):} The prefill phase (encoding static knowledge/rules) and decoding phase (generating dynamic steps) exhibit different sensitivities. A unified strategy fails to balance these distinct needs.
\end{enumerate}

\noindent We emphasize that \method{} is intentionally simple: it is a \emph{task-motivated combination} of two design choices derived from \bcmk{}---(i) shot-aware semantic-unit preservation in prefill, sharpening the chunk-level retention insight of ChunkKV with explicit few-shot boundaries, and (ii) explicit prefill/decoding separation, addressing the dynamic information needs identified in Observation~6. We do not claim novelty for generic attention-based ranking; rather, our contribution is to show that this targeted, lightweight combination is sufficient to recover the reasoning performance lost by token-level methods, thereby validating the central conclusion of \bcmk{}: preserving \emph{semantic units} matters more than fine-grained token selection under aggressive budgets.

\vspace{-5pt}
\subsection{Implementation}
The \textbf{\method{}} (Prefill-Decoding Separated \textbf{Shot}-aware \textbf{KV} Cache Compression), which separates the compression strategy for prefill and decoding phases. The key insight is that the prefill phase KV cache, which contains crucial prompt information, should be compressed once and remain fixed, while the decoding phase KV cache can be dynamically compressed with different strategies.

Given a prompt with $n$ shots and tokens generated, we define:
\vspace{-0.3cm}
\begin{equation}
    KV_{\text{total}} = KV_{\text{prefill}} \cup KV_{\text{decoding}}
\end{equation}
\vspace{-0.6cm}

For the prefill phase, guided by \textbf{Observation 4} (Chunk-Level Superiority) which highlights the necessity of maintaining semantic chunks, we compute shot importance and preserve complete shot examples:
\vspace{-0.5cm}
\begin{equation}
    \text{Score}_{\text{prefill}}^l(s_i) = \frac{1}{k_i} \sum_{t \in s_i} \sum_{h=1}^H \alpha_{t,h}^l
\end{equation}
\vspace{-0.5cm}

where $s_i$ represents the $i$-th shot example containing $k_i$ tokens. The term $\alpha_{t,h}^l$ denotes the accumulated attention scores of token $t$ within shot $s_i$ in attention head $h$ at transformer layer $l$. Following standard practices in KV compression~\citep{h2o, snapkv}, we treat the accumulated attention weight as a \emph{practical heuristic} for ranking semantic units rather than a principled measure of semantic importance: it is empirically effective at preserving coherent few-shot structure under a budget, and we report sensitivity to shot ordering and exemplar choice in the appendix. The compression is performed independently for each layer, allowing different layers to retain different subsets of shots based on their specific attention patterns. Once the prefill phase KV cache is compressed, it remains fixed throughout the generation process.

Given a prefill compression ratio $r_p$, we prioritize shots with higher scores while ensuring that the total number of preserved tokens does not exceed the KV cache limit. Specifically, the shots are ranked by their scores and selected in descending order until they reach the compression budget $r_p \times |KV_{\text{prefill}}|$. This shot-level selection strategy helps to maintain the semantic coherence of important examples while adhering to memory constraints.

\vspace{-0.5cm}
\begin{equation}
    KV^C_{\text{prefill}, l} = \text{Compress}(\{s_i | s_i \in S^*_{\text{preserved}, l}\})
\end{equation}
\begin{equation}
    \label{eq:prefill_select}
    S^*_{\text{preserved}, l} = \operatorname{TopK}\left(\{s_1, \dots, s_n\}, \text{Score}_{\text{prefill}}^l, N_{\text{budget}}\right)
\end{equation}
\vspace{-0.5cm}

\vspace{-0.5cm}
\begin{equation}
    \text{subject to:} \quad \sum_{s_i \in S} k_i \leq r_p \times |KV_{\text{prefill}}|
\end{equation}
\vspace{-0.5cm}

Here, $KV^C_{\text{prefill}, l}$ represents the compressed KV cache for prefilling at layer $l$, and $S^*_{\text{preserved}, l}$ represents the selected subset of shots to be preserved for that layer. where $\operatorname{TopK}$ selects the subset of shots with the highest importance scores such that the total token count fits within the budget defined by $N_{\text{budget}} \approx r_p \times |KV_{\text{prefill}}|$. The second equation enforces the memory constraint: the total number of tokens ($k_i$) in the selected shots must not exceed the allocated budget, which is determined by the prefill compression ratio $r_p$ multiplied by the original KV cache size.

For the decoding phase, addressing \textbf{Observation 6} (Long-Context Generation Sensitivity), we employ a dynamic token-level compression to manage the evolving context independently for each layer:
\vspace{-0.3cm}
\begin{equation}
    \text{Score}_{\text{decoding}}^l(t) = \sum_{h=1}^H \alpha_{t,h}^l
\end{equation}
\vspace{-0.5cm}

Here, for a previously generated token $t$, $\alpha_{t,h}^l$ is similarly defined as the attention weight assigned by the query vector of the current token being generated to the key vector of token $t$, within head $h$ at layer $l$. Thus, $\text{Score}_{\text{decoding}}^l(t)$ represents the total attention received by token $t$ at layer $l$ from the current generation step.

Given a decoding compression ratio $r_d$, we select the tokens with the highest scores to preserve. The compressed decoding KV cache $KV^C_{\text{decoding}, l}$ retains only the top-$k$ tokens where $k = r_d \times |KV_{\text{decoding}}|$, effectively maintaining the most influential context tokens while reducing memory usage:
\vspace{-0.3cm}
\begin{equation}
    \begin{split}
        KV^C_{\text{decoding}, l} = \text{TopK}(&KV_{\text{decoding}}, \text{Score}_{\text{decoding}}^l, \\
        &k = r_d \times |KV_{\text{decoding}}|)
    \end{split}
\end{equation}
\vspace{-0.5cm}

Finally, we combine compressed prefill and decoding KV caches to form the total compressed KV cache for each layer:

\vspace{-0.5cm}
\begin{equation}
    KV_{\text{total}, l} = KV^C_{\text{prefill}, l} \cup KV^C_{\text{decoding}, l}
\end{equation}
\vspace{-0.5cm}

\begin{table*}[t]
    \caption{LLaMA-3-8B-Instruct on HotpotQA under 10\% and 30\% compression. \method{} adapts to this non-ICL document-QA setting by treating each sentence as a semantic unit.}
    \centering
    \vspace{-5pt}

    \resizebox{1\linewidth}{!}{
    \begin{tabular}{l|c|cccccc}
    \toprule
    \textbf{Comp.\ \%} & FullKV & StreamingLLM & H2O & SnapKV & PyramidKV & ChunkKV & \method{} (Ours) \\ \midrule
    10\% & \multirow{2}{*}{45.55} & 40.27 & 40.84 & 43.36 & 43.80 & 43.27 & 43.60 \\
    30\% &       & 42.18 & 42.75 & 44.12 & 44.56 & 44.05 & 44.38 \\
    \bottomrule
    \end{tabular}
    }

    \vspace{-15pt}
    \label{tab:hotpotqa_shotkv}

\end{table*}

\vspace{-0.3cm}
\subsection{Empirical Results}
In this section, we evaluate the effectiveness of \method{} on the \bcmk{}. As illustrated in \cref{fig:kv_cache_compression}, \method{} consistently achieves top-tier performance across various tasks, demonstrating its robustness as a compression strategy. To provide a more granular analysis, we focus on two particularly challenging scenarios identified in our benchmark analysis: many-shot \textit{Arithmetic Reasoning}, where preserving reasoning chains is critical, and \textit{Long-Context Generation}, which demands sustained coherence over extended outputs. We additionally report a non-ICL generalization study on HotpotQA.

\vspace{-0.3cm}
\paragraph{Baseline.} For \textit{LG-GSM8K} evaluation, we employ three state-of-the-art unified compression methods as baselines: StreamingLLM~\cite{streamingllm}, H2O~\cite{h2o}, and PyramidInfer~\cite{yang2024pyramidinfer}. We conduct experiments using LLaMA-3-8B-Instruct~\cite{dubey2024llama} on the \textit{LG-GSM8K} benchmark~\cite{longgenbench}, maintaining consistent parameters with Observation 6 ($K = 35$, $T = 20$). For many-shot \textit{Arithmetic Reasoning} experiments, we follow the configuration detailed in Observation 4.

\begin{table}[t]
    \caption{KV cache compression methods' performance on Many-shot \textit{Arithmetic Reasoning}}
    \vspace{-5pt}

    \centering
 
    \resizebox{1\linewidth}{!}{
    \begin{tabular}{l|c|cccc}
    \toprule
    Method \& Comp. \% & 100\% & 40\% & 30\% & 20\% & 10\% \\
    \midrule
    \rowcolor{gray!30} FullKV & 82.35 & - & - & - & - \\
    \midrule
    StreamingLLM & - & 80.37 & 78.35 & 75.37 & 74.32 \\
    H2O & - & 78.32 & 79.32 & 74.28 & 51.27 \\
    PyramidKV & - & 78.34 & 79.34 & 78.32 & 70.37 \\
    SnapKV & - & 79.35 & 80.38 & 79.34 & 68.27 \\
    ChunkKV & - & 78.32 & 79.32 & 78.35 & 79.32 \\
    Random Shot & - & 72.50 & 68.23 & 62.50 & 51.34 \\
    \rowcolor{red!20} \method{}(Ours) & - & \textbf{81.07} & \textbf{80.82} & \textbf{80.57} & \textbf{80.37} \\

    \bottomrule
    \end{tabular}
    }

    \vspace{-10pt}

    \label{tab:kv-compression-many-gsm8k}

\end{table}

\vspace{-0.8em}

\paragraph{Main Results.} As shown in Table~\ref{tab:longgenbench_gsm8k_performance}, \method{} achieves 47.33\% accuracy on \textit{LG-GSM8K} at 40\% compression, surpassing the FullKV baseline. Furthermore, on many-shot \textit{Arithmetic Reasoning} (Table~\ref{tab:kv-compression-many-gsm8k}), \method{} maintains \textbf{80.37\%} accuracy even at a 10\% ratio. \textbf{This significantly outperforms a \textit{Random Shot} baseline (51.34\%), where the $>$29\% gap confirms the efficacy of our attention-based semantic unit preservation.} 



This superior performance can be attributed to two key design choices: (1) the preservation of complete shot examples ensures the semantic coherence necessary for mathematical reasoning, and (2) the separation of prefill and decoding phases allows for flexible, task-appropriate token retention. These results suggest that our shot-aware strategy is particularly effective for tasks requiring complex reasoning chains. Appendix~\ref{appendix:lggsm8k} provides more detailed ablation analysis on \textit{LG-GSM8K}.


\vspace{-0.8em}
\paragraph{Latency and Throughput}
We further compare the inference efficiency of ShotKV and the FullKV baseline in terms of latency and throughput under different input and output sequence lengths. These results demonstrate that \method{} not only maintains model performance under aggressive KV cache compression, but also brings tangible efficiency benefits for long-context inference.

\begin{table}[t]
    \caption{Latency and throughput comparison among \method{}, FullKV, and representative compression baselines under different input-output configurations. Tested on the A40 server with batch size 1. Percentages show the relative gain over FullKV.}
    \vspace{-5pt}

    \centering
    \resizebox{1\linewidth}{!}{
    \begin{tabular}{l|cc|cc}
    \toprule
    \multirow{2}{*}{\textbf{Method}} & \multicolumn{2}{c|}{\textbf{Sequence Length}} & \multicolumn{2}{c}{\textbf{Performance Metrics}} \\
    \cmidrule{2-5}
    & \textbf{Input} & \textbf{Output} & \textbf{Latency(s)} $\downarrow$ & \textbf{Throughput(T/S)} $\uparrow$ \\
    \midrule
    FullKV   & 4096 & 4096 & 175.50          & 37.73          \\
    ChunkKV  & 4096 & 4096 & 160.32 (8.6\%)  & 41.30 (9.5\%)  \\
    SnapKV   & 4096 & 4096 & 163.45 (6.9\%)  & 40.51 (7.4\%)  \\
    \method{} & 4096 & 4096 & 162.85 (7.2\%)  & 41.12 (9.0\%)  \\
    \midrule
    FullKV   & 8192 & 4096 & 183.42          & 55.93          \\
    \method{} & 8192 & 4096 & 162.78 (11.3\%) & 63.24 (13.1\%) \\
    \bottomrule
    \end{tabular}
    }
   
    \vspace{-15pt}

    \label{tab:efficiency}
\end{table}

\vspace{-5pt}
\paragraph{Generalization to Non-ICL Tasks (HotpotQA).}
\label{appendix:hotpotqa}

For a document QA setting without few-shot ICL, we adapt \method{} by treating each sentence as a coherent semantic unit (analogous to a \emph{shot}). Even under an aggressive 10\% compression ratio on LLaMA-3-8B-Instruct, \method{} remains competitive with the best-performing method, as shown in Table~\ref{tab:hotpotqa_shotkv}.

\vspace{-10pt}
\section{Limitations}
\label{sec:limitations}
\paragraph{Deployment scope.} Our method requires direct access to the KV cache, so it targets self-hosted or open-weight deployments (e.g., LLaMA, Mistral, DeepSeek, Qwen) rather than closed API-only systems. This setting matches all KV-compression baselines we compare against and is the regime where KV-level intervention is actionable.

\vspace{-10pt}
\paragraph{Scoring is a practical heuristic.} The attention-derived score in \method{} is an effective operational signal for ranking semantic units under a budget; it is not a principled semantic-importance oracle. 

\vspace{-10pt}
\paragraph{Orthogonal compression axes.} Our scope is token-level retention; complementary axes such as KV quantization~\citep{li2025antkv} and broader capability-preservation analyses~\citep{tang2025lottery} can be combined with \method{} but are out of scope here.

\vspace{-10pt}
\paragraph{Prompt-structure dependence.} \method{} is most effective when the prompt contains recoverable semantic units (few-shot exemplars or sentence-level chunks). The non-ICL HotpotQA experiment (\cref{tab:hotpotqa_shotkv}) shows the design transfers to document QA via sentence-as-shot adaptation, but extension to fully zero-shot, unstructured long-document settings (e.g., book-scale summarization) remains open. An extended discussion of these limitations is provided in \cref{app:limitations}.

\vspace{-5pt}
\section{Conclusion}
This paper presents \bcmk{}, a benchmark for systematically evaluating the effects of KV cache compression on various fundamental LLM capabilities. Our findings reveal that performance degradation is highly task dependent, with \textbf{High-Density Reasoning} (e.g., arithmetic reasoning) and long-context generation being particularly sensitive (\textit{Task-Dependent Degradation} and \textit{Long-Context Generation Sensitivity}). We also highlight that compression sensitivity is influenced by a confluence of factors, including inherent model characteristics such as training dynamics—\textbf{where we identify the unique robustness of reasoning models like DeepSeek-R1} (\textit{Model-Type Robustness})—prompt-level attributes like length (\textit{Prompt Length Vulnerability}), and the reliance on in-context examples (\textit{Prompt-Gain Sensitivity}). Crucially, we demonstrate the importance of preserving the semantic integrity of prompt components, especially at a chunk or shot level, for complex tasks where current token-level methods often struggle (\textit{Chunk-Level Superiority}).

Based on these insights, we developed \method{}, an insight-driven strategy designed to validate our findings regarding the necessity of semantic preservation. By distinctively managing prefill and decoding phases and prioritizing shot-level \textbf{Semantic Units}, \method{} effectively mitigates information loss in sensitive tasks. The empirical success of \method{}—\textbf{achieving significant accuracy recovery while delivering tangible latency reductions}—confirms the critical role of semantic integrity in reasoning tasks and underscores the potential for more nuanced, task-aware compression strategies.

\section*{Impact Statement}
\label{app:impact_statement}

This work advances the field of efficient large language model deployment through systematic analysis and improvement of KV cache compression techniques. Our research has several potential societal impacts:

First, by enabling more efficient memory usage in LLMs while maintaining performance, our work contributes to reducing the computational resources and energy consumption required for AI deployment. This has positive environmental implications and makes AI technology more accessible to researchers and organizations with limited computing resources.
 
Second, our proposed \method{} method specifically improves performance on long-context arithmetic reasoning tasks, which could enhance the practical applications of LLMs in education, scientific computing, and other fields requiring complex mathematical reasoning. This could lead to more reliable AI-assisted learning and problem-solving tools.

However, we acknowledge that making LLMs more efficient could accelerate their widespread adoption, potentially raising concerns about AI's impact on employment and privacy. While our work focuses on technical improvements, we encourage the research community to carefully consider these broader implications when deploying such technologies.

We believe the benefits of more efficient and capable AI systems outweigh potential risks, particularly as our work promotes more sustainable and accessible AI development. Nevertheless, we emphasize the importance of responsible deployment and continued ethical consideration in the application of these technologies.

\section*{Acknowledgments}
This work was partially supported by the National Natural Science Foundation of China under Grant No.~62272122, and Hong Kong CRF grants under Grant No.~C7004-22G and C6015-23G. This work was also supported by the National Natural Science Foundation of China (Grant No.~62506318); Guangdong Provincial Department of Education Project (Grant No.~2024KQNCX028); CAAI-Ant Group Research Fund; Scientific Research Projects for the Higher-educational Institutions (Grant No.~2024312096), Education Bureau of Guangzhou Municipality; and Guangzhou-HKUST(GZ) Joint Funding Program (Grant No.~2025A03J3957), Education Bureau of Guangzhou Municipality.

\bibliography{custom}

@misc{huang2026mdnparallelizingstepwisemomentum,
      title={MDN: Parallelizing Stepwise Momentum for Delta Linear Attention}, 
      author={Yulong Huang and Xiang Liu and Hongxiang Huang and Xiaopeng Lin and Zunchang Liu and Xiaowen Chu and Zeke Xie and Bojun Cheng},
      year={2026},
      eprint={2605.05838},
      archivePrefix={arXiv},
      primaryClass={cs.LG},
      url={https://arxiv.org/abs/2605.05838}, 
}

@article{tang2026your,
  title={Is Your LLM-as-a-Recommender Agent Trustable? LLMs' Recommendation is Easily Hacked by Biases (Preferences)},
  author={Tang, Zichen and Zhang, Zirui and Wang, Qian and Tang, Zhenheng and Li, Bo and Chu, Xiaowen},
  journal={arXiv preprint arXiv:2603.17417},
  year={2026}
}

@article{tang2026memory-survey,
	doi = {10.20944/preprints202603.0359.v2},
	url = {https://doi.org/10.20944/preprints202603.0359.v2},
	year = 2026,
	month = {March},
	publisher = {Preprints},
	author = {Zhenheng Tang and Xin He and Tiancheng Zhao and Fanjunduo Wei and Xiang Liu and Peijie Dong and Qian Wang and Qi Li and Huacan Wang and Ronghao Chen and Sen Hu and Weidong Guo and Yu Xu and Haolan Chen and Kunfeng Lai and Kaiyong Zhao and Keyan Ding and Ivor W. Tsang and Yew-Soon Ong and Bo Li and Xiaowen Chu},
	title = {LLM Agent Memory: A Survey from a Unified Representation–Management Perspective},
	journal = {Preprints}
}

@article{yu2026rethinking,
  title={Rethinking Deep Research from the Perspective of Web Content Distribution Matching},
  author={Yu, Zixuan and Tang, Zhenheng and Liu, Tongliang and Zhang, Chengqi and Chu, Xiaowen and Han, Bo},
  journal={arXiv preprint arXiv:2603.07241},
  year={2026}
}

@inproceedings{
tang2026ghost,
title={Ghost in the Cloud: Your Geo-Distributed Large Language Models Training is Easily Manipulated},
author={Zichen TANG and Zhenheng Tang and Gaoning Pan and Buhua Liu and Xin He and Kunfeng Lai and Xiaowen Chu and Bo Li},
booktitle={The Fourteenth International Conference on Learning Representations},
year={2026},
url={https://openreview.net/forum?id=FwnmQnVc7g}
}

@inproceedings{zhu2025oraclekv,
 author = {Zhu, Yuanbing and Tang, Zhenheng and Liu, Xiang and Li, Ang and Li, Bo and Chu, Xiaowen and Han, Bo},
 booktitle = {ICML 2025 Workshop on Long-Context Foundation Models},
 title = {OracleKV: Oracle Guidance for Question-Independent KV Cache Compression},
 year = {2025}
}

@inproceedings{liu2025flowkv,
  title={{FlowKV}: Enhancing Multi-Turn Conversational Coherence in {LLM}s via Isolated Key-Value Cache Management},
  author={Liu, Xiang and Chen, Hong and Hu, Xuming and Chu, Xiaowen},
  booktitle={NeurIPS Workshop on Multi-Turn Interactions in Large Language Models},
  year={2025},
  url={https://arxiv.org/abs/2505.15347}
}

@article{kvpress,
 author = {Devoto, Alessio and Jeblick, Maximilian and J{\'e}gou, Simon},
 journal = {ArXiv preprint},
 title = {Expected Attention: KV Cache Compression by Estimating Attention from Future Queries Distribution},
 url = {https://arxiv.org/abs/2510.00636},
 volume = {abs/2510.00636},
 year = {2025}
}

@article{kim2025kvzip,
 author = {Kim, Jang-Hyun and Kim, Jinuk and Kwon, Sangwoo and Lee, Jae W and Yun, Sangdoo and Song, Hyun Oh},
 journal = {ArXiv preprint},
 title = {KVzip: Query-Agnostic KV Cache Compression with Context Reconstruction},
 url = {https://arxiv.org/abs/2505.23416},
 volume = {abs/2505.23416},
 year = {2025}
}

@inproceedings{scope,
  title={{SCOPE}: Optimizing Key-Value Cache Compression in Long-context Generation},
  author={Wu, Jialong and Wang, Zhenglin and Zhang, Linhai and Lai, Yilong and He, Yulan and Zhou, Deyu},
  booktitle={Proceedings of the 63rd Annual Meeting of the Association for Computational Linguistics (Volume 1: Long Papers)},
  pages={10775--10790},
  year={2025},
  address={Vienna, Austria},
  publisher={Association for Computational Linguistics}
}

@inproceedings{chunkkv,
  title={{ChunkKV}: Semantic-Preserving {KV} Cache Compression for Efficient Long-Context {LLM} Inference},
  author={Liu, Xiang and Tang, Zhenheng and Dong, Peijie and Li, Zeyu and Liu, Yue and Li, Bo and Hu, Xuming and Chu, Xiaowen},
  booktitle={Advances in Neural Information Processing Systems},
  volume={38},
  year={2025},
  url={https://openreview.net/forum?id=8sglLco8Ti}
}

@inproceedings{fei-etal-2024-extending,
 address = {Bangkok, Thailand and virtual meeting},
 author = {Fei, Weizhi  and
Niu, Xueyan  and
Zhou, Pingyi  and
Hou, Lu  and
Bai, Bo  and
Deng, Lei  and
Han, Wei},
 booktitle = {Findings of the Association for Computational Linguistics ACL 2024},
 doi = {10.18653/v1/2024.findings-acl.306},
 editor = {Ku, Lun-Wei  and
Martins, Andre  and
Srikumar, Vivek},
 pages = {5169--5181},
 publisher = {Association for Computational Linguistics},
 title = {Extending Context Window of Large Language Models via Semantic Compression},
 url = {https://aclanthology.org/2024.findings-acl.306},
 year = {2024}
}

@inproceedings{jiang-etal-2024-longllmlingua,
 address = {Bangkok, Thailand},
 author = {Huiqiang Jiang and Qianhui Wu and and Xufang Luo and Dongsheng Li and Chin-Yew Lin and Yuqing Yang and Lili Qiu},
 booktitle = {Proceedings of the 62nd Annual Meeting of the Association for Computational Linguistics (Volume 1: Long Papers)},
 editor = {Ku, Lun-Wei  and
Martins, Andre  and
Srikumar, Vivek},
 pages = {1658--1677},
 publisher = {Association for Computational Linguistics},
 title = {{L}ong{LLML}ingua: Accelerating and Enhancing {LLM}s in Long Context Scenarios via Prompt Compression},
 url = {https://aclanthology.org/2024.acl-long.91},
 year = {2024}
}

@inproceedings{jiang-etal-2023-llmlingua,
 address = {Singapore},
 author = {Huiqiang Jiang and Qianhui Wu and Chin-Yew Lin and Yuqing Yang and Lili Qiu},
 booktitle = {Proceedings of the 2023 Conference on Empirical Methods in Natural Language Processing},
 doi = {10.18653/v1/2023.emnlp-main.825},
 editor = {Bouamor, Houda  and
Pino, Juan  and
Bali, Kalika},
 pages = {13358--13376},
 publisher = {Association for Computational Linguistics},
 title = {{LLML}ingua: Compressing Prompts for Accelerated Inference of Large Language Models},
 url = {https://aclanthology.org/2023.emnlp-main.825},
 year = {2023}
}

@article{wang2023recursively,
 author = {Wang, Qingyue and Ding, Liang and Cao, Yanan and Tian, Zhiliang and Wang, Shi and Tao, Dacheng and Guo, Li},
 journal = {ArXiv preprint},
 title = {Recursively summarizing enables long-term dialogue memory in large language models},
 url = {https://arxiv.org/abs/2308.15022},
 volume = {abs/2308.15022},
 year = {2023}
}

@misc{zhou2023recurrentgpt,
 archiveprefix = {arXiv},
 author = {Wangchunshu Zhou and Yuchen Eleanor Jiang and Peng Cui and Tiannan Wang and Zhenxin Xiao and Yifan Hou and Ryan Cotterell and Mrinmaya Sachan},
 eprint = {2305.13304},
 primaryclass = {cs.CL},
 title = {RecurrentGPT: Interactive Generation of (Arbitrarily) Long Text},
 year = {2023}
}

@inproceedings{Chevalier2023AdaptingLM,
 address = {Singapore},
 author = {Chevalier, Alexis  and
Wettig, Alexander  and
Ajith, Anirudh  and
Chen, Danqi},
 booktitle = {Proceedings of the 2023 Conference on Empirical Methods in Natural Language Processing},
 doi = {10.18653/v1/2023.emnlp-main.232},
 editor = {Bouamor, Houda  and
Pino, Juan  and
Bali, Kalika},
 pages = {3829--3846},
 publisher = {Association for Computational Linguistics},
 title = {Adapting Language Models to Compress Contexts},
 url = {https://aclanthology.org/2023.emnlp-main.232},
 year = {2023}
}

@inproceedings{wingate-etal-2022-prompt,
 address = {Abu Dhabi, United Arab Emirates},
 author = {Wingate, David  and
Shoeybi, Mohammad  and
Sorensen, Taylor},
 booktitle = {Findings of the Association for Computational Linguistics: EMNLP 2022},
 doi = {10.18653/v1/2022.findings-emnlp.412},
 pages = {5621--5634},
 publisher = {Association for Computational Linguistics},
 title = {Prompt Compression and Contrastive Conditioning for Controllability and Toxicity Reduction in Language Models},
 url = {https://aclanthology.org/2022.findings-emnlp.412},
 year = {2022}
}

@inproceedings{hotpotqa,
 address = {Brussels, Belgium},
 author = {Yang, Zhilin  and
Qi, Peng  and
Zhang, Saizheng  and
Bengio, Yoshua  and
Cohen, William  and
Salakhutdinov, Ruslan  and
Manning, Christopher D.},
 booktitle = {Proceedings of the 2018 Conference on Empirical Methods in Natural Language Processing},
 doi = {10.18653/v1/D18-1259},
 pages = {2369--2380},
 publisher = {Association for Computational Linguistics},
 title = {{H}otpot{QA}: A Dataset for Diverse, Explainable Multi-hop Question Answering},
 url = {https://aclanthology.org/D18-1259},
 year = {2018}
}

@inproceedings{tay2020long,
 author = {Yi Tay and
Mostafa Dehghani and
Samira Abnar and
Yikang Shen and
Dara Bahri and
Philip Pham and
Jinfeng Rao and
Liu Yang and
Sebastian Ruder and
Donald Metzler},
 bibsource = {dblp computer science bibliography, https://dblp.org},
 booktitle = {9th International Conference on Learning Representations, {ICLR} 2021,
Virtual Event, Austria, May 3-7, 2021},
 publisher = {OpenReview.net},
 title = {Long Range Arena : {A} Benchmark for Efficient Transformers},
 url = {https://openreview.net/forum?id=qVyeW-grC2k},
 year = {2021}
}

@inproceedings{hsieh2024ruler,
  title={{RULER}: What's the Real Context Size of Your Long-Context Language Models?},
  author={Hsieh, Cheng-Ping and Sun, Simeng and Kriman, Samuel and Acharya, Shantanu and Rekesh, Dima and Jia, Fei and Zhang, Yang and Ginsburg, Boris},
  booktitle={First Conference on Language Modeling},
  year={2024},
  url={https://openreview.net/forum?id=kIoBbc76Sy}
}

@article{liu2024lost,
 address = {Cambridge, MA},
 author = {Liu, Nelson F.  and
Lin, Kevin  and
Hewitt, John  and
Paranjape, Ashwin  and
Bevilacqua, Michele  and
Petroni, Fabio  and
Liang, Percy},
 doi = {10.1162/tacl_a_00638},
 journal = {Transactions of the Association for Computational Linguistics},
 pages = {157--173},
 publisher = {MIT Press},
 title = {Lost in the Middle: How Language Models Use Long Contexts},
 url = {https://aclanthology.org/2024.tacl-1.9},
 volume = {12},
 year = {2024}
}

@misc{longchat,
 author = {Dacheng Li and Rulin Shao and others},
 title = {How Long Can Open-Source {LLMs} Truly Promise on Context Length?},
 url = {https://lmsys.org/blog/2023-06-29-longchat},
 year = {2023}
}

@inproceedings{mohtashami2023landmark,
  title={Landmark Attention: Random-Access Infinite Context Length for Transformers},
  author={Mohtashami, Amirkeivan and Jaggi, Martin},
  booktitle={Advances in Neural Information Processing Systems},
  volume={36},
  year={2023}
}

@inproceedings{an2023eval,
  title={{L-Eval}: Instituting Standardized Evaluation for Long Context Language Models},
  author={An, Chenxin and Gong, Shansan and Zhong, Ming and Zhao, Xingjian and Li, Mukai and Zhang, Jun and Kong, Lingpeng and Qiu, Xipeng},
  booktitle={Proceedings of the 62nd Annual Meeting of the Association for Computational Linguistics (Volume 1: Long Papers)},
  pages={14388--14411},
  year={2024},
  address={Bangkok, Thailand},
  publisher={Association for Computational Linguistics}
}

@inproceedings{shaham2023zeroscrolls,
 address = {Singapore},
 author = {Shaham, Uri  and
Ivgi, Maor  and
Efrat, Avia  and
Berant, Jonathan  and
Levy, Omer},
 booktitle = {Findings of the Association for Computational Linguistics: EMNLP 2023},
 doi = {10.18653/v1/2023.findings-emnlp.536},
 editor = {Bouamor, Houda  and
Pino, Juan  and
Bali, Kalika},
 pages = {7977--7989},
 publisher = {Association for Computational Linguistics},
 title = {{Z}ero{SCROLLS}: A Zero-Shot Benchmark for Long Text Understanding},
 url = {https://aclanthology.org/2023.findings-emnlp.536},
 year = {2023}
}

@inproceedings{zhang2024infty,
  title={$\infty$-Bench: Extending Long Context Evaluation Beyond 100{K} Tokens},
  author={Zhang, Xinrong and Chen, Yingfa and Hu, Shengding and Xu, Zihang and Chen, Junhao and Hao, Moo Khai and Han, Xu and Thai, Zhen Leng and Wang, Shuo and Liu, Zhiyuan and Sun, Maosong},
  booktitle={Proceedings of the 62nd Annual Meeting of the Association for Computational Linguistics (Volume 1: Long Papers)},
  pages={15262--15277},
  year={2024},
  address={Bangkok, Thailand},
  publisher={Association for Computational Linguistics}
}

@inproceedings{liu2024minicache,
  title={{MiniCache}: {KV} Cache Compression in Depth Dimension for Large Language Models},
  author={Liu, Akide and Liu, Jing and Pan, Zizheng and He, Yefei and Haffari, Gholamreza and Zhuang, Bohan},
  booktitle={Advances in Neural Information Processing Systems},
  volume={37},
  year={2024}
}

@inproceedings{sun2024yoco,
  title={You Only Cache Once: Decoder-Decoder Architectures for Language Models},
  author={Sun, Yutao and Dong, Li and Zhu, Yi and Huang, Shaohan and Wang, Wenhui and Ma, Shuming and Zhang, Quanlu and Wang, Jianyong and Wei, Furu},
  booktitle={Advances in Neural Information Processing Systems},
  volume={37},
  year={2024}
}

@inproceedings{brandon2024reducing,
  title={Reducing Transformer Key-Value Cache Size with Cross-Layer Attention},
  author={Brandon, William and Mishra, Mayank and Nrusimha, Aniruddha and Panda, Rameswar and Ragan-Kelly, Jonathan},
  booktitle={Advances in Neural Information Processing Systems},
  volume={37},
  year={2024}
}

@inproceedings{wu2024layercondensedkvcacheefficient,
  title={Layer-Condensed {KV} Cache for Efficient Inference of Large Language Models},
  author={Wu, Haoyi and Tu, Kewei},
  booktitle={Proceedings of the 62nd Annual Meeting of the Association for Computational Linguistics (Volume 1: Long Papers)},
  pages={11175--11188},
  year={2024},
  address={Bangkok, Thailand},
  publisher={Association for Computational Linguistics},
  url={https://aclanthology.org/2024.acl-long.602/},
  doi={10.18653/v1/2024.acl-long.602}
}

@article{deepseekr1,
  title={{DeepSeek-R1} Incentivizes Reasoning in {LLM}s through Reinforcement Learning},
  author={{DeepSeek-AI} and Guo, Daya and Yang, Dejian and Zhang, Haowei and Song, Junxiao and Zhang, Ruoyu and Xu, Runxin and Zhu, Qihao and Ma, Shirong and Wang, Peiyi and others},
  journal={Nature},
  year={2025},
  doi={10.1038/s41586-025-09422-z}
}

@inproceedings{longbenchv2,
  title={{LongBench v2}: Towards Deeper Understanding and Reasoning on Realistic Long-context Multitasks},
  author={Bai, Yushi and Tu, Shangqing and Zhang, Jiajie and Peng, Hao and Wang, Xiaozhi and Lv, Xin and Cao, Shulin and Xu, Jiazheng and Hou, Lei and Dong, Yuxiao and Tang, Jie and Li, Juanzi},
  booktitle={Proceedings of the 63rd Annual Meeting of the Association for Computational Linguistics (Volume 1: Long Papers)},
  pages={3639--3664},
  year={2025},
  address={Vienna, Austria},
  publisher={Association for Computational Linguistics}
}

@inproceedings{longbench,
  title={{LongBench}: A Bilingual, Multitask Benchmark for Long Context Understanding},
  author={Bai, Yushi and Lv, Xin and Zhang, Jiajie and Lyu, Hongchang and Tang, Jiankai and Huang, Zhidian and Du, Zhengxiao and Liu, Xiao and Zeng, Aohan and Hou, Lei and Dong, Yuxiao and Tang, Jie and Li, Juanzi},
  booktitle={Proceedings of the 62nd Annual Meeting of the Association for Computational Linguistics (Volume 1: Long Papers)},
  pages={3119--3137},
  year={2024},
  address={Bangkok, Thailand},
  publisher={Association for Computational Linguistics},
  doi={10.18653/v1/2024.acl-long.172}
}

@misc{grok,
 author = {X.AI},
 title = {Announcing Grok-1.5},
 url = {https://x.ai/blog/grok-1.5},
 year = {2024}
}

@misc{jamba,
 author = {AI21},
 title = {Introducing Jamba: AI21's Groundbreaking SSM-Transformer Model},
 url = {https://www.ai21.com/blog/announcing-jamba},
 year = {2024}
}

@inproceedings{ge2023model,
  title={Model Tells You What to Discard: Adaptive {KV} Cache Compression for {LLM}s},
  author={Ge, Suyu and Zhang, Yunan and Liu, Liyuan and Zhang, Minjia and Han, Jiawei and Gao, Jianfeng},
  booktitle={The Twelfth International Conference on Learning Representations},
  year={2024},
  url={https://openreview.net/forum?id=uNrFpDPMyo}
}

@article{deepseekv3,
 author = {Liu, Aixin and Feng, Bei and Xue, Bing and Wang, Bingxuan and Wu, Bochao and Lu, Chengda and Zhao, Chenggang and Deng, Chengqi and Zhang, Chenyu and Ruan, Chong and others},
 journal = {ArXiv preprint},
 title = {DeepSeek-V3 Technical Report},
 url = {https://arxiv.org/abs/2412.19437},
 volume = {abs/2412.19437},
 year = {2024}
}

@misc{deepseekv2,
 archiveprefix = {arXiv},
 author = {DeepSeek-AI},
 eprint = {2405.04434},
 primaryclass = {cs.CL},
 title = {DeepSeek-V2: A Strong, Economical, and Efficient Mixture-of-Experts Language Model},
 year = {2024}
}

@misc{claude3,
 author = {Anthropic},
 title = {Introducing the next generation of Claude},
 url = {https://www.anthropic.com/news/claude-3-family},
 year = {2024}
}

@article{geminiteam2024gemini,
 author = {Reid, Machel and Savinov, Nikolay and Teplyashin, Denis and Lepikhin, Dmitry and Lillicrap, Timothy and Alayrac, Jean-baptiste and Soricut, Radu and Lazaridou, Angeliki and Firat, Orhan and Schrittwieser, Julian and others},
 journal = {ArXiv preprint},
 title = {Gemini 1.5: Unlocking multimodal understanding across millions of tokens of context},
 url = {https://arxiv.org/abs/2403.05530},
 volume = {abs/2403.05530},
 year = {2024}
}

@inproceedings{chen2023longlora,
 author = {Chen, Yukang and Qian, Shengju and Tang, Haotian and Lai, Xin and Liu, Zhijian and Han, Song and Jia, Jiaya},
 booktitle = {The Twelfth International Conference on Learning Representations},
 title = {LongLoRA: Efficient Fine-tuning of Long-Context Large Language Models},
 year = {2023}
}

@inproceedings{peng2024yarn,
 author = {Bowen Peng and Jeffrey Quesnelle and Honglu Fan and Enrico Shippole},
 booktitle = {The Twelfth International Conference on Learning Representations},
 title = {Ya{RN}: Efficient Context Window Extension of Large Language Models},
 url = {https://openreview.net/forum?id=wHBfxhZu1u},
 year = {2024}
}

@inproceedings{xiong2023effective,
 address = {Mexico City, Mexico},
 author = {Xiong, Wenhan  and
Liu, Jingyu  and
Molybog, Igor  and
Zhang, Hejia  and
Bhargava, Prajjwal  and
Hou, Rui  and
Martin, Louis  and
Rungta, Rashi  and
Sankararaman, Karthik Abinav  and
Oguz, Barlas  and
Khabsa, Madian  and
Fang, Han  and
Mehdad, Yashar  and
Narang, Sharan  and
Malik, Kshitiz  and
Fan, Angela  and
Bhosale, Shruti  and
Edunov, Sergey  and
Lewis, Mike  and
Wang, Sinong  and
Ma, Hao},
 booktitle = {Proceedings of the 2024 Conference of the North American Chapter of the Association for Computational Linguistics: Human Language Technologies (Volume 1: Long Papers)},
 editor = {Duh, Kevin  and
Gomez, Helena  and
Bethard, Steven},
 pages = {4643--4663},
 publisher = {Association for Computational Linguistics},
 title = {Effective Long-Context Scaling of Foundation Models},
 url = {https://aclanthology.org/2024.naacl-long.260},
 year = {2024}
}

@article{chen2023extending,
 author = {Chen, Shouyuan and Wong, Sherman and Chen, Liangjian and Tian, Yuandong},
 journal = {ArXiv preprint},
 title = {Extending context window of large language models via positional interpolation},
 url = {https://arxiv.org/abs/2306.15595},
 volume = {abs/2306.15595},
 year = {2023}
}

@inproceedings{streamingllm,
 author = {Guangxuan Xiao and Yuandong Tian and Beidi Chen and Song Han and Mike Lewis},
 booktitle = {The Twelfth International Conference on Learning Representations},
 title = {Efficient Streaming Language Models with Attention Sinks},
 url = {https://openreview.net/forum?id=NG7sS51zVF},
 year = {2024}
}

@article{jacobs2023deepspeed,
 author = {Sam Ade Jacobs and others},
 journal = {ArXiv preprint},
 title = {{DeepSpeed Ulysses}: System Optimizations for Enabling Training of Extreme Long Sequence {Transformer} Models},
 url = {https://arxiv.org/abs/2309.14509},
 volume = {abs/2309.14509},
 year = {2023}
}

@inproceedings{flash-attn,
  title={{FlashAttention}: Fast and Memory-Efficient Exact Attention with {IO}-Awareness},
  author={Dao, Tri and Fu, Daniel and Ermon, Stefano and Rudra, Atri and R{\'e}, Christopher},
  booktitle={Advances in Neural Information Processing Systems},
  volume={35},
  pages={16344--16359},
  year={2022}
}

@inproceedings{flash-attn2,
 author = {Dao, Tri},
 booktitle = {International Conference on Learning Representations (ICLR)},
 title = {Flash{A}ttention-2: Faster Attention with Better Parallelism and Work Partitioning},
 year = {2024}
}

@article{touvron2023llama,
 author = {Touvron, Hugo and Lavril, Thibaut and Izacard, Gautier and Martinet, Xavier and Lachaux, Marie-Anne and Lacroix, Timoth{\'e}e and Rozi{\`e}re, Baptiste and Goyal, Naman and Hambro, Eric and Azhar, Faisal and others},
 journal = {ArXiv preprint},
 title = {Llama: Open and efficient foundation language models},
 url = {https://arxiv.org/abs/2302.13971},
 volume = {abs/2302.13971},
 year = {2023}
}

@article{touvron2023llama2,
 author = {Touvron, Hugo and Martin, Louis and Stone, Kevin and Albert, Peter and Almahairi, Amjad and Babaei, Yasmine and Bashlykov, Nikolay and Batra, Soumya and Bhargava, Prajjwal and Bhosale, Shruti and others},
 journal = {ArXiv preprint},
 title = {Llama 2: Open foundation and fine-tuned chat models},
 url = {https://arxiv.org/abs/2307.09288},
 volume = {abs/2307.09288},
 year = {2023}
}

@article{chowdhery2022palm,
  title={{PaLM}: Scaling Language Modeling with Pathways},
  author={Chowdhery, Aakanksha and Narang, Sharan and Devlin, Jacob and Bosma, Maarten and Mishra, Gaurav and Roberts, Adam and Barham, Paul and Chung, Hyung Won and Sutton, Charles and Gehrmann, Sebastian and others},
  journal={Journal of Machine Learning Research},
  volume={24},
  number={240},
  pages={1--113},
  year={2023}
}

@article{raffel2020exploring,
 author = {Colin Raffel and
Noam Shazeer and
Adam Roberts and
Katherine Lee and
Sharan Narang and
Michael Matena and
Yanqi Zhou and
Wei Li and
Peter J. Liu},
 bibsource = {dblp computer science bibliography, https://dblp.org},
 journal = {J. Mach. Learn. Res.},
 pages = {140:1--140:67},
 title = {Exploring the Limits of Transfer Learning with a Unified Text-to-Text
Transformer},
 url = {http://jmlr.org/papers/v21/20-074.html},
 volume = {21},
 year = {2020}
}

@inproceedings{longgenbench,
 address = {Miami, Florida, USA},
 author = {Liu, Xiang  and
Dong, Peijie  and
Hu, Xuming  and
Chu, Xiaowen},
 booktitle = {Findings of the Association for Computational Linguistics: EMNLP 2024},
 editor = {Al-Onaizan, Yaser  and
Bansal, Mohit  and
Chen, Yun-Nung},
 pages = {865--883},
 publisher = {Association for Computational Linguistics},
 title = {{L}ong{G}en{B}ench: Long-context Generation Benchmark},
 url = {https://aclanthology.org/2024.findings-emnlp.48},
 year = {2024}
}

@article{needle,
 author = {Gregory Kamradt},
 journal = {Github},
 title = {{Needle In A Haystack} - Pressure Testing {LLM}s},
 url = {https://github.com/gkamradt/LLMTest_NeedleInAHaystack/tree/main},
 year = {2023}
}

@article{gsm8k,
 author = {Cobbe, Karl and Kosaraju, Vineet and Bavarian, Mohammad and Chen, Mark and Jun, Heewoo and Kaiser, Lukasz and Plappert, Matthias and Tworek, Jerry and Hilton, Jacob and Nakano, Reiichiro and others},
 journal = {ArXiv preprint},
 title = {Training verifiers to solve math word problems},
 url = {https://arxiv.org/abs/2110.14168},
 volume = {abs/2110.14168},
 year = {2021}
}

@inproceedings{csqa,
 address = {Minneapolis, Minnesota},
 author = {Talmor, Alon  and
Herzig, Jonathan  and
Lourie, Nicholas  and
Berant, Jonathan},
 booktitle = {Proceedings of the 2019 Conference of the North {A}merican Chapter of the Association for Computational Linguistics: Human Language Technologies, Volume 1 (Long and Short Papers)},
 doi = {10.18653/v1/N19-1421},
 pages = {4149--4158},
 publisher = {Association for Computational Linguistics},
 title = {{C}ommonsense{QA}: A Question Answering Challenge Targeting Commonsense Knowledge},
 url = {https://aclanthology.org/N19-1421},
 year = {2019}
}

@inproceedings{mmlu,
  title={Measuring Massive Multitask Language Understanding},
  author={Hendrycks, Dan and Burns, Collin and Basart, Steven and Zou, Andy and Mazeika, Mantas and Song, Dawn and Steinhardt, Jacob},
  booktitle={International Conference on Learning Representations},
  year={2021}
}

@inproceedings{pope2023efficiently,
  title={Efficiently Scaling Transformer Inference},
  author={Pope, Reiner and Douglas, Sholto and Chowdhery, Aakanksha and Devlin, Jacob and Bradbury, James and Heek, Jonathan and Xiao, Kefan and Agrawal, Shivani and Dean, Jeff},
  booktitle={Proceedings of Machine Learning and Systems},
  volume={5},
  year={2023}
}

@inproceedings{adnan2024keyformer,
  title={{Keyformer}: {KV} Cache Reduction through Key Tokens Selection for Efficient Generative Inference},
  author={Adnan, Muhammad and Arunkumar, Akhil and Jain, Gaurav and Nair, Prashant J. and Soloveychik, Ilya and Kamath, Purushotham},
  booktitle={Proceedings of Machine Learning and Systems},
  volume={6},
  year={2024}
}

@inproceedings{yao2024cacheblend,
  title={{CacheBlend}: Fast Large Language Model Serving for {RAG} with Cached Knowledge Fusion},
  author={Yao, Jiayi and Li, Hanchen and Liu, Yuhan and Ray, Siddhant and Cheng, Yihua and Zhang, Qizheng and Du, Kuntai and Lu, Shan and Jiang, Junchen},
  booktitle={Proceedings of the Twentieth European Conference on Computer Systems},
  series={EuroSys '25},
  year={2025},
  doi={10.1145/3689031.3696098}
}

@inproceedings{zhangcam,
 author = {Zhang, Yuxin and Du, Yuxuan and Luo, Gen and Zhong, Yunshan and Zhang, Zhenyu and Liu, Shiwei and Ji, Rongrong},
 booktitle = {Forty-first International Conference on Machine Learning},
 title = {CaM: Cache Merging for Memory-efficient LLMs Inference},
 year = {2024}
}

@inproceedings{yang2024pyramidinfer,
  title={{PyramidInfer}: Pyramid {KV} Cache Compression for High-throughput {LLM} Inference},
  author={Yang, Dongjie and Han, XiaoDong and Gao, Yan and Hu, Yao and Zhang, Shilin and Zhao, Hai},
  booktitle={Findings of the Association for Computational Linguistics: ACL 2024},
  pages={3258--3270},
  year={2024},
  publisher={Association for Computational Linguistics}
}

@inproceedings{guo2024attention,
  title={Attention Score is not All You Need for Token Importance Indicator in {KV} Cache Reduction: Value Also Matters},
  author={Guo, Zhiyu and Kamigaito, Hidetaka and Watanabe, Taro},
  booktitle={Proceedings of the 2024 Conference on Empirical Methods in Natural Language Processing},
  pages={21158--21166},
  year={2024},
  address={Miami, Florida, USA},
  publisher={Association for Computational Linguistics}
}

@inproceedings{fu2024lazyllm,
 author = {Qichen Fu and Minsik Cho and Thomas Merth and Sachin Mehta and Mohammad Rastegari and Mahyar Najibi},
 booktitle = {Workshop on Efficient Systems for Foundation Models II @ ICML2024},
 title = {Lazy{LLM}: Dynamic Token Pruning for Efficient Long Context {LLM} Inference},
 url = {https://openreview.net/forum?id=gGZD1dsJqZ},
 year = {2024}
}

@inproceedings{liu2024scissorhands,
  title={Scissorhands: Exploiting the Persistence of Importance Hypothesis for {LLM} {KV} Cache Compression at Test Time},
  author={Liu, Zichang and Desai, Aditya and Liao, Fangshuo and Wang, Weitao and Xie, Victor and Xu, Zhaozhuo and Kyrillidis, Anastasios and Shrivastava, Anshumali},
  booktitle={Advances in Neural Information Processing Systems},
  volume={36},
  year={2023}
}

@inproceedings{tang2024quest,
  title={{QUEST}: Query-Aware Sparsity for Efficient Long-Context {LLM} Inference},
  author={Tang, Jiaming and Zhao, Yilong and Zhu, Kan and Xiao, Guangxuan and Kasikci, Baris and Han, Song},
  booktitle={Proceedings of the 41st International Conference on Machine Learning},
  series={Proceedings of Machine Learning Research},
  volume={235},
  pages={47901--47911},
  year={2024},
  publisher={PMLR}
}

@inproceedings{feng2024ada,
  title={{Ada-KV}: Optimizing {KV} Cache Eviction by Adaptive Budget Allocation for Efficient {LLM} Inference},
  author={Feng, Yuan and Lv, Junlin and Cao, Yukun and Xie, Xike and Zhou, S. Kevin},
  booktitle={Advances in Neural Information Processing Systems},
  volume={38},
  year={2025},
  url={https://openreview.net/forum?id=tcisuhGsQZ}
}

@inproceedings{yuan2024kv,
  title={{KV} Cache Compression, But What Must We Give in Return? A Comprehensive Benchmark of Long Context Capable Approaches},
  author={Yuan, Jiayi and Liu, Hongyi and Zhong, Shaochen and Chuang, Yu-Neng and Li, Songchen and Wang, Guanchu and Le, Duy and Jin, Hongye and Chaudhary, Vipin and Xu, Zhaozhuo and Liu, Zirui and Hu, Xia},
  booktitle={Findings of the Association for Computational Linguistics: EMNLP 2024},
  pages={4623--4648},
  year={2024},
  address={Miami, Florida, USA},
  publisher={Association for Computational Linguistics}
}

@inproceedings{lin2021truthfulqa,
  title={{TruthfulQA}: Measuring How Models Mimic Human Falsehoods},
  author={Lin, Stephanie and Hilton, Jacob and Evans, Owain},
  booktitle={Proceedings of the 60th Annual Meeting of the Association for Computational Linguistics (Volume 1: Long Papers)},
  pages={3214--3252},
  year={2022},
  address={Dublin, Ireland},
  publisher={Association for Computational Linguistics},
  doi={10.18653/v1/2022.acl-long.229}
}

@inproceedings{hartvigsen2022toxigen,
  title={{ToxiGen}: A Large-Scale Machine-Generated Dataset for Adversarial and Implicit Hate Speech Detection},
  author={Hartvigsen, Thomas and Gabriel, Saadia and Palangi, Hamid and Sap, Maarten and Ray, Dipankar and Kamar, Ece},
  booktitle={Proceedings of the 60th Annual Meeting of the Association for Computational Linguistics (Volume 1: Long Papers)},
  pages={3309--3326},
  year={2022},
  address={Dublin, Ireland},
  publisher={Association for Computational Linguistics}
}

@inproceedings{liang2021towards,
 author = {Liang, Paul Pu and Wu, Chiyu and Morency, Louis-Philippe and Salakhutdinov, Ruslan},
 booktitle = {International Conference on Machine Learning},
 organization = {PMLR},
 pages = {6565--6576},
 title = {Towards understanding and mitigating social biases in language models},
 year = {2021}
}

@article{zhu2023promptbench,
  title={{PromptBench}: A Unified Library for Evaluation of Large Language Models},
  author={Zhu, Kaijie and Zhao, Qinlin and Chen, Hao and Wang, Jindong and Xie, Xing},
  journal={Journal of Machine Learning Research},
  volume={25},
  number={254},
  pages={1--22},
  year={2024}
}

@inproceedings{shen2024anything,
 author = {Shen, Xinyue and Chen, Zeyuan and Backes, Michael and Shen, Yun and Zhang, Yang},
 booktitle = {Proceedings of the 2024 on ACM SIGSAC Conference on Computer and Communications Security},
 pages = {1671--1685},
 title = {" do anything now": Characterizing and evaluating in-the-wild jailbreak prompts on large language models},
 year = {2024}
}

@inproceedings{deng2023multilingual,
  title={Multilingual Jailbreak Challenges in Large Language Models},
  author={Deng, Yue and Zhang, Wenxuan and Pan, Sinno Jialin and Bing, Lidong},
  booktitle={The Twelfth International Conference on Learning Representations},
  year={2024},
  url={https://openreview.net/forum?id=vESNKdEMGp}
}

@inproceedings{li2024should,
  title={Should We Really Edit Language Models? On the Evaluation of Edited Language Models},
  author={Li, Qi and Liu, Xiang and Tang, Zhenheng and Dong, Peijie and Li, Zeyu and Pan, Xinglin and Chu, Xiaowen},
  booktitle={Advances in Neural Information Processing Systems},
  volume={37},
  year={2024}
}

@article{pyramidkv,
 author = {Cai, Zefan and Zhang, Yichi and Gao, Bofei and Liu, Yuliang and Liu, Tianyu and Lu, Keming and Xiong, Wayne and Dong, Yue and Chang, Baobao and Hu, Junjie and others},
 journal = {ArXiv preprint},
 title = {Pyramidkv: Dynamic kv cache compression based on pyramidal information funneling},
 url = {https://arxiv.org/abs/2406.02069},
 volume = {abs/2406.02069},
 year = {2024}
}

@inproceedings{h2o,
  title={{H}$_2${O}: Heavy-Hitter Oracle for Efficient Generative Inference of Large Language Models},
  author={Zhang, Zhenyu and Sheng, Ying and Zhou, Tianyi and Chen, Tianlong and Zheng, Lianmin and Cai, Ruisi and Song, Zhao and Tian, Yuandong and R{\'e}, Christopher and Barrett, Clark and Wang, Zhangyang and Chen, Beidi},
  booktitle={Advances in Neural Information Processing Systems},
  volume={36},
  year={2023}
}

@inproceedings{snapkv,
  title={{SnapKV}: {LLM} Knows What You Are Looking for Before Generation},
  author={Li, Yuhong and Huang, Yingbing and Yang, Bowen and Venkitesh, Bharat and Locatelli, Acyr and Ye, Hanchen and Cai, Tianle and Lewis, Patrick and Chen, Deming},
  booktitle={Advances in Neural Information Processing Systems},
  volume={37},
  year={2024}
}

@misc{eval-harness,
 author = {Gao, Leo and Tow, Jonathan and Abbasi, Baber and Biderman, Stella and Black, Sid and DiPofi, Anthony and Foster, Charles and Golding, Laurence and Hsu, Jeffrey and Le Noac'h, Alain and Li, Haonan and McDonell, Kyle and Muennighoff, Niklas and Ociepa, Chris and Phang, Jason and Reynolds, Laria and Schoelkopf, Hailey and Skowron, Aviya and Sutawika, Lintang and Tang, Eric and Thite, Anish and Wang, Ben and Wang, Kevin and Zou, Andy},
 doi = {10.5281/zenodo.10256836},
 publisher = {Zenodo},
 title = {A framework for few-shot language model evaluation},
 url = {https://zenodo.org/records/10256836},
 version = {v0.4.0},
 year = {2023}
}

@article{jiang2023mistral,
 author = {Jiang, Albert Q and Sablayrolles, Alexandre and Mensch, Arthur and Bamford, Chris and Chaplot, Devendra Singh and Casas, Diego de las and Bressand, Florian and Lengyel, Gianna and Lample, Guillaume and Saulnier, Lucile and others},
 journal = {ArXiv preprint},
 title = {Mistral 7B},
 url = {https://arxiv.org/abs/2310.06825},
 volume = {abs/2310.06825},
 year = {2023}
}

@article{dubey2024llama,
 author = {Dubey, Abhimanyu and Jauhri, Abhinav and Pandey, Abhinav and Kadian, Abhishek and Al-Dahle, Ahmad and Letman, Aiesha and Mathur, Akhil and Schelten, Alan and Yang, Amy and Fan, Angela and others},
 journal = {ArXiv preprint},
 title = {The llama 3 herd of models},
 url = {https://arxiv.org/abs/2407.21783},
 volume = {abs/2407.21783},
 year = {2024}
}

@inproceedings{agarwal2024many,
  title={Many-Shot In-Context Learning},
  author={Agarwal, Rishabh and Singh, Avi and Zhang, Lei M. and Bohnet, Bernd and Rosias, Luis and Chan, Stephanie and Zhang, Biao and Anand, Ankesh and Abbas, Zaheer and Nova, Azade and Co-Reyes, John D. and Chu, Eric and Behbahani, Feryal and Faust, Aleksandra and Larochelle, Hugo},
  booktitle={Advances in Neural Information Processing Systems},
  volume={37},
  year={2024}
}

@inproceedings{luo2024jailbreakv,
 author = {Weidi Luo and Siyuan Ma and Xiaogeng Liu and Xiaoyu Guo and Chaowei Xiao},
 booktitle = {First Conference on Language Modeling},
 title = {JailBreakV: A Benchmark for Assessing the Robustness of MultiModal Large Language Models against Jailbreak Attacks},
 url = {https://openreview.net/forum?id=GC4mXVfquq},
 year = {2024}
}

@article{chen2021evaluating,
 author = {Chen, Mark and Tworek, Jerry and Jun, Heewoo and Yuan, Qiming and Pinto, Henrique Ponde De Oliveira and Kaplan, Jared and Edwards, Harri and Burda, Yuri and Joseph, Nicholas and Brockman, Greg and others},
 journal = {ArXiv preprint},
 title = {Evaluating large language models trained on code},
 url = {https://arxiv.org/abs/2107.03374},
 volume = {abs/2107.03374},
 year = {2021}
}

@article{austin2021program,
 author = {Austin, Jacob and Odena, Augustus and Nye, Maxwell and Bosma, Maarten and Michalewski, Henryk and Dohan, David and Jiang, Ellen and Cai, Carrie and Terry, Michael and Le, Quoc and others},
 journal = {ArXiv preprint},
 title = {Program synthesis with large language models},
 url = {https://arxiv.org/abs/2108.07732},
 volume = {abs/2108.07732},
 year = {2021}
}

@inproceedings{bbh,
  title={Challenging {BIG}-Bench Tasks and Whether Chain-of-Thought Can Solve Them},
  author={Suzgun, Mirac and Scales, Nathan and Sch{\"a}rli, Nathanael and Gehrmann, Sebastian and Tay, Yi and Chung, Hyung Won and Chowdhery, Aakanksha and Le, Quoc and Chi, Ed and Zhou, Denny and Wei, Jason},
  booktitle={Findings of the Association for Computational Linguistics: ACL 2023},
  pages={13003--13051},
  year={2023},
  address={Toronto, Canada},
  publisher={Association for Computational Linguistics},
  doi={10.18653/v1/2023.findings-acl.824}
}

@article{bigbench,
  title={Beyond the Imitation Game: Quantifying and Extrapolating the Capabilities of Language Models},
  author={Srivastava, Aarohi and Rastogi, Abhinav and Rao, Abhishek and others},
  journal={Transactions on Machine Learning Research},
  issn={2835-8856},
  year={2023},
  url={https://openreview.net/forum?id=uyTL5Bvosj}
}

@inproceedings{math,
  title={Measuring Mathematical Problem Solving with the {MATH} Dataset},
  author={Hendrycks, Dan and Burns, Collin and Kadavath, Saurav and Arora, Akul and Basart, Steven and Tang, Eric and Song, Dawn and Steinhardt, Jacob},
  booktitle={Advances in Neural Information Processing Systems Track on Datasets and Benchmarks},
  year={2021}
}

@inproceedings{mathqa,
 address = {Minneapolis, Minnesota},
 author = {Amini, Aida  and
Gabriel, Saadia  and
Lin, Shanchuan  and
Koncel-Kedziorski, Rik  and
Choi, Yejin  and
Hajishirzi, Hannaneh},
 booktitle = {Proceedings of the 2019 Conference of the North {A}merican Chapter of the Association for Computational Linguistics: Human Language Technologies, Volume 1 (Long and Short Papers)},
 doi = {10.18653/v1/N19-1245},
 pages = {2357--2367},
 publisher = {Association for Computational Linguistics},
 title = {{M}ath{QA}: Towards Interpretable Math Word Problem Solving with Operation-Based Formalisms},
 url = {https://aclanthology.org/N19-1245},
 year = {2019}
}

@inproceedings{tang2025lottery,
  title={The Lottery {LLM} Hypothesis, Rethinking What Abilities Should {LLM} Compression Preserve?},
  author={Tang, Zhenheng and Liu, Xiang and Wang, Qian and Dong, Peijie and He, Bingsheng and Chu, Xiaowen and Li, Bo},
  booktitle={ICLR Blogposts Track},
  year={2025},
  url={https://arxiv.org/abs/2502.17535}
}

@inproceedings{dong2025compressedllm,
  title={Can Compressed {LLM}s Truly Act? An Empirical Evaluation of Agentic Capabilities in {LLM} Compression},
  author={Dong, Peijie and Tang, Zhenheng and Liu, Xiang and Li, Lujun and Chu, Xiaowen and Li, Bo},
  booktitle={Proceedings of the 42nd International Conference on Machine Learning},
  series={Proceedings of Machine Learning Research},
  publisher={PMLR},
  year={2025},
  url={https://proceedings.mlr.press/v267/dong25k.html}
}

@article{li2025antkv,
  title={{AnTKV}: Anchor Token-Aware Sub-Bit Vector Quantization for {KV} Cache in Large Language Models},
  author={Li, Zeyu and Xiao, Chuanfu and Wang, Yang and Liu, Xiang and Tang, Zhenheng and Lu, Baotong and Yang, Mao and Chen, Xinyu and Chu, Xiaowen},
  journal={arXiv preprint arXiv:2506.19505},
  year={2025},
  url={https://arxiv.org/abs/2506.19505}
}

@article{chen2026sonic,
  title={{SONIC}: Segmented Optimized Nexus for Information Compression in Key-Value Caching},
  author={Chen, Hong and Liu, Xiang and Wang, Bo and Fan, Yuxuan and Chu, Yuanlin and Li, Zongluo and Chu, Xiaowen and Hu, Xuming},
  journal={arXiv preprint arXiv:2601.21927},
  year={2026},
  url={https://arxiv.org/abs/2601.21927}
}

@inproceedings{liu2026diffadapt,
  title={{DiffAdapt}: Difficulty-Adaptive Reasoning for Token-Efficient {LLM} Inference},
  author={Liu, Xiang and Hu, Xuming and Chu, Xiaowen and Choi, Eunsol},
  booktitle={The Fourteenth International Conference on Learning Representations},
  year={2026},
  url={https://arxiv.org/abs/2510.19669}
}

@inproceedings{li2026reasoningserving,
  title={Reasoning Language Model Inference Serving Unveiled: An Empirical Study},
  author={Li, Qi and Wu, Junpan and Liu, Xiang and Wang, Yuxin and Li, Zeyu and Tang, Zhenheng and Chen, Yuhan and Shi, Shaohuai and Chu, Xiaowen},
  booktitle={The Fourteenth International Conference on Learning Representations},
  year={2026},
  url={https://arxiv.org/abs/2510.18672}
}

@inproceedings{pan2024lisa,
  title={{LISA}: Layerwise Importance Sampling for Memory-Efficient Large Language Model Fine-Tuning},
  author={Pan, Rui and Liu, Xiang and Diao, Shizhe and Pi, Renjie and Zhang, Jipeng and Han, Chi and Zhang, Tong},
  booktitle={Advances in Neural Information Processing Systems 37 (NeurIPS 2024)},
  year={2024},
  url={http://papers.nips.cc/paper_files/paper/2024/hash/687163285b8affc8ee933bdca8e75747-Abstract-Conference.html}
}

@article{zhang2023dissecting,
  title={Dissecting the Runtime Performance of the Training, Fine-tuning, and Inference of Large Language Models},
  author={Zhang, Longteng and Liu, Xiang and Li, Zeyu and Pan, Xinglin and Dong, Peijie and Fan, Ruibo and Guo, Rui and Wang, Xin and Luo, Qiong and Shi, Shaohuai and Chu, Xiaowen},
  journal={arXiv preprint arXiv:2311.03687},
  year={2023}
}
\bibliographystyle{icml2026}

\newpage
\appendix
\onecolumn

\section{Code Availability}
\label{app:code_availability}
The implementation of \method{}, \bcmk{} evaluation scripts, and the configuration files used in our experiments are released at \url{https://github.com/Zefan-Cai/KVCache-Factory}. \bcmk{} extends this open-source KV cache compression framework with additional fundamental-capability benchmarks (world knowledge, commonsense reasoning, arithmetic reasoning, code generation, safety, and long-context generation) and the integrated \method{} compression strategy. All experiments use publicly available models (e.g., LLaMA-3.1, Mistral-7B, DeepSeek-R1, Qwen3) and standard academic datasets evaluated through the \texttt{lm-evaluation-harness} and \texttt{KVpress} interfaces. Detailed hyperparameters and experimental configurations are provided in Appendix~\ref{app:experiments_design}.

\section{Extended Limitations}
\label{app:limitations}

While \bcmk{} provides a comprehensive evaluation of KV cache compression on fundamental capabilities and \method{} demonstrates significant improvements, there are a few aspects that warrant further exploration.
First, our analysis primarily focuses on open-weights models (e.g., LLaMA-3, Mistral, DeepSeek) where accessing and manipulating the KV cache is straightforward. Extending these findings to proprietary, API-based models remains a direction for future research due to current access restrictions on internal inference states.
Second, although \method{} achieves superior performance by preserving semantic units, it relies on a heuristic for shot importance that, while effective, introduces a negligible calculation step during the prefill phase. As shown in our efficiency analysis, the throughput gains from compression significantly outweigh this cost, though further kernel-level optimizations could be explored.
Finally, while we cover five distinct categories of fundamental abilities, the landscape of LLM capabilities is vast. Future work could extend our framework to assess more specialized domains or multimodal tasks as efficient inference techniques evolve.

\section{Extended Related Work}
\label{appendix:related_work}

\paragraph{Key--value Cache Optimization Techniques}

KV cache is the core component in LLM inference, which avoids repetitive computations by caching Key and Value vectors. However, the cost of caching KV increases exponentially with the expansion of the model size and the length of the context~\cite{pope2023efficiently}. Some approaches have been published to alleviate the problem. For example, KV Compression designs efficient content selection strategies to filter and manage tokens~\cite{h2o,adnan2024keyformer}. Some methods identify important tokens by focusing on high attention allocation~\cite{snapkv}, while others optimize token selection by combining attention scores with value vector norms to improve importance evaluation~\cite{guo2024attention}. Techniques like PyramidInfer reduce critical contexts layer by layer based on the distribution of attention scores~\cite{yang2024pyramidinfer}, and StreamingLLM preserves attention sinks to maintain stable performance in extended sequences~\cite{streamingllm}. Researchers reduce storage costs by merging similar context representations and solving input disturbances caused by compression~\cite{zhangcam}. For example, CaM~\cite{zhangcam} works by integrating the KV cache to be dropped into the retained cache in proportion to the attention weight.  In addition, \citet{yao2024cacheblend} proposes CacheBlend to achieve a selective KV recompute. Only partial KVs of crucial tokens are updated to reduce the delay in the prefill stage and increase the throughput. In addition, the dynamic budget allocation method is also used to optimize the KV cache, which adjusts the resource allocation in real time according to the importance of the context, providing a balance between performance and efficiency in multitask inference scenarios~\cite{pyramidkv,feng2024ada,kim2025kvzip,kvpress,liu2025flowkv,zhu2025oraclekv}.\citet{scope} proposes a prefill-decoding separation strategy to optimize the KV cache compression.

\paragraph{Evaluation of LLMs' Fundamental Abilities}
Accurately evaluating the fundamental capabilities of large language models is crucial to understand their true potential and limitations. The evaluation typically spans across several key dimensions: world knowledge tasks like MMLU~\cite{mmlu},BBH~\cite{bbh} assess models' grasp of diverse domains through multiple-choice questions; commonsense reasoning tasks such as CSQA~\cite{csqa} evaluate inference and context understanding abilities; arithmetic reasoning benchmarks like GSM8K~\cite{gsm8k} test mathematical problem-solving capabilities through step-by-step reasoning; code generation tasks including HumanEval~\cite{chen2021evaluating,yu2026rethinking} measure the ability to generate functionally correct code; and safety evaluations using benchmarks like JailBreakV~\cite{luo2024jailbreakv} assess models' robustness against harmful content generation. Additionally, long-context benchmarks such as LongBench~\cite{longbench,longbenchv2} and Need-In-A-Haystack (NIAH)~\cite{needle} aiming to evaluate models' long-context summarization and retrieval capabilities. Furthermore, LongGenBench~\cite{longgenbench} evaluates the models' ability to process and generate responses for extended input sequences. And recently, in-context many-shot learning has been recognized as a long-context reasoning paradigm~\cite{agarwal2024many,tang2026your}, which considers the number of shots as a critical factor in the performance of LLM.
Although these tasks typically employ automatic evaluation metrics for standardization, KV cache compression may introduce unique challenges, particularly in tasks requiring complex reasoning chains or extensive knowledge retrieval.

\paragraph{KV cache sharing}
Recent work has explored various strategies for sharing KV caches across transformer layers. The Layer Condensed KV Cache (LCKV) \citep{wu2024layercondensedkvcacheefficient} computes the KV only for the top layer and pairs them with queries from all layers, while optionally retaining standard attention for a few top and bottom layers to mitigate performance degradation. Similarly, You Only Cache Once (YOCO) \citep{sun2024yoco} computes KVs exclusively for the top layer but pairs them with queries from only the top half of layers, employing efficient attention in the bottom layers to maintain a constant cache size. In contrast, Cross-Layer Attention (CLA) \citep{brandon2024reducing} divides layers into groups, pairing queries from all layers in each group with KVs from that group's bottom layer. MiniCache \citep{liu2024minicache} introduces a novel method that merges KV caches in layering while enabling recovery during compute-in-place operations, optimizing the size of the KV cache. These methods illustrate various trade-offs between computation, memory usage, and model performance when sharing KV caches across transformer layers.

\paragraph{Prompting Compression}
Recent advances in prompt compression have yielded innovative approaches to information density optimization in natural language processing. Research by \citet{wingate-etal-2022-prompt} demonstrates how soft prompting techniques can achieve higher information density per token. Building upon this foundation, AutoCompressor \citep{Chevalier2023AdaptingLM} leverages soft prompts to both condense input sequences and expand model context windows. Parallel developments by \citet{zhou2023recurrentgpt} and \citet{wang2023recursively} showcase iterative summarization strategies using LLMs, establishing persistent memory mechanisms particularly beneficial for narrative construction and conversational systems. The progressive development of the LLMLingua framework \citep{jiang-etal-2023-llmlingua,jiang-etal-2024-longllmlingua,fei-etal-2024-extending} has advanced prompt compression capabilities across extended context processing, logical reasoning, and retrieval-augmented generation. Notable contributions from \citet{fei-etal-2024-extending} demonstrate effective context management through automated segmentation and semantic condensation using pre-trained language models.

\paragraph{General Tasks}
General tasks refer to evaluating the overall performance of LLMs under mathematical inference, logic reasoning, and common knowledge. GSM8K~\cite{gsm8k} and MMLU~\cite{mmlu} are representative tasks. The former focuses on the step-by-step reasoning ability of mathematical problem solving, while the latter covers assessment of common sense and expertise in multiple areas. Besides, MATH~\cite{math} spans various mathematical fields, ranging from elementary algebra to calculus, aiming to improve the mathematical problem-solving capabilities of LLMs. Meanwhile, MathQA~\cite{mathqa} is a large-scale dataset comprising approximately 37,000 multiple-choice questions with precise annotations, designed to enhance the interpretability and performance of LLMs. In addition, BBH~\cite{bbh}, a subset of BIG-Bench~\cite{bigbench}, focuses on challenging tasks. BBH includes multi-step reasoning problems, highlighting the importance of Chain-of-Thought prompting in LLMs. Similarly, CSQA~\cite{csqa} is a task that combines knowledge graph-based multi-step reasoning with conversational capabilities. CSQA emphasizes inference and context understanding grounded in knowledge graphs. Normally, the general tasks apply automatic evaluation metrics (e.g. multi-choice accuracy) to ensure comparability and standardization. However, optimization strategies like KV cache compression may introduce challenges in executing the mentioned tasks. Filtering and dropping of contexts are involved in the compression strategy which may lead to an intermediate inference steps missing. In addition, in tasks such as MMLU that are highly dependent on knowledge coverage, compression may weaken the model's ability to capture long context or rare domain knowledge~\cite{yuan2024kv}.

\paragraph{Security Tasks}
Security tasks focus on assessing the robustness and protections of LLMs against harmful content, including truthfulness~\cite{lin2021truthfulqa}, toxicity~\cite{hartvigsen2022toxigen}, and bias~\cite{liang2021towards,tang2026your}. Recently, researchers noticed the weakness of LLMs in adversarial prompts~\cite{zhu2023promptbench,tang2026ghost}, especially in generating illegal or inappropriate content under jailbreak prompts. \citet{shen2024anything} analyze the jailbreak prompts in real cases to reveal the failure of model security mechanism under complex malicious input. Meanwhile, \citet{deng2023multilingual} demonstrates the multilingual jailbreak makes model security in low-resource languages easier to bypass, significantly increasing the probability that users of low-resource languages will generate insecure content. Similar to general tasks, KV optimization techniques can cause the model to ignore potential security threats when dealing with jailbreak prompts, thereby improving the success rate of adversarial prompts~\cite{li2024should}.

\paragraph{Code Generation Tasks}
Code generation tasks test the capacities of LLMs to generate code, which not only requires that the model can generate syntactic code based on natural language description but also has certain logical reasoning abilities. HumanEval~\cite{chen2021evaluating} and MBPP~\cite{austin2021program} are the commonly used benchmarks. They measure the functional correctness of the model by testing the results of the code's execution.

\paragraph{Long-context Tasks}
Recent developments in evaluating long-context models have produced a comprehensive ecosystem of benchmarks, focusing on both comprehension depth and retrieval efficiency. In the comprehension domain, $\infty$-Bench~\citep{zhang2024infty} has established new standards by crafting evaluation scenarios exceeding 100,000 tokens, while LongBench~\citep{longbench,longbenchv2} introduced multilingual assessment frameworks spanning document comprehension, text synthesis, and programming tasks. Further enriching this landscape, ZeroSCROLLS~\citep{shaham2023zeroscrolls} and L-Eval~\citep{an2023eval} have expanded evaluation criteria to encompass real-world applications, particularly in query-based content summarization. The emergence of many-shot learning as a distinct paradigm for extended context processing~\cite{agarwal2024many} has added another dimension to this field. Notable contributions from LongGenBench~\cite{longgenbench} have advanced evaluation methodologies by combining extensive response generation requirements with efficient, cost-effective quality metrics.

The development of retrieval-focused benchmarks has taken a distinct approach, predominantly utilizing constructed datasets that enable precise experimental control, particularly in managing input sequence lengths. This methodology helps neutralize variations in model performance stemming from differences in training approaches. Substantial research efforts have yielded specialized synthetic frameworks for assessing retrieval capabilities~\citep{needle, mohtashami2023landmark, longchat, liu2024lost, hsieh2024ruler}, while concurrent investigations have revealed the broader implications of extended context processing for enhanced reasoning capabilities~\citep{tay2020long}.

\section{Experiment Details}
\label{app:experiments_design}

\subsection{Detail Results}
\label{sec:detail-results}
This section provide the detailed results of experiments in this paper, the results are shown in the format of $x_{y}$, where $x$ is the performance of the method and $y$ is the $\Delta P$ from the \cref{eq:performance_change}.

\paragraph{Observation 1.} \textbf{KV cache compression methods show task-dependent performance degradation, WK and CSR are more robust to KV cache compression.}

The detailed results of different KV cache compression methods are shown in \cref{tab:kv-compression-obs1}, different tasks exhibit notably varied sensitivities to KV cache compression, particularly under aggressive compression ratios. At a 10\% compression ratio, MMLU demonstrates remarkable resilience with less than 1\% average performance degradation, while GSM8K experiences a severe average performance drop exceeding 35\%. Other tasks show moderate to significant degradation, ranging from 6.5\% to 17.2\%. This substantial variation in compression sensitivity across tasks suggests that the effectiveness of KV cache compression is highly task-dependent, necessitating careful consideration of the specific task requirements when determining appropriate compression ratios.

The \cref{tab:kv-compression-r1} compares the performance of R1-Distill-Llama-8B and LLaMA-3.1-8B-Instruct under different compression ratios. R1-Distill-Llama-8B demonstrates more robust performance under compression compared to LLaMA-3.1-8B-Instruct. While both models start with similar baseline performance (0.6938 vs 0.7945), R1-Distill shows significantly less performance degradation under aggressive compression. Specifically, at 30\% compression ratio, R1-Distill maintains a performance of 0.6407 (-7.66\%), while LLaMA-3.1-8B-Instruct drops to 0.7469 (-6.00\%). The difference becomes more pronounced at 10\% compression ratio, where R1-Distill achieves 0.5840 (-15.82\%) compared to LLaMA-3.1-8B-Instruct's sharp decline to 0.5143 (-35.30\%). This suggests that the multi-step reasoning capabilities of R1-Distill contribute to its resilience against aggressive KV cache compression, particularly in maintaining reasoning coherence under limited context conditions.
\begin{table}[h]
    \caption{Performance Comparison of Different KV Cache Compression Methods on Instruction-Tuning Model and Multi-Step Reasoning Model}
    \centering
    \resizebox{\textwidth}{!}{
    \begin{tabular}{lcccccc|c}
    \toprule
    \textbf{Benchmark} & \textbf{Ratio} & \textbf{StreamingLLM} & \textbf{H2O} & \textbf{SnapKV} & \textbf{PyramidKV} & \textbf{ChunkKV} & \textbf{Average $\uparrow$} \\
    \midrule
        \multirow{10}{*}{\textit{R1-AR}}
    & Baseline  & \multicolumn{5}{c|}{R1-Distill-Llama-8B FullKV: 0.6938} & \\ \cmidrule{2-8}
    & $90\%$  & $0.7167_{(+3.30\%)}$ & $0.6900_{(-0.55\%)}$ & $0.6933_{(-0.07\%)}$ & $0.7100_{(+2.34\%)}$ & $0.6867_{(-1.02\%)}$ & $0.6993_{(+0.79\%)}$ \\
    & $80\%$  & $0.6867_{(-1.02\%)}$ & $0.6933_{(-0.07\%)}$ & $0.6933_{(-0.07\%)}$ & $0.7067_{(+1.86\%)}$ & $0.6767_{(-2.47\%)}$ & $0.6913_{(-0.36\%)}$ \\
    & $70\%$  & $0.6933_{(-0.07\%)}$ & $0.6633_{(-4.40\%)}$ & $0.7100_{(+2.34\%)}$ & $0.7100_{(+2.34\%)}$ & $0.7000_{(+0.89\%)}$ & $0.6953_{(+0.22\%)}$ \\
    & $60\%$  & $0.6833_{(-1.51\%)}$ & $0.6900_{(-0.55\%)}$ & $0.6900_{(-0.55\%)}$ & $0.7133_{(+2.81\%)}$ & $0.7067_{(+1.86\%)}$ & $0.6967_{(+0.42\%)}$ \\
    & $50\%$  & $0.6700_{(-3.43\%)}$ & $0.6967_{(+0.42\%)}$ & $0.7067_{(+1.86\%)}$ & $0.7000_{(+0.89\%)}$ & $0.6867_{(-1.02\%)}$ & $0.6920_{(-0.26\%)}$ \\
    & $40\%$  & $0.6767_{(-2.47\%)}$ & $0.6800_{(-1.99\%)}$ & $0.5967_{(-13.99\%)}$ & $0.6967_{(+0.42\%)}$ & $0.7133_{(+2.81\%)}$ & $0.6727_{(-3.04\%)}$ \\
    & $30\%$  & $0.6600_{(-4.87\%)}$ & $0.5900_{(-14.96\%)}$ & $0.5833_{(-15.93\%)}$ & $0.6700_{(-3.43\%)}$ & $0.7000_{(+0.89\%)}$ & $0.6407_{(-7.66\%)}$ \\
    & $20\%$  & $0.6200_{(-10.64\%)}$ & $0.4933_{(-28.90\%)}$ & $0.5633_{(-18.81\%)}$ & $0.6833_{(-1.51\%)}$ & $0.6533_{(-5.84\%)}$ & $0.6026_{(-13.14\%)}$ \\
    & $10\%$  & $0.5167_{(-25.53\%)}$ & $0.5567_{(-19.76\%)}$ & $0.5767_{(-16.88\%)}$ & $0.6267_{(-9.67\%)}$ & $0.6433_{(-7.28\%)}$ & $0.5840_{(-15.82\%)}$ \\

    \midrule
    \multirow{10}{*}{\textit{AR}}
    & Baseline  & \multicolumn{5}{c|}{LLaMA-3.1-8B-Instruct FullKV: 0.7945} & \\ \cmidrule{2-8}
    & $90\%$  & $0.7695_{(-3.10\%)}$ & $0.7923_{(-0.30\%)}$ & $0.7839_{(-1.30\%)}$ & $0.7854_{(-1.10\%)}$ & $0.7824_{(-1.50\%)}$ & $0.7827_{(-1.50\%)}$ \\
    & $80\%$  & $0.7642_{(-3.80\%)}$ & $0.7938_{(-0.10\%)}$ & $0.7824_{(-1.50\%)}$ & $0.7900_{(-0.60\%)}$ & $0.7824_{(-1.50\%)}$ & $0.7826_{(-1.50\%)}$ \\
    & $70\%$  & $0.7642_{(-3.80\%)}$ & $0.7900_{(-0.60\%)}$ & $0.7923_{(-0.30\%)}$ & $0.7983_{(+0.50\%)}$ & $0.7809_{(-1.70\%)}$ & $0.7851_{(-1.20\%)}$ \\
    & $60\%$  & $0.7650_{(-3.70\%)}$ & $0.7809_{(-1.70\%)}$ & $0.7885_{(-0.80\%)}$ & $0.7923_{(-0.30\%)}$ & $0.7885_{(-0.80\%)}$ & $0.7830_{(-1.50\%)}$ \\
    & $50\%$  & $0.7657_{(-3.60\%)}$ & $0.7854_{(-1.10\%)}$ & $0.7847_{(-1.20\%)}$ & $0.7854_{(-1.10\%)}$ & $0.7824_{(-1.50\%)}$ & $0.7807_{(-1.70\%)}$ \\
    & $40\%$  & $0.7491_{(-5.70\%)}$ & $0.7688_{(-3.20\%)}$ & $0.7756_{(-2.40\%)}$ & $0.7839_{(-1.30\%)}$ & $0.7763_{(-2.30\%)}$ & $0.7707_{(-3.00\%)}$ \\
    & $30\%$  & $0.7051_{(-11.20\%)}$ & $0.7225_{(-9.10\%)}$ & $0.7619_{(-4.10\%)}$ & $0.7718_{(-2.90\%)}$ & $0.7733_{(-2.70\%)}$ & $0.7469_{(-6.00\%)}$ \\
    & $20\%$  & $0.6384_{(-19.70\%)}$ & $0.6406_{(-19.40\%)}$ & $0.6884_{(-13.40\%)}$ & $0.7142_{(-10.10\%)}$ & $0.7763_{(-2.30\%)}$ & $0.6916_{(-13.00\%)}$ \\
    & $10\%$  & $0.4784_{(-39.80\%)}$ & $0.4503_{(-43.30\%)}$ & $0.5034_{(-36.60\%)}$ & $0.4829_{(-39.20\%)}$ & $0.6566_{(-17.40\%)}$ & $0.5143_{(-35.30\%)}$ \\
    \bottomrule
    \end{tabular}
    }

    \label{tab:kv-compression-r1}
\end{table}

\begin{table}[!h]
    \caption{Performance Comparison of Different KV Cache Compression Methods on \bcmk{}}

    \centering

    \resizebox{\textwidth}{!}{
    \begin{tabular}{lcccccc|c}
    \toprule
    \textbf{Benchmark} & \textbf{Ratio} & \textbf{StreamingLLM} & \textbf{H2O} & \textbf{SnapKV} & \textbf{PyramidKV} & \textbf{ChunkKV} & \textbf{Average $\uparrow$} \\
    \midrule
    \multirow{10}{*}{\textit{WK}} 
    & Baseline  & \multicolumn{5}{c|}{FullKV: 0.6882} & \\ \cmidrule{2-8}
    & $90\%$  & $0.6882_{(+0.00\%)}$ & $0.6882_{(+0.00\%)}$ & $0.6882_{(+0.00\%)}$ & $0.6882_{(+0.00\%)}$ & $0.6882_{(+0.00\%)}$ & $0.6882_{(+0.00\%)}$ \\
    & $80\%$  & $0.6882_{(+0.00\%)}$ & $0.6882_{(+0.00\%)}$ & $0.6882_{(+0.00\%)}$ & $0.6882_{(+0.00\%)}$ & $0.6882_{(+0.00\%)}$ & $0.6882_{(+0.00\%)}$ \\
    & $70\%$  & $0.6881_{(-0.01\%)}$ & $0.6882_{(+0.00\%)}$ & $0.6882_{(+0.00\%)}$ & $0.6882_{(+0.00\%)}$ & $0.6882_{(+0.00\%)}$ & $0.6882_{(+0.00\%)}$ \\
    & $60\%$  & $0.6881_{(-0.01\%)}$ & $0.6882_{(+0.00\%)}$ & $0.6882_{(+0.00\%)}$ & $0.6882_{(+0.00\%)}$ & $0.6882_{(+0.00\%)}$ & $0.6882_{(+0.00\%)}$ \\
    & $50\%$  & $0.6881_{(-0.01\%)}$ & $0.6882_{(+0.00\%)}$ & $0.6882_{(+0.00\%)}$ & $0.6882_{(+0.00\%)}$ & $0.6882_{(+0.00\%)}$ & $0.6882_{(+0.00\%)}$ \\
    & $40\%$  & $0.6879_{(-0.04\%)}$ & $0.6882_{(+0.00\%)}$ & $0.6882_{(+0.00\%)}$ & $0.6882_{(+0.00\%)}$ & $0.6882_{(+0.00\%)}$ & $0.6881_{(-0.01\%)}$ \\
    & $30\%$  & $0.6876_{(-0.09\%)}$ & $0.6880_{(-0.03\%)}$ & $0.6880_{(-0.03\%)}$ & $0.6882_{(+0.00\%)}$ & $0.6882_{(+0.00\%)}$ & $0.6880_{(-0.03\%)}$ \\
    & $20\%$  & $0.6859_{(-0.33\%)}$ & $0.6878_{(-0.06\%)}$ & $0.6880_{(-0.03\%)}$ & $0.6882_{(+0.00\%)}$ & $0.6882_{(+0.00\%)}$ & $0.6876_{(-0.08\%)}$ \\    
    & $10\%$  & $0.6787_{(-1.38\%)}$ & $0.6852_{(-0.44\%)}$ & $0.6831_{(-0.74\%)}$ & $0.6882_{(0.00\%)}$ & $0.6842_{(-0.58\%)}$ & $0.6839_{(-0.63\%)}$ \\
    
    \midrule
    \multirow{10}{*}{\textit{AR}}
    & Baseline  & \multicolumn{5}{c|}{FullKV: 0.7945} & \\ \cmidrule{2-8}
    & $90\%$  & $0.7695_{(-3.10\%)}$ & $0.7923_{(-0.30\%)}$ & $0.7839_{(-1.30\%)}$ & $0.7854_{(-1.10\%)}$ & $0.7824_{(-1.50\%)}$ & $0.7827_{(-1.50\%)}$ \\
    & $80\%$  & $0.7642_{(-3.80\%)}$ & $0.7938_{(-0.10\%)}$ & $0.7824_{(-1.50\%)}$ & $0.7900_{(-0.60\%)}$ & $0.7824_{(-1.50\%)}$ & $0.7826_{(-1.50\%)}$ \\
    & $70\%$  & $0.7642_{(-3.80\%)}$ & $0.7900_{(-0.60\%)}$ & $0.7923_{(-0.30\%)}$ & $0.7983_{(+0.50\%)}$ & $0.7809_{(-1.70\%)}$ & $0.7851_{(-1.20\%)}$ \\
    & $60\%$  & $0.7650_{(-3.70\%)}$ & $0.7809_{(-1.70\%)}$ & $0.7885_{(-0.80\%)}$ & $0.7923_{(-0.30\%)}$ & $0.7885_{(-0.80\%)}$ & $0.7830_{(-1.50\%)}$ \\
    & $50\%$  & $0.7657_{(-3.60\%)}$ & $0.7854_{(-1.10\%)}$ & $0.7847_{(-1.20\%)}$ & $0.7854_{(-1.10\%)}$ & $0.7824_{(-1.50\%)}$ & $0.7807_{(-1.70\%)}$ \\
    & $40\%$  & $0.7491_{(-5.70\%)}$ & $0.7688_{(-3.20\%)}$ & $0.7756_{(-2.40\%)}$ & $0.7839_{(-1.30\%)}$ & $0.7763_{(-2.30\%)}$ & $0.7707_{(-3.00\%)}$ \\
    & $30\%$  & $0.7051_{(-11.20\%)}$ & $0.7225_{(-9.10\%)}$ & $0.7619_{(-4.10\%)}$ & $0.7718_{(-2.90\%)}$ & $0.7733_{(-2.70\%)}$ & $0.7469_{(-6.00\%)}$ \\
    & $20\%$  & $0.6384_{(-19.70\%)}$ & $0.6406_{(-19.40\%)}$ & $0.6884_{(-13.40\%)}$ & $0.7142_{(-10.10\%)}$ & $0.7763_{(-2.30\%)}$ & $0.6916_{(-13.00\%)}$ \\
    & $10\%$  & $0.4784_{(-39.80\%)}$ & $0.4503_{(-43.30\%)}$ & $0.5034_{(-36.60\%)}$ & $0.4829_{(-39.20\%)}$ & $0.6566_{(-17.40\%)}$ & $0.5143_{(-35.30\%)}$ \\
    \midrule
    \multirow{10}{*}{\textit{CSR}}
    & Baseline  & \multicolumn{5}{c|}{FullKV: 0.7748} & \\ \cmidrule{2-8}
    & $90\%$  & $0.7748_{(+0.00\%)}$ & $0.7748_{(+0.00\%)}$ & $0.7748_{(+0.00\%)}$ & $0.7748_{(+0.00\%)}$ & $0.7748_{(+0.00\%)}$ & $0.7748_{(+0.00\%)}$ \\
    & $80\%$  & $0.7748_{(+0.00\%)}$ & $0.7748_{(+0.00\%)}$ & $0.7748_{(+0.00\%)}$ & $0.7748_{(+0.00\%)}$ & $0.7748_{(+0.00\%)}$ & $0.7748_{(+0.00\%)}$ \\
    & $70\%$  & $0.7748_{(+0.00\%)}$ & $0.7748_{(+0.00\%)}$ & $0.7748_{(+0.00\%)}$ & $0.7748_{(+0.00\%)}$ & $0.7748_{(+0.00\%)}$ & $0.7748_{(+0.00\%)}$ \\
    & $60\%$  & $0.7748_{(+0.00\%)}$ & $0.7748_{(+0.00\%)}$ & $0.7748_{(+0.00\%)}$ & $0.7748_{(+0.00\%)}$ & $0.7748_{(+0.00\%)}$ & $0.7748_{(+0.00\%)}$ \\
    & $50\%$  & $0.7748_{(+0.00\%)}$ & $0.7748_{(+0.00\%)}$ & $0.7748_{(+0.00\%)}$ & $0.7748_{(+0.00\%)}$ & $0.7748_{(+0.00\%)}$ & $0.7748_{(+0.00\%)}$ \\
    & $40\%$  & $0.7748_{(+0.00\%)}$ & $0.7748_{(+0.00\%)}$ & $0.7748_{(+0.00\%)}$ & $0.7748_{(+0.00\%)}$ & $0.7748_{(+0.00\%)}$ & $0.7748_{(+0.00\%)}$ \\
    & $30\%$  & $0.7748_{(+0.00\%)}$ & $0.7748_{(+0.00\%)}$ & $0.7748_{(+0.00\%)}$ & $0.7748_{(+0.00\%)}$ & $0.7748_{(+0.00\%)}$ & $0.7748_{(+0.00\%)}$ \\
    & $20\%$  & $0.7174_{(-7.40\%)}$ & $0.7748_{(+0.00\%)}$ & $0.7740_{(-0.10\%)}$ & $0.7748_{(+0.00\%)}$ & $0.7699_{(-0.60\%)}$ & $0.7622_{(-1.60\%)}$ \\
    & $10\%$  & $0.6806_{(-12.20\%)}$ & $0.7510_{(-3.10\%)}$ & $0.7191_{(-7.20\%)}$ & $0.7723_{(-0.30\%)}$ & $0.7002_{(-9.60\%)}$ & $0.7246_{(-6.50\%)}$ \\

    \midrule
    \multirow{10}{*}{\textit{SA}}
    & Baseline  & \multicolumn{5}{c|}{FullKV: 0.8895} & \\ \cmidrule{2-8}
    & $90\%$  & $0.8893_{(-0.00\%)}$ & $0.8890_{(-0.10\%)}$ & $0.8894_{(-0.00\%)}$ & $0.8893_{(-0.00\%)}$ & $0.8896_{(+0.00\%)}$ & $0.8893_{(-0.00\%)}$ \\
    & $80\%$  & $0.8878_{(-0.20\%)}$ & $0.8885_{(-0.10\%)}$ & $0.8895_{(+0.00\%)}$ & $0.8891_{(-0.00\%)}$ & $0.8894_{(-0.00\%)}$ & $0.8889_{(-0.10\%)}$ \\
    & $70\%$  & $0.8872_{(-0.30\%)}$ & $0.8879_{(-0.20\%)}$ & $0.8896_{(+0.00\%)}$ & $0.8889_{(-0.10\%)}$ & $0.8895_{(+0.00\%)}$ & $0.8886_{(-0.10\%)}$ \\
    & $60\%$  & $0.8845_{(-0.60\%)}$ & $0.8848_{(-0.50\%)}$ & $0.8892_{(-0.00\%)}$ & $0.8887_{(-0.10\%)}$ & $0.8899_{(+0.00\%)}$ & $0.8874_{(-0.20\%)}$ \\
    & $50\%$  & $0.8849_{(-0.50\%)}$ & $0.8749_{(-1.60\%)}$ & $0.8886_{(-0.10\%)}$ & $0.8884_{(-0.10\%)}$ & $0.8894_{(-0.00\%)}$ & $0.8852_{(-0.50\%)}$ \\
    & $40\%$  & $0.8734_{(-1.80\%)}$ & $0.8557_{(-3.80\%)}$ & $0.8880_{(-0.20\%)}$ & $0.8877_{(-0.20\%)}$ & $0.8900_{(+0.10\%)}$ & $0.8790_{(-1.20\%)}$ \\
    & $30\%$  & $0.8329_{(-6.40\%)}$ & $0.8015_{(-9.90\%)}$ & $0.8858_{(-0.40\%)}$ & $0.8899_{(+0.00\%)}$ & $0.8846_{(-0.60\%)}$ & $0.8589_{(-3.50\%)}$ \\
    & $20\%$  & $0.6501_{(-26.90\%)}$ & $0.7178_{(-19.30\%)}$ & $0.8806_{(-1.00\%)}$ & $0.8751_{(-1.60\%)}$ & $0.8902_{(+0.10\%)}$ & $0.8028_{(-9.70\%)}$ \\
    & $10\%$  & $0.5314_{(-40.30\%)}$ & $0.6544_{(-26.40\%)}$ & $0.8434_{(-5.20\%)}$ & $0.8556_{(-3.80\%)}$ & $0.8799_{(-1.10\%)}$ & $0.7529_{(-15.40\%)}$ \\

    \midrule
    \multirow{10}{*}{\textit{CG}}
    & Baseline  & \multicolumn{5}{c|}{FullKV: 0.5122} & \\ \cmidrule{2-8}
    & $90\%$  & $0.5061_{(-1.20\%)}$ & $0.5183_{(+1.20\%)}$ & $0.5122_{(+0.00\%)}$ & $0.5122_{(+0.00\%)}$ & $0.5122_{(+0.00\%)}$ & $0.5122_{(+0.00\%)}$ \\
    & $80\%$  & $0.5061_{(-1.20\%)}$ & $0.5183_{(+1.20\%)}$ & $0.5183_{(+1.20\%)}$ & $0.5305_{(+3.60\%)}$ & $0.5061_{(-1.20\%)}$ & $0.5159_{(+0.70\%)}$ \\
    & $70\%$  & $0.5000_{(-2.40\%)}$ & $0.5244_{(+2.40\%)}$ & $0.5122_{(+0.00\%)}$ & $0.5183_{(+1.20\%)}$ & $0.5122_{(+0.00\%)}$ & $0.5134_{(+0.20\%)}$ \\
    & $60\%$  & $0.5061_{(-1.20\%)}$ & $0.5366_{(+4.80\%)}$ & $0.5366_{(+4.80\%)}$ & $0.5305_{(+3.60\%)}$ & $0.5244_{(+2.40\%)}$ & $0.5268_{(+2.90\%)}$ \\
    & $50\%$  & $0.4939_{(-3.60\%)}$ & $0.5427_{(+6.00\%)}$ & $0.5061_{(-1.20\%)}$ & $0.4939_{(-3.60\%)}$ & $0.4878_{(-4.80\%)}$ & $0.5049_{(-1.40\%)}$ \\
    & $40\%$  & $0.4817_{(-6.00\%)}$ & $0.5427_{(+6.00\%)}$ & $0.5244_{(+2.40\%)}$ & $0.4939_{(-3.60\%)}$ & $0.5000_{(-2.40\%)}$ & $0.5085_{(-0.70\%)}$ \\
    & $30\%$  & $0.4817_{(-6.00\%)}$ & $0.5305_{(+3.60\%)}$ & $0.5000_{(-2.40\%)}$ & $0.4939_{(-3.60\%)}$ & $0.4817_{(-6.00\%)}$ & $0.4976_{(-2.90\%)}$ \\
    & $20\%$  & $0.4634_{(-9.50\%)}$ & $0.5061_{(-1.20\%)}$ & $0.4939_{(-3.60\%)}$ & $0.4695_{(-8.30\%)}$ & $0.4878_{(-4.80\%)}$ & $0.4841_{(-5.50\%)}$ \\
    & $10\%$  & $0.3659_{(-28.60\%)}$ & $0.4634_{(-9.50\%)}$ & $0.4268_{(-16.70\%)}$ & $0.4207_{(-17.90\%)}$ & $0.4451_{(-13.10\%)}$ & $0.4244_{(-17.20\%)}$ \\
    \bottomrule
    \end{tabular}
    }
    \label{tab:kv-compression-obs1}
\end{table}

On safety-focused evaluations, we observe that aggressive compression can disproportionately degrade performance, plausibly because compression may discard or fragment subtle safety-critical keywords and phrases present in system prompts; this disruption can weaken safety constraints during generation.

\paragraph{Observation 2.} \textbf{Multi-step reasoning LLMs are more robust to KV cache compression.}
As shown in \cref{tab:gsm8k-instruct}, while instruct-tuned models achieve superior baseline performance (0.7945 vs 0.5122), they demonstrate heightened sensitivity to KV cache compression. This sensitivity becomes particularly pronounced at aggressive compression ratios. At 10\% compression ratio, instruct-tuned models suffer an average performance degradation of 35.3\% (from 0.7945 to 0.5143), nearly double the degradation observed in non-instruct-tuned models which show a 17.2\% drop (from 0.5122 to 0.4244). In contrast, R1-Distill-Llama-8B shows better resilience to compression, with only a 15.82\% performance drop (from 0.6938 to 0.5840) at 10\% compression ratio. This pattern suggests that while instruction tuning enhances model capabilities, it also makes the model more dependent on maintaining complete context information. However, models trained with multi-step reasoning capabilities like R1-Distill demonstrate better robustness against aggressive compression, likely due to their enhanced ability to maintain reasoning coherence even with limited context.
We hypothesize that the reinforcement learning objective that explicitly incentivizes multi-step reasoning in DeepSeek-R1 yields more structured and robust internal representations of reasoning chains, making them less fragile to KV cache compression.

\begin{table}[!h]
    \centering

    \resizebox{\textwidth}{!}{
        \begin{tabular}{lcccccc|c}
            \toprule
            \textbf{Setting} & \textbf{Ratio} & \textbf{StreamingLLM} & \textbf{H2O} & \textbf{SnapKV} & \textbf{PyramidKV} & \textbf{ChunkKV} & \textbf{Average $\uparrow$ } \\
            \midrule

            \multirow{10}{*}{w/ Instruct Tuning}
            & Baseline  & \multicolumn{5}{c|}{FullKV: 0.7945} & \\ \cmidrule{2-8}
            & $90\%$  & $0.7695_{(-3.10\%)}$ & $0.7923_{(-0.30\%)}$ & $0.7839_{(-1.30\%)}$ & $0.7854_{(-1.10\%)}$ & $0.7824_{(-1.50\%)}$ & $0.7827_{(-1.50\%)}$ \\
            & $80\%$  & $0.7642_{(-3.80\%)}$ & $0.7938_{(-0.10\%)}$ & $0.7824_{(-1.50\%)}$ & $0.7900_{(-0.60\%)}$ & $0.7824_{(-1.50\%)}$ & $0.7826_{(-1.50\%)}$ \\
            & $70\%$  & $0.7642_{(-3.80\%)}$ & $0.7900_{(-0.60\%)}$ & $0.7923_{(-0.30\%)}$ & $0.7983_{(+0.50\%)}$ & $0.7809_{(-1.70\%)}$ & $0.7851_{(-1.20\%)}$ \\
            & $60\%$  & $0.7650_{(-3.70\%)}$ & $0.7809_{(-1.70\%)}$ & $0.7885_{(-0.80\%)}$ & $0.7923_{(-0.30\%)}$ & $0.7885_{(-0.80\%)}$ & $0.7830_{(-1.50\%)}$ \\
            & $50\%$  & $0.7657_{(-3.60\%)}$ & $0.7854_{(-1.10\%)}$ & $0.7847_{(-1.20\%)}$ & $0.7854_{(-1.10\%)}$ & $0.7824_{(-1.50\%)}$ & $0.7807_{(-1.70\%)}$ \\
            & $40\%$  & $0.7491_{(-5.70\%)}$ & $0.7688_{(-3.20\%)}$ & $0.7756_{(-2.40\%)}$ & $0.7839_{(-1.30\%)}$ & $0.7763_{(-2.30\%)}$ & $0.7707_{(-3.00\%)}$ \\
            & $30\%$  & $0.7051_{(-11.20\%)}$ & $0.7225_{(-9.10\%)}$ & $0.7619_{(-4.10\%)}$ & $0.7718_{(-2.90\%)}$ & $0.7733_{(-2.70\%)}$ & $0.7469_{(-6.00\%)}$ \\
            & $20\%$  & $0.6384_{(-19.70\%)}$ & $0.6406_{(-19.40\%)}$ & $0.6884_{(-13.40\%)}$ & $0.7142_{(-10.10\%)}$ & $0.7763_{(-2.30\%)}$ & $0.6916_{(-13.00\%)}$ \\
            & $10\%$  & $0.4784_{(-39.80\%)}$ & $0.4503_{(-43.30\%)}$ & $0.5034_{(-36.60\%)}$ & $0.4829_{(-39.20\%)}$ & $0.6566_{(-17.40\%)}$ & $0.5143_{(-35.30\%)}$ \\
            \midrule

            \multirow{10}{*}{w/ R1 Distill}
        & Baseline  & \multicolumn{5}{c|}{R1-Distill-Llama-8B FullKV: 0.6938} & \\ \cmidrule{2-8}
        & $90\%$  & $0.7167_{(+3.30\%)}$ & $0.6900_{(-0.55\%)}$ & $0.6933_{(-0.07\%)}$ & $0.7100_{(+2.34\%)}$ & $0.6867_{(-1.02\%)}$ & $0.6993_{(+0.79\%)}$ \\
        & $80\%$  & $0.6867_{(-1.02\%)}$ & $0.6933_{(-0.07\%)}$ & $0.6933_{(-0.07\%)}$ & $0.7067_{(+1.86\%)}$ & $0.6767_{(-2.47\%)}$ & $0.6913_{(-0.36\%)}$ \\
        & $70\%$  & $0.6933_{(-0.07\%)}$ & $0.6633_{(-4.40\%)}$ & $0.7100_{(+2.34\%)}$ & $0.7100_{(+2.34\%)}$ & $0.7000_{(+0.89\%)}$ & $0.6953_{(+0.22\%)}$ \\
        & $60\%$  & $0.6833_{(-1.51\%)}$ & $0.6900_{(-0.55\%)}$ & $0.6900_{(-0.55\%)}$ & $0.7133_{(+2.81\%)}$ & $0.7067_{(+1.86\%)}$ & $0.6967_{(+0.42\%)}$ \\
        & $50\%$  & $0.6700_{(-3.43\%)}$ & $0.6967_{(+0.42\%)}$ & $0.7067_{(+1.86\%)}$ & $0.7000_{(+0.89\%)}$ & $0.6867_{(-1.02\%)}$ & $0.6920_{(-0.26\%)}$ \\
        & $40\%$  & $0.6767_{(-2.47\%)}$ & $0.6800_{(-1.99\%)}$ & $0.5967_{(-13.99\%)}$ & $0.6967_{(+0.42\%)}$ & $0.7133_{(+2.81\%)}$ & $0.6727_{(-3.04\%)}$ \\
        & $30\%$  & $0.6600_{(-4.87\%)}$ & $0.5900_{(-14.96\%)}$ & $0.5833_{(-15.93\%)}$ & $0.6700_{(-3.43\%)}$ & $0.7000_{(+0.89\%)}$ & $0.6407_{(-7.66\%)}$ \\
        & $20\%$  & $0.6200_{(-10.64\%)}$ & $0.4933_{(-28.90\%)}$ & $0.5633_{(-18.81\%)}$ & $0.6833_{(-1.51\%)}$ & $0.6533_{(-5.84\%)}$ & $0.6026_{(-13.14\%)}$ \\
        & $10\%$  & $0.5167_{(-25.53\%)}$ & $0.5567_{(-19.76\%)}$ & $0.5767_{(-16.88\%)}$ & $0.6267_{(-9.67\%)}$ & $0.6433_{(-7.28\%)}$ & $0.5840_{(-15.82\%)}$ \\
        \midrule

        \multirow{10}{*}{w/o Instruct Tuning} 
        & Baseline  & \multicolumn{5}{c|}{FullKV: 0.5122} & \\ \cmidrule{2-8}
        & $90\%$  & $0.5061_{(-1.20\%)}$ & $0.5183_{(+1.20\%)}$ & $0.5122_{(+0.00\%)}$ & $0.5122_{(+0.00\%)}$ & $0.5122_{(+0.00\%)}$ & $0.5122_{(+0.00\%)}$ \\
        & $80\%$  & $0.5061_{(-1.20\%)}$ & $0.5183_{(+1.20\%)}$ & $0.5183_{(+1.20\%)}$ & $0.5305_{(+3.60\%)}$ & $0.5061_{(-1.20\%)}$ & $0.5159_{(+0.70\%)}$ \\
        & $70\%$  & $0.5000_{(-2.40\%)}$ & $0.5244_{(+2.40\%)}$ & $0.5122_{(+0.00\%)}$ & $0.5183_{(+1.20\%)}$ & $0.5122_{(+0.00\%)}$ & $0.5134_{(+0.20\%)}$ \\
        & $60\%$  & $0.5061_{(-1.20\%)}$ & $0.5366_{(+4.80\%)}$ & $0.5366_{(+4.80\%)}$ & $0.5305_{(+3.60\%)}$ & $0.5244_{(+2.40\%)}$ & $0.5268_{(+2.90\%)}$ \\
        & $50\%$  & $0.4939_{(-3.60\%)}$ & $0.5427_{(+6.00\%)}$ & $0.5061_{(-1.20\%)}$ & $0.4939_{(-3.60\%)}$ & $0.4878_{(-4.80\%)}$ & $0.5049_{(-1.40\%)}$ \\
        & $40\%$  & $0.4817_{(-6.00\%)}$ & $0.5427_{(+6.00\%)}$ & $0.5244_{(+2.40\%)}$ & $0.4939_{(-3.60\%)}$ & $0.5000_{(-2.40\%)}$ & $0.5085_{(-0.70\%)}$ \\
        & $30\%$  & $0.4817_{(-6.00\%)}$ & $0.5305_{(+3.60\%)}$ & $0.5000_{(-2.40\%)}$ & $0.4939_{(-3.60\%)}$ & $0.4817_{(-6.00\%)}$ & $0.4976_{(-2.90\%)}$ \\
        & $20\%$  & $0.4634_{(-9.50\%)}$ & $0.5061_{(-1.20\%)}$ & $0.4939_{(-3.60\%)}$ & $0.4695_{(-8.30\%)}$ & $0.4878_{(-4.80\%)}$ & $0.4841_{(-5.50\%)}$ \\
        & $10\%$  & $0.3659_{(-28.60\%)}$ & $0.4634_{(-9.50\%)}$ & $0.4268_{(-16.70\%)}$ & $0.4207_{(-17.90\%)}$ & $0.4451_{(-13.10\%)}$ & $0.4244_{(-17.20\%)}$ \\
        \bottomrule
    \end{tabular}
    }
    \caption{KV Cache Compression Performance Comparison on \textit{Arithmetic Reasoning} with Different Instruction TuningSettings}
    \label{tab:gsm8k-instruct}
    \end{table}

\paragraph{Observation 3.} \textbf{Short prompt length is more sensitive to KV cache compression.}
As demonstrated in \cref{tab:gsm8k-shots}, the impact of KV cache compression varies significantly with the number of shots in the prompt. One-shot prompts show extreme vulnerability to aggressive compression, with performance plummeting from 0.7149 to 0.0452 (a 93.7\% drop) at 10\% compression ratio. This sensitivity gradually decreases as the number of shots increases. For instance, at the same compression ratio, 4-shot prompts show a 46.2\% performance drop (from 0.7597 to 0.4088), while 8-shot prompts demonstrate relatively better resilience with a 35.3\% reduction (from 0.7945 to 0.5143). This pattern suggests that longer prompts with more examples provide redundancy that helps maintain model performance under compression, while shorter prompts lack this buffer against information loss.

\begin{table}[h]
    \caption{Performance Comparison of Different Shot Numbers on GSM8K}

    \centering

    \resizebox{\textwidth}{!}{
        \begin{tabular}{lcccccc|c}
            \toprule
            \textbf{Shot} & \textbf{Ratio} & \textbf{StreamingLLM} & \textbf{H2O} & \textbf{SnapKV} & \textbf{PyramidKV} & \textbf{ChunkKV} & \textbf{Average $\uparrow$} \\
            \midrule
            \multirow{10}{*}{1-shot} 
            & Baseline & \multicolumn{5}{c|}{FullKV: 0.7149} & \\ \cmidrule{2-8}
            & $90\%$  & $0.7013_{(-1.90\%)}$ & $0.7172_{(+0.30\%)}$ & $0.7142_{(-0.10\%)}$ & $0.7020_{(-1.80\%)}$ & $0.7172_{(+0.30\%)}$ & $0.7104_{(-0.60\%)}$ \\
            & $80\%$  & $0.6892_{(-3.60\%)}$ & $0.7089_{(-0.80\%)}$ & $0.7066_{(-1.20\%)}$ & $0.6952_{(-2.80\%)}$ & $0.7081_{(-1.00\%)}$ & $0.7016_{(-1.90\%)}$ \\
            & $70\%$  & $0.6816_{(-4.70\%)}$ & $0.6914_{(-3.30\%)}$ & $0.6945_{(-2.90\%)}$ & $0.6884_{(-3.70\%)}$ & $0.7127_{(-0.30\%)}$ & $0.6937_{(-3.00\%)}$ \\
            & $60\%$  & $0.6884_{(-3.70\%)}$ & $0.6831_{(-4.40\%)}$ & $0.6914_{(-3.30\%)}$ & $0.6816_{(-4.70\%)}$ & $0.6990_{(-2.20\%)}$ & $0.6887_{(-3.70\%)}$ \\
            & $50\%$  & $0.6952_{(-2.80\%)}$ & $0.6596_{(-7.70\%)}$ & $0.6611_{(-7.50\%)}$ & $0.6717_{(-6.00\%)}$ & $0.6732_{(-5.80\%)}$ & $0.6722_{(-6.00\%)}$ \\
            & $40\%$  & $0.6657_{(-6.90\%)}$ & $0.6202_{(-13.20\%)}$ & $0.6065_{(-15.20\%)}$ & $0.6475_{(-9.40\%)}$ & $0.6050_{(-15.40\%)}$ & $0.6290_{(-12.00\%)}$ \\
            & $30\%$  & $0.5118_{(-28.40\%)}$ & $0.5004_{(-30.00\%)}$ & $0.5042_{(-29.50\%)}$ & $0.5898_{(-17.50\%)}$ & $0.4011_{(-43.90\%)}$ & $0.5015_{(-29.90\%)}$ \\
            & $20\%$  & $0.2320_{(-67.50\%)}$ & $0.2714_{(-62.00\%)}$ & $0.2654_{(-62.90\%)}$ & $0.3973_{(-44.40\%)}$ & $0.1319_{(-81.60\%)}$ & $0.2596_{(-63.70\%)}$ \\
            & $10\%$  & $0.0296_{(-95.90\%)}$ & $0.0243_{(-96.60\%)}$ & $0.0296_{(-95.90\%)}$ & $0.1236_{(-82.70\%)}$ & $0.0190_{(-97.30\%)}$ & $0.0452_{(-93.70\%)}$ \\
            \midrule
            \multirow{10}{*}{2-shot}
            & Baseline & \multicolumn{5}{c|}{FullKV: 0.7574} &\\ \cmidrule{2-8}
            & $90\%$  & $0.7544_{(-0.40\%)}$ & $0.7604_{(+0.40\%)}$ & $0.7574_{(+0.00\%)}$ & $0.7612_{(+0.50\%)}$ & $0.7627_{(+0.70\%)}$ & $0.7592_{(+0.20\%)}$ \\
            & $80\%$  & $0.7551_{(-0.30\%)}$ & $0.7521_{(-0.70\%)}$ & $0.7559_{(-0.20\%)}$ & $0.7559_{(-0.20\%)}$ & $0.7589_{(+0.20\%)}$ & $0.7556_{(-0.20\%)}$ \\
            & $70\%$  & $0.7521_{(-0.70\%)}$ & $0.7453_{(-1.60\%)}$ & $0.7566_{(-0.10\%)}$ & $0.7574_{(+0.00\%)}$ & $0.7642_{(+0.90\%)}$ & $0.7551_{(-0.30\%)}$ \\
            & $60\%$  & $0.7475_{(-1.30\%)}$ & $0.7506_{(-0.90\%)}$ & $0.7521_{(-0.70\%)}$ & $0.7589_{(+0.20\%)}$ & $0.7695_{(+1.60\%)}$ & $0.7557_{(-0.20\%)}$ \\
            & $50\%$  & $0.7460_{(-1.50\%)}$ & $0.7437_{(-1.80\%)}$ & $0.7437_{(-1.80\%)}$ & $0.7604_{(+0.40\%)}$ & $0.7619_{(+0.60\%)}$ & $0.7511_{(-0.80\%)}$ \\
            & $40\%$  & $0.7445_{(-1.70\%)}$ & $0.7081_{(-6.50\%)}$ & $0.7202_{(-4.90\%)}$ & $0.7309_{(-3.50\%)}$ & $0.7650_{(+1.00\%)}$ & $0.7337_{(-3.10\%)}$ \\
            & $30\%$  & $0.7506_{(-0.90\%)}$ & $0.6133_{(-19.00\%)}$ & $0.6657_{(-12.10\%)}$ & $0.7036_{(-7.10\%)}$ & $0.7445_{(-1.70\%)}$ & $0.6955_{(-8.20\%)}$ \\
            & $20\%$  & $0.6217_{(-17.90\%)}$ & $0.4412_{(-41.70\%)}$ & $0.4936_{(-34.80\%)}$ & $0.5534_{(-26.90\%)}$ & $0.5368_{(-29.10\%)}$ & $0.5293_{(-30.10\%)}$ \\
            & $10\%$  & $0.1516_{(-80.00\%)}$ & $0.1759_{(-76.80\%)}$ & $0.1622_{(-78.60\%)}$ & $0.2244_{(-70.40\%)}$ & $0.0735_{(-90.30\%)}$ & $0.1575_{(-79.20\%)}$ \\
            \midrule
            \multirow{10}{*}{4-shot}
            & Baseline & \multicolumn{5}{c|}{FullKV: 0.7597} & \\ \cmidrule{2-8}
            & $90\%$  & $0.7597_{(+0.00\%)}$ & $0.7604_{(+0.10\%)}$ & $0.7650_{(+0.70\%)}$ & $0.7642_{(+0.60\%)}$ & $0.7657_{(+0.80\%)}$ & $0.7630_{(+0.40\%)}$ \\
            & $80\%$  & $0.7559_{(-0.50\%)}$ & $0.7688_{(+1.20\%)}$ & $0.7695_{(+1.30\%)}$ & $0.7680_{(+1.10\%)}$ & $0.7642_{(+0.60\%)}$ & $0.7653_{(+0.70\%)}$ \\
            & $70\%$  & $0.7597_{(+0.00\%)}$ & $0.7695_{(+1.30\%)}$ & $0.7680_{(+1.10\%)}$ & $0.7710_{(+1.50\%)}$ & $0.7726_{(+1.70\%)}$ & $0.7682_{(+1.10\%)}$ \\
            & $60\%$  & $0.7369_{(-3.00\%)}$ & $0.7726_{(+1.70\%)}$ & $0.7688_{(+1.20\%)}$ & $0.7635_{(+0.50\%)}$ & $0.7718_{(+1.60\%)}$ & $0.7627_{(+0.40\%)}$ \\
            & $50\%$  & $0.7475_{(-1.60\%)}$ & $0.7612_{(+0.20\%)}$ & $0.7619_{(+0.30\%)}$ & $0.7665_{(+0.90\%)}$ & $0.7635_{(+0.50\%)}$ & $0.7601_{(+0.10\%)}$ \\
            & $40\%$  & $0.7165_{(-5.70\%)}$ & $0.7339_{(-3.40\%)}$ & $0.7377_{(-2.90\%)}$ & $0.7483_{(-1.50\%)}$ & $0.7612_{(+0.20\%)}$ & $0.7395_{(-2.70\%)}$ \\
            & $30\%$  & $0.6558_{(-13.70\%)}$ & $0.6603_{(-13.10\%)}$ & $0.7111_{(-6.40\%)}$ & $0.7263_{(-4.40\%)}$ & $0.7597_{(+0.00\%)}$ & $0.7026_{(-7.50\%)}$ \\
            & $20\%$  & $0.6224_{(-18.10\%)}$ & $0.5625_{(-26.00\%)}$ & $0.6065_{(-20.20\%)}$ & $0.6543_{(-13.90\%)}$ & $0.7468_{(-1.70\%)}$ & $0.6385_{(-16.00\%)}$ \\
            & $10\%$  & $0.4708_{(-38.00\%)}$ & $0.3980_{(-47.60\%)}$ & $0.3995_{(-47.40\%)}$ & $0.4321_{(-43.10\%)}$ & $0.3434_{(-54.80\%)}$ & $0.4088_{(-46.20\%)}$ \\
            \midrule
            \multirow{10}{*}{6-shot}
            & Baseline & \multicolumn{5}{c|}{FullKV: 0.7680} & \\ \cmidrule{2-8}
            & $90\%$  & $0.7551_{(-1.70\%)}$ & $0.7748_{(+0.90\%)}$ & $0.7839_{(+2.10\%)}$ & $0.7794_{(+1.50\%)}$ & $0.7794_{(+1.50\%)}$ & $0.7745_{(+0.90\%)}$ \\
            & $80\%$  & $0.7642_{(-0.50\%)}$ & $0.7756_{(+1.00\%)}$ & $0.7809_{(+1.70\%)}$ & $0.7741_{(+0.80\%)}$ & $0.7786_{(+1.40\%)}$ & $0.7747_{(+0.90\%)}$ \\
            & $70\%$  & $0.7513_{(-2.20\%)}$ & $0.7771_{(+1.20\%)}$ & $0.7809_{(+1.70\%)}$ & $0.7771_{(+1.20\%)}$ & $0.7786_{(+1.40\%)}$ & $0.7730_{(+0.70\%)}$ \\
            & $60\%$  & $0.7468_{(-2.80\%)}$ & $0.7748_{(+0.90\%)}$ & $0.7733_{(+0.70\%)}$ & $0.7771_{(+1.20\%)}$ & $0.7809_{(+1.70\%)}$ & $0.7706_{(+0.30\%)}$ \\
            & $50\%$  & $0.7407_{(-3.60\%)}$ & $0.7718_{(+0.50\%)}$ & $0.7718_{(+0.50\%)}$ & $0.7771_{(+1.20\%)}$ & $0.7718_{(+0.50\%)}$ & $0.7666_{(-0.20\%)}$ \\
            & $40\%$  & $0.7377_{(-3.90\%)}$ & $0.7506_{(-2.30\%)}$ & $0.7771_{(+1.20\%)}$ & $0.7688_{(+0.10\%)}$ & $0.7854_{(+2.30\%)}$ & $0.7639_{(-0.50\%)}$ \\
            & $30\%$  & $0.7058_{(-8.10\%)}$ & $0.7255_{(-5.50\%)}$ & $0.7392_{(-3.70\%)}$ & $0.7491_{(-2.50\%)}$ & $0.7763_{(+1.10\%)}$ & $0.7392_{(-3.70\%)}$ \\
            & $20\%$  & $0.5921_{(-22.90\%)}$ & $0.6232_{(-18.80\%)}$ & $0.6732_{(-12.30\%)}$ & $0.6960_{(-9.40\%)}$ & $0.7665_{(-0.20\%)}$ & $0.6702_{(-12.70\%)}$ \\
            & $10\%$  & $0.4572_{(-40.50\%)}$ & $0.4481_{(-41.60\%)}$ & $0.4958_{(-35.40\%)}$ & $0.4458_{(-41.90\%)}$ & $0.5565_{(-27.50\%)}$ & $0.4807_{(-37.40\%)}$ \\
            \midrule
            \multirow{10}{*}{8-shot}
            & Baseline & \multicolumn{5}{c|}{FullKV: 0.7945} & \\ \cmidrule{2-8}
            & $90\%$  & $0.7695_{(-3.10\%)}$ & $0.7923_{(-0.30\%)}$ & $0.7839_{(-1.30\%)}$ & $0.7854_{(-1.10\%)}$ & $0.7824_{(-1.50\%)}$ & $0.7827_{(-1.50\%)}$ \\
            & $80\%$  & $0.7642_{(-3.80\%)}$ & $0.7938_{(-0.10\%)}$ & $0.7824_{(-1.50\%)}$ & $0.7900_{(-0.60\%)}$ & $0.7824_{(-1.50\%)}$ & $0.7826_{(-1.50\%)}$ \\
            & $70\%$  & $0.7642_{(-3.80\%)}$ & $0.7900_{(-0.60\%)}$ & $0.7923_{(-0.30\%)}$ & $0.7983_{(+0.50\%)}$ & $0.7809_{(-1.70\%)}$ & $0.7851_{(-1.20\%)}$ \\
            & $60\%$  & $0.7650_{(-3.70\%)}$ & $0.7809_{(-1.70\%)}$ & $0.7885_{(-0.80\%)}$ & $0.7923_{(-0.30\%)}$ & $0.7885_{(-0.80\%)}$ & $0.7830_{(-1.50\%)}$ \\
            & $50\%$  & $0.7657_{(-3.60\%)}$ & $0.7854_{(-1.10\%)}$ & $0.7847_{(-1.20\%)}$ & $0.7854_{(-1.10\%)}$ & $0.7824_{(-1.50\%)}$ & $0.7807_{(-1.70\%)}$ \\
            & $40\%$  & $0.7491_{(-5.70\%)}$ & $0.7688_{(-3.20\%)}$ & $0.7756_{(-2.40\%)}$ & $0.7839_{(-1.30\%)}$ & $0.7763_{(-2.30\%)}$ & $0.7707_{(-3.00\%)}$ \\
            & $30\%$  & $0.7051_{(-11.20\%)}$ & $0.7225_{(-9.10\%)}$ & $0.7619_{(-4.10\%)}$ & $0.7718_{(-2.90\%)}$ & $0.7733_{(-2.70\%)}$ & $0.7469_{(-6.00\%)}$ \\
            & $20\%$  & $0.6384_{(-19.70\%)}$ & $0.6406_{(-19.40\%)}$ & $0.6884_{(-13.40\%)}$ & $0.7142_{(-10.10\%)}$ & $0.7763_{(-2.30\%)}$ & $0.6916_{(-13.00\%)}$ \\
            & $10\%$  & $0.4784_{(-39.80\%)}$ & $0.4503_{(-43.30\%)}$ & $0.5034_{(-36.60\%)}$ & $0.4829_{(-39.20\%)}$ & $0.6566_{(-17.40\%)}$ & $0.5143_{(-35.30\%)}$ \\
    \bottomrule
    \end{tabular}
    }
    \label{tab:gsm8k-shots}
    \end{table}

\paragraph{Observation 4.} \textbf{Chunk-level compression is more effective for long-context structured reasoning tasks.}
As shown in \cref{tab:kv-compression}, ChunkKV demonstrates superior robustness across different compression ratios, particularly under aggressive compression settings. While other methods show significant performance degradation at 10\% compression ratio (StreamingLLM: -9.8\%, H2O: -37.8\%, SnapKV: -17.1\%, PyramidKV: -14.6\%), ChunkKV maintains relatively stable performance with only a -3.7\% drop. This stark contrast in performance suggests that chunk-level compression better preserves the essential contextual information needed for complex reasoning tasks. The method's effectiveness likely stems from its ability to maintain the structural integrity of related context segments, which is particularly crucial for tasks requiring extended logical reasoning and arithmetic operations.

\begin{table}[h]
    \caption{Performance Comparison of Different KV Cache Compression Methods on Many-shot GSM8K}
    \centering

    \resizebox{\textwidth}{!}{
        \begin{tabular}{lcccccc|c}
            \toprule
            \textbf{Benchmark} & \textbf{Ratio} & \textbf{StreamingLLM} & \textbf{H2O} & \textbf{SnapKV} & \textbf{PyramidKV} & \textbf{ChunkKV} & \textbf{Average $\uparrow$ } \\
            \midrule
            \multirow{21}{*}{\makecell{Many-shot \\GSM8K}}
            & Baseline & \multicolumn{5}{c|}{LLaMA-3.1-8B-Instruct FullKV: 0.8235} &  \\ \cmidrule{2-8}
            & $90\%$  & $0.7728_{(-6.16\%)}$ & $0.8142_{(-1.13\%)}$ & $0.8137_{(-1.19\%)}$ & $0.7932_{(-3.68\%)}$ & $0.8233_{(-0.02\%)}$ & $0.8034_{(-2.44\%)}$ \\
            & $80\%$  & $0.7935_{(-3.64\%)}$ & $0.8334_{(+1.20\%)}$ & $0.8138_{(-1.18\%)}$ & $0.8037_{(-2.40\%)}$ & $0.7932_{(-3.68\%)}$ & $0.8075_{(-1.94\%)}$ \\
            & $70\%$  & $0.8038_{(-2.39\%)}$ & $0.8136_{(-1.20\%)}$ & $0.7832_{(-4.89\%)}$ & $0.7932_{(-3.68\%)}$ & $0.8037_{(-2.40\%)}$ & $0.7995_{(-2.91\%)}$ \\
            & $60\%$  & $0.7932_{(-3.68\%)}$ & $0.8142_{(-1.13\%)}$ & $0.8037_{(-2.40\%)}$ & $0.7935_{(-3.64\%)}$ & $0.8038_{(-2.39\%)}$ & $0.8017_{(-2.65\%)}$ \\
            & $50\%$  & $0.7934_{(-3.65\%)}$ & $0.8137_{(-1.19\%)}$ & $0.7932_{(-3.68\%)}$ & $0.7932_{(-3.68\%)}$ & $0.7835_{(-4.86\%)}$ & $0.7954_{(-3.41\%)}$ \\
            & $40\%$  & $0.8037_{(-2.40\%)}$ & $0.7832_{(-4.89\%)}$ & $0.7935_{(-3.64\%)}$ & $0.7834_{(-4.87\%)}$ & $0.7832_{(-4.89\%)}$ & $0.7894_{(-4.14\%)}$ \\
            & $30\%$  & $0.7835_{(-4.86\%)}$ & $0.7932_{(-3.68\%)}$ & $0.8038_{(-2.39\%)}$ & $0.7934_{(-3.65\%)}$ & $0.7932_{(-3.68\%)}$ & $0.7934_{(-3.65\%)}$ \\
            & $20\%$  & $0.7537_{(-8.47\%)}$ & $0.7428_{(-9.80\%)}$ & $0.7934_{(-3.65\%)}$ & $0.7832_{(-4.89\%)}$ & $0.7835_{(-4.86\%)}$ & $0.7713_{(-6.34\%)}$ \\
            & $10\%$  & $0.7432_{(-9.75\%)}$ & $0.5127_{(-37.74\%)}$ & $0.6827_{(-17.10\%)}$ & $0.7037_{(-14.55\%)}$ & $0.7932_{(-3.68\%)}$ & $0.6871_{(-16.56\%)}$ \\
            \cmidrule{2-8}
            & Baseline & \multicolumn{5}{c|}{R1-Distill-Llama-8B FullKV: 0.7123} &  \\ \cmidrule{2-8}
            & $90\%$  & $0.7123_{(+1.42\%)}$ & $0.6612_{(-5.85\%)}$ & $0.6534_{(-6.96\%)}$ & $0.6912_{(-1.58\%)}$ & $0.6923_{(-1.42\%)}$ & $0.6821_{(-2.88\%)}$ \\
            & $80\%$  & $0.7234_{(+3.00\%)}$ & $0.6534_{(-6.96\%)}$ & $0.7123_{(+1.42\%)}$ & $0.6423_{(-8.54\%)}$ & $0.7123_{(+1.42\%)}$ & $0.6887_{(-1.94\%)}$ \\
            & $70\%$  & $0.7412_{(+5.54\%)}$ & $0.6523_{(-7.12\%)}$ & $0.7234_{(+3.00\%)}$ & $0.6923_{(-1.42\%)}$ & $0.7234_{(+3.00\%)}$ & $0.7065_{(+0.60\%)}$ \\
            & $60\%$  & $0.7423_{(+5.69\%)}$ & $0.6912_{(-1.58\%)}$ & $0.6912_{(-1.58\%)}$ & $0.6823_{(-2.85\%)}$ & $0.6634_{(-5.54\%)}$ & $0.6941_{(-1.17\%)}$ \\
            & $50\%$  & $0.7234_{(+3.00\%)}$ & $0.7134_{(+1.58\%)}$ & $0.7312_{(+4.12\%)}$ & $0.7123_{(+1.42\%)}$ & $0.7123_{(+1.42\%)}$ & $0.7185_{(+2.31\%)}$ \\
            & $40\%$  & $0.7123_{(+1.42\%)}$ & $0.6923_{(-1.42\%)}$ & $0.6923_{(-1.42\%)}$ & $0.7023_{(+0.00\%)}$ & $0.7234_{(+3.00\%)}$ & $0.7045_{(+0.31\%)}$ \\
            & $30\%$  & $0.6523_{(-7.12\%)}$ & $0.7312_{(+4.12\%)}$ & $0.6634_{(-5.54\%)}$ & $0.7423_{(+5.69\%)}$ & $0.6912_{(-1.58\%)}$ & $0.6961_{(-0.88\%)}$ \\
            & $20\%$  & $0.6912_{(-1.58\%)}$ & $0.5834_{(-16.93\%)}$ & $0.5123_{(-27.05\%)}$ & $0.6823_{(-2.85\%)}$ & $0.6634_{(-5.54\%)}$ & $0.6265_{(-10.79\%)}$ \\
            & $10\%$  & $0.6323_{(-9.97\%)}$ & $0.5423_{(-22.78\%)}$ & $0.5412_{(-22.94\%)}$ & $0.5923_{(-15.66\%)}$ & $0.6823_{(-2.85\%)}$ & $0.5981_{(-14.84\%)}$ \\
            \bottomrule
        \end{tabular}
    }

    \label{tab:kv-compression}
\end{table}



\subsection{Matched-setting comparison: retrieval vs. high-density reasoning}
\label{appendix:matched-setting}
To directly substantiate the claim in \cref{sec:introduction} that high-density reasoning is more compression-sensitive than retrieval-oriented long-context tasks, we compare relative degradation under \emph{matched} compression methods and ratios. We use LLaMA-3.1-8B-Instruct, fix the compression methods (ChunkKV and H2O), and apply identical compression ratios (10\% and 30\%) to two task families: a retrieval-oriented long-context workload (LongBench) and a high-density reasoning workload (many-shot GSM8K, prompt $\approx$4K tokens). \cref{tab:matched-setting} shows that LongBench-style retrieval remains comparatively stable, while many-shot GSM8K degrades sharply at 10\%---supporting our framing that retrieval-oriented benchmarks underestimate the impact of compression on reasoning.

\begin{table}[h]
    \caption{Matched-setting comparison on LLaMA-3.1-8B-Instruct. Same model, methods, and compression ratios applied to retrieval-oriented (LongBench) vs.\ high-density reasoning (many-shot GSM8K). $\Delta$ is absolute drop from FullKV.}
    \centering
    \resizebox{0.95\textwidth}{!}{
    \begin{tabular}{l|l|c|cc|cc}
    \toprule
    \multirow{2}{*}{\textbf{Method}} & \multirow{2}{*}{\textbf{Task family}} & \textbf{FullKV} & \multicolumn{2}{c|}{\textbf{10\%}} & \multicolumn{2}{c}{\textbf{30\%}} \\
    \cmidrule{4-7}
                 &                                 &       & Score & $\Delta$ & Score & $\Delta$ \\
    \midrule
    ChunkKV      & Retrieval-oriented (LongBench)  & 41.46 & 40.51 & $-0.95$  & 41.59 & $+0.13$ \\
    ChunkKV      & High-density reasoning (GSM8K)  & 79.45 & 65.66 & $-13.79$ & 77.63 & $-1.82$ \\
    \midrule
    H2O          & Retrieval-oriented (LongBench)  & 41.46 & 37.06 & $-4.40$  & 39.23 & $-2.23$ \\
    H2O          & High-density reasoning (GSM8K)  & 79.45 & 45.03 & $-34.42$ & 64.06 & $-15.39$ \\
    \bottomrule
    \end{tabular}}
    \vspace{-5pt}
    \label{tab:matched-setting}
\end{table}

\subsection{Ablation: Prefill-only vs. \method{}}
\label{appendix:lggsm8k}
To assess the contribution of the decoding-phase compression, we ablate it by retaining only prefill compression and removing dynamic decoding-side compression. The ablation is run on two long-context settings.

\paragraph{LG-GSM8K (LLaMA-3.1-8B-Instruct).}
As summarized in \cref{tab:ablation_lg_gsm8k}, the prefill-only variant substantially underperforms full \method{} across all compression ratios, confirming the importance of the prefill--decoding separation.

\begin{table}[h]
    \caption{LLaMA-3.1-8B-Instruct on LG-GSM8K: \method{} vs. Prefill-only.}
    \centering
    \resizebox{0.6\textwidth}{!}{
    \begin{tabular}{l|cccc}
    \toprule
    \textbf{Method} & \textbf{40\%} & \textbf{35\%} & \textbf{30\%} & \textbf{25\%} \\
    \midrule
    \method{}    & 47.33 & 41.33 & 38.33 & 26.83 \\
    Prefill-only & 42.15 & 30.14 & 27.63 & 18.59 \\
    \bottomrule
    \end{tabular}}
    \vspace{-5pt}
    \label{tab:ablation_lg_gsm8k}
\end{table}

\paragraph{Many-shot GSM8K (LLaMA-3.1-8B-Instruct).}
The same pattern holds on many-shot arithmetic reasoning (\cref{tab:ablation_gsm8k}): removing the decoding-side compression component reduces accuracy by $\sim$2--3 absolute points across ratios, with the gap widest under aggressive (10\%) compression.

\begin{table}[h]
    \caption{LLaMA-3.1-8B-Instruct on many-shot GSM8K: \method{} vs. Prefill-only.}
    \centering
    \resizebox{0.6\textwidth}{!}{
    \begin{tabular}{l|cccc}
    \toprule
    \textbf{Method} & \textbf{40\%} & \textbf{30\%} & \textbf{20\%} & \textbf{10\%} \\
    \midrule
    \method{}    & 81.07 & 80.82 & 80.57 & 80.37 \\
    Prefill-only & 79.07 & 78.82 & 78.57 & 77.26 \\
    \bottomrule
    \end{tabular}}
    \vspace{-5pt}
    \label{tab:ablation_gsm8k}
\end{table}

\subsection{Shot ordering and exemplar robustness}
\label{appendix:robust-shots}
To probe whether \method{}'s attention-based scoring is robust to superficial perturbations of the prompt, we run two controlled variations on many-shot GSM8K with LLaMA-3.1-8B-Instruct: \emph{(i) Re-ordered}, where we shuffle the order of the few-shot exemplars while keeping the set fixed; and \emph{(ii) Diff-Example}, where we replace the exemplar set with a different sample of GSM8K problems of the same size. \cref{tab:shot-robust} shows that \method{} is broadly stable to shot reordering and reasonably stable across exemplar choices: variations are within $\sim$1.0 absolute point across all compression ratios. We note that this analysis demonstrates robustness to ordering and exemplar replacement; a fully principled validation of the score under arbitrary semantic corruption remains future work.

\begin{table}[h]
    \caption{Shot-ordering and exemplar-choice robustness of \method{} on many-shot GSM8K (LLaMA-3.1-8B-Instruct).}
    \centering
    \resizebox{0.7\textwidth}{!}{
    \begin{tabular}{l|cccc}
    \toprule
    \textbf{Method} & \textbf{40\%} & \textbf{30\%} & \textbf{20\%} & \textbf{10\%} \\
    \midrule
    \method{}                       & 81.07 & 80.82 & 80.57 & 80.37 \\
    \method{} (Re-ordered 1)        & 80.82 & 80.57 & 80.34 & 80.15 \\
    \method{} (Re-ordered 2)        & 81.12 & 80.87 & 80.62 & 80.22 \\
    \method{} (Diff-Example 1)      & 80.87 & 80.14 & 79.96 & 79.72 \\
    \method{} (Diff-Example 2)      & 80.97 & 80.94 & 80.91 & 80.88 \\
    \bottomrule
    \end{tabular}}
    \vspace{-5pt}
    \label{tab:shot-robust}
\end{table}

\subsection{Extended efficiency comparison}
\label{appendix:efficiency-extended}
\cref{tab:efficiency} in the main paper compares end-to-end efficiency of \method{}, FullKV, and two representative compression baselines at the 4K--4K setting. For completeness, \cref{tab:efficiency-extended} reports the same configuration on the A40 server with batch size~1 across all four methods. \method{} achieves efficiency comparable to ChunkKV and SnapKV; the design objective is not a radically different speed profile, but rather to preserve reasoning-relevant semantic structure under similar efficiency budgets.

\begin{table}[h]
    \caption{Extended efficiency comparison on A40, batch size 1, input/output 4096/4096. Percentages denote relative gain over FullKV.}
    \centering
    \resizebox{0.7\textwidth}{!}{
    \begin{tabular}{l|cc}
    \toprule
    \textbf{Method} & \textbf{Latency (s)} $\downarrow$ & \textbf{Throughput (T/S)} $\uparrow$ \\
    \midrule
    FullKV    & 175.50          & 37.73          \\
    ChunkKV   & 160.32 (8.6\%)  & 41.30 (9.5\%)  \\
    SnapKV    & 163.45 (6.9\%)  & 40.51 (7.4\%)  \\
    \method{} & 162.85 (7.2\%)  & 41.12 (9.0\%)  \\
    \bottomrule
    \end{tabular}}
    \vspace{-5pt}
    \label{tab:efficiency-extended}
\end{table}

\subsection{More experiments on other models}
\label{appendix:other-models}
To further validate the generality of our findings, we evaluate the impact of KV cache compression on additional model families beyond LLaMA-3.1.

\begin{figure}[t]
    \centering
    \includegraphics[width=1\textwidth]{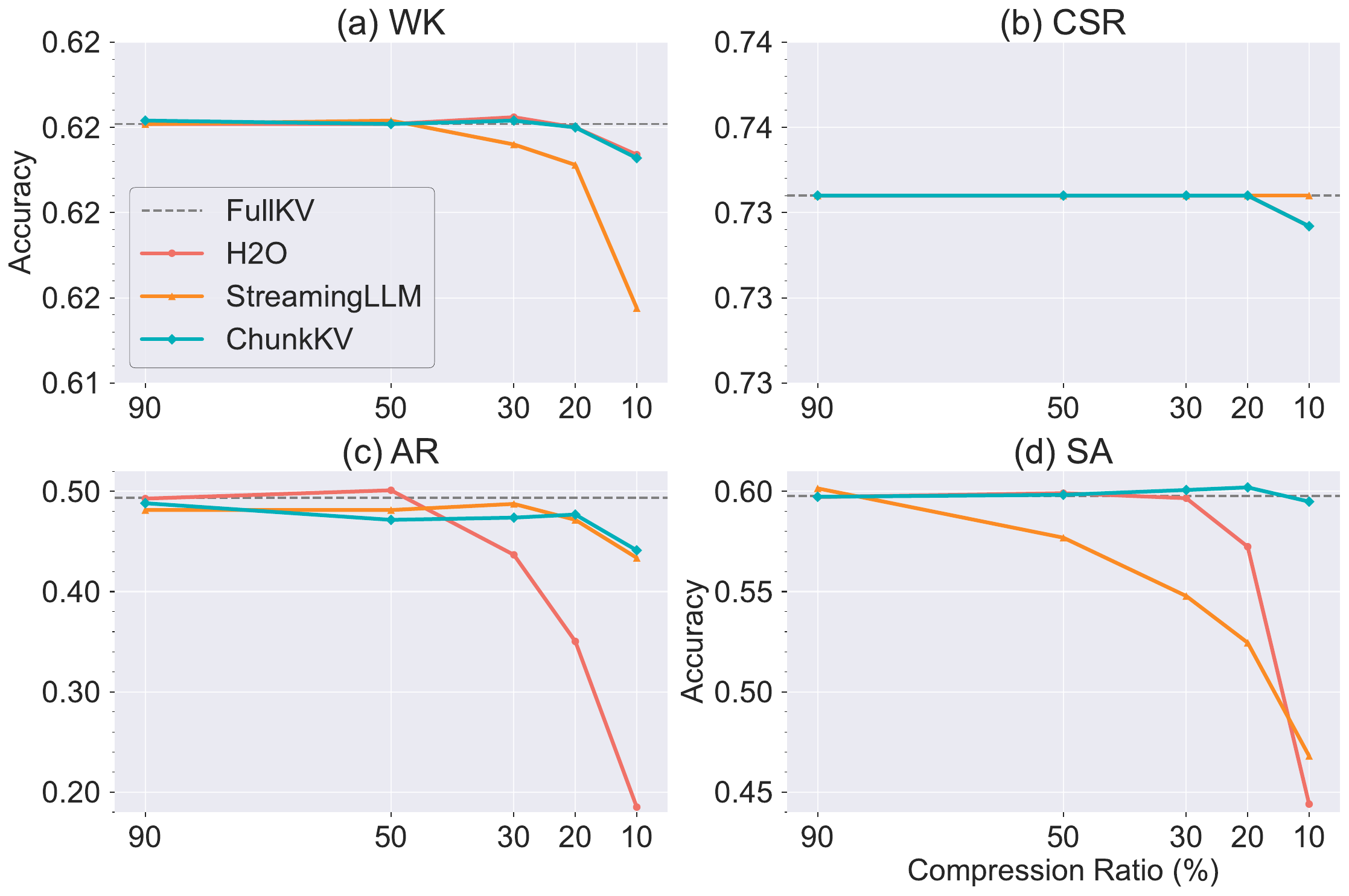}
    \caption{Performance Comparison of KV Cache Compression Methods Across Tasks with Mistral-7B-Instruct.}
    \label{fig:kv_cache_compression_rebuttal_mistral_7b_instruct}
\end{figure}

\paragraph{Mistral-7B-Instruct.}
As shown in Figure~\ref{fig:kv_cache_compression_rebuttal_mistral_7b_instruct}, we observe that various KV cache compression methods lead to significant performance degradation across multiple fundamental tasks, especially under aggressive compression ratios. This result demonstrates that the reduction in foundation abilities due to KV cache compression is not limited to a specific model family, but is a general phenomenon affecting different LLM architectures.

\paragraph{Qwen3-4B-Base.}
\label{appendix:qwen3}
We evaluate Qwen3-4B-Base on \bcmk{} to test cross-family generality. \cref{tab:qwen3} reports performance on Arithmetic Reasoning (AR) and World Knowledge (WK) at 10\% and 30\% compression with two representative methods. The same pattern observed on the LLaMA and Mistral families holds on Qwen3: aggressive token-level compression (H2O at 10\%) causes a substantial collapse on AR (87.79~$\rightarrow$~49.20), while WK remains comparatively robust; chunk-level retention (ChunkKV) significantly mitigates the AR degradation (87.79~$\rightarrow$~64.41 at 10\%). Together with the LLaMA-3.1 / Mistral / DeepSeek-R1 results elsewhere in the paper, this provides cross-family evidence for the High-Density-Reasoning vulnerability identified by \bcmk{}. The full \method{} pipeline on Qwen3 follows the same trend; we report the complete sweep below as it became available.

\begin{table}[t]
    \caption{Qwen3-4B-Base on \bcmk{}: AR (Arithmetic Reasoning) and WK (World Knowledge) at 10\% and 30\% compression.}
    \centering
    \resizebox{0.85\textwidth}{!}{
    \begin{tabular}{l|c|cc|cc}
    \toprule
    \multirow{2}{*}{\textbf{Task}} & \multirow{2}{*}{\textbf{FullKV}} & \multicolumn{2}{c|}{\textbf{H2O (token-level)}} & \multicolumn{2}{c}{\textbf{ChunkKV (chunk-level)}} \\
    \cmidrule{3-6}
       &       & 10\%   & 30\%   & 10\%   & 30\%   \\
    \midrule
    AR & 87.79 & 49.20  & 73.78  & 64.41  & 80.41 \\
    WK & 72.99 & 72.03  & 71.23  & 72.99  & 72.31 \\
    \bottomrule
    \end{tabular}}
    \vspace{-5pt}
    \label{tab:qwen3}
\end{table}

\section{\method{} Pseudocode and Algorithmic Details}
\label{app:shotkv-pseudocode}
This section provides the detailed algorithmic description of \method{}.

\subsection{Pseudocode}
The detailed algorithm of \method{} is presented in Algorithm~\ref{alg:shotkv}. Our method consists of two main phases: prefill compression and decoding compression. During the prefill phase, we compute an importance score for each shot by averaging the attention weights across all tokens, heads, and layers within that shot. This score $\text{Score}_{\text{prefill}}(s_i)$ is normalized by the shot length $k_i$ to avoid bias towards longer shots. Shots are then sorted by their scores and preserved until reaching the specified prefill ratio $r_p$.

In the decoding phase, compression is performed dynamically at each step. For each token in the decoding KV cache, we calculate its importance score $\text{Score}_{\text{decoding}}(t)$ by summing attention weights across all heads and layers. The top-k tokens are retained based on the decoding ratio $r_d$. Finally, the compressed KV cache is formed by combining both the preserved prefill and decoding caches.

This two-phase approach allows for different compression strategies during prefill and decoding stages, recognizing their distinct roles in the inference process. The shot-aware design during prefill ensures that the most informative examples are preserved, while the token-level compression during decoding maintains essential recent context.
\begin{algorithm}[t]
    \caption{ShotKV: Shot-aware KV Cache Compression}
    \label{alg:shotkv}
    \begin{algorithmic}[1]
        \REQUIRE Prompt with $n$ shots $\{s_1,...,s_n\}$, prefill ratio $r_p$, decoding ratio $r_d$, number of layers $L$
        \ENSURE Compressed KV cache $KV_{\text{total}}$

        \STATE Initialize $KV_{\text{total}} \leftarrow \emptyset$
        
        \FOR{each layer $l \in \{1, ..., L\}$}
            \STATE // Phase 1: Prefill Compression (Layer-wise)
            \FOR{each shot $s_i$ in $\{s_1,...,s_n\}$}
                \STATE Compute $\text{Score}_{\text{prefill}}^l(s_i) = \frac{1}{k_i} \sum_{t \in s_i} \sum_{h=1}^H \alpha_{t,h}^l$
            \ENDFOR
            \STATE Sort shots by $\text{Score}_{\text{prefill}}^l(s_i)$ in descending order
            \STATE $S_{\text{preserved}, l} \leftarrow$ Select shots until $\sum_{s_i} k_i \leq r_p \times |KV_{\text{prefill}}|$
            \STATE $KV^C_{\text{prefill}, l} \leftarrow \text{Compress}(\{s_i | s_i \in S_{\text{preserved}, l}\})$
            
            \STATE // Phase 2: Decoding Compression (Layer-wise, dynamic)
            \FOR{each decoding step}
                \FOR{each token $t$ in $KV_{\text{decoding}}$}
                    \STATE Compute $\text{Score}_{\text{decoding}}^l(t) = \sum_{h=1}^H \alpha_{t,h}^l$
                \ENDFOR
                \STATE $k \leftarrow r_d \times |KV_{\text{decoding}}|$
                \STATE $KV^C_{\text{decoding}, l} \leftarrow \text{TopK}(KV_{\text{decoding}}, \text{Score}_{\text{decoding}}^l, k)$
            \ENDFOR

            \STATE $KV_{\text{total}, l} \leftarrow KV^C_{\text{prefill}, l} \cup KV^C_{\text{decoding}, l}$
            \STATE $KV_{\text{total}} \leftarrow KV_{\text{total}} \cup \{KV_{\text{total}, l}\}$
        \ENDFOR
        
        \STATE \textbf{return} $KV_{\text{total}}$
    \end{algorithmic}
    \end{algorithm}

\section{Evaluation Benchmark}

\label{app:eval_bench}
\subsection{Dataset Details}
\label{app:eval_bench_dataset}

Detailed statistics for each benchmark dataset are provided in Table~\ref{tab:dataset_statistic}. For HotpotQA, we only report results under the 10\% compression ratio using the LLaMA-3-8B-Instruct model.

\begin{table}[t]

    \caption{The statistics of the datasets used in this paper. 
    \textsc{\# Test} denote the number of training data and test data, respectively.
    }
    \centering
    
    \footnotesize
    \resizebox{\linewidth}{!}{
    \begin{tabular}{l|c|r|c|c}
    \toprule
    \textsc{Dataset}  & \textsc{Task Type}  & \textsc{\# Test} & \textsc{Metric} & \textsc{Evaluation Method}\\ \midrule
    
    MMLU~\cite{mmlu}
    & World Knowledge 
    & 14,079 
    & Accuracy     
    & Generation-Based \\
    
    GSM8K~\cite{gsm8k}
    & Arithmetic      
    & 1,319   
    & Exact match  
    & Generation-Based \\
    
    CSQA~\cite{csqa}
    & Commonsense
    & 1,221
    & Accuracy 
    & Generation-Based \\
    
    HumanEval~\cite{chen2021evaluating}
    & Code Generation
    & 164
    & Pass@1 rate 
    & Generation-Based \\
    
    JailBreakV~\cite{luo2024jailbreakv}
    & Safety
    & 28,000
    & Attack success rate     
    & Generation-Based \\
    
    HotpotQA~\cite{hotpotqa}
    & Document QA (Multi-hop)
    & 7,405
    & Accuracy
    & Generation-Based \\
    
    LongGenBench~\cite{longgenbench}
    & Long-Context Generation
    & 23,000
    & Accuracy
    & Generation-Based \\
    \bottomrule
    \end{tabular}
    }
    \vspace{1em}

    \label{tab:dataset_statistic}
    \end{table}

The hyper-parameters for different observations are shown in \cref{tab:hyperparameters}. The temperature for the experiments are set to 0 for ensuring the deterministic results.

\begin{table}[t]
    \caption{Hyperparameters for Different Observations}
    \centering

    \resizebox{\columnwidth}{!}{%
    \begin{tabular}{l|ccccc|cc}
    \toprule
    \multirow{2}{*}{Benchmarks} & Obs 1 & Obs 2 & Obs 3 & Obs 4 & Obs 5 & \multicolumn{2}{c}{Obs 6} \\
    \cmidrule{2-6} \cmidrule{7-8}
    & \multicolumn{5}{c|}{Number of Shots} & $K$ & $T$ \\
    \midrule
    MMLU~\cite{mmlu} & 5 & 5 & - & - & 0,5 & - & - \\
    CommonsenseQA~\cite{csqa} & 4 & 4 & - & - & - & - & - \\
    GSM8K~\cite{gsm8k} & 8 & 8 & 1-8 & 50 & 0,8 & - & - \\
    HumanEval~\cite{chen2021evaluating} & 8 & 8 & - & - & - & - & - \\
    JailBreakV~\cite{luo2024jailbreakv} & 8 & 8 & - & - & - & - & - \\ \midrule
    LongGenBench-GSM8K~\cite{longgenbench} & - & - & - & - & - & 35 & 20 \\
    \bottomrule
    \end{tabular}
    }
    \label{tab:hyperparameters}
\end{table}

\end{document}